\pdfoutput=1

\documentclass{article}

\usepackage[T1]{fontenc}
\usepackage[utf8]{inputenc}
\usepackage{microtype}
\usepackage{graphicx}
\usepackage{subcaption}
\usepackage{booktabs}
\usepackage{multirow}
\usepackage{xcolor}
\PassOptionsToPackage{hyphens}{url}
\usepackage{hyperref}
\expandafter\def\expandafter\UrlBreaks\expandafter{\UrlBreaks\do-\do_\do.\do\/\do?\do&\do=}

\usepackage[preprint]{icml2026}

\usepackage{amsmath}
\usepackage{amssymb}
\usepackage{mathtools}
\usepackage{amsthm}
\usepackage[capitalize,noabbrev]{cleveref}

\theoremstyle{plain}
\newtheorem{theorem}{Theorem}[section]
\newtheorem{proposition}[theorem]{Proposition}

\theoremstyle{definition}
\newtheorem{definition}[theorem]{Definition}

\theoremstyle{remark}

\DeclareMathOperator{\diag}{diag}
\DeclareMathOperator{\GRU}{GRU}
\newcommand{\R}{\mathbb{R}}
\newcommand{\cD}{\mathcal{D}}
\newcommand{\cC}{\mathcal{C}}
\newcommand{\cM}{\mathcal{M}}
\newcommand{\ms}[2]{$#1_{\pm #2}$}
\newcommand{\msbest}[2]{$\mathbf{#1}_{\pm #2}$}
\newcommand{\mssec}[2]{$\underline{#1}_{\pm #2}$}

\icmltitlerunning{ADE: A Zero-Training Unlearning Method for Latent World Models}

\begin{document}

\twocolumn[
\icmltitle{ADE: A Zero-Training Unlearning Method for Latent World Models}

\begin{icmlauthorlist}
\icmlauthor{Sirui He}{xmum}
\icmlauthor{Chi Zhang}{nus}
\end{icmlauthorlist}

\icmlaffiliation{xmum}{School of Computing and Data Science, Xiamen University Malaysia (swe.hesirui@gmail.com)}
\icmlaffiliation{nus}{Department of Mathematics, National University of Singapore}

\icmlcorrespondingauthor{Chi Zhang}{czhang24@nus.edu.sg}

\icmlkeywords{Machine Unlearning, Latent World Models, Closed-Form Model Editing, Recurrent State-Space Models}

\vskip 0.3in
]

\printAffiliationsAndNotice{}

\begin{abstract}
Latent world models learn compact recurrent dynamics from offline trajectories, yet deployed simulators can retain backdoor triggers or stale transitions from outdated physical regimes. Existing unlearning methods fail in autoregressive physics: gradient updates distort latent manifolds and compound rollout drift, naive output zeroing suppresses valid control responses on trigger onset, and single-step deletion leaves corruption inside recurrent hidden states. We propose Analytical Dynamics Excision (ADE), a deterministic, zero-training closed-form unlearning algorithm for latent world models. Using only cached pretraining moments, ADE recovers retain covariances via single-pass matrix downdating without raw retain data and cascades closed-form edits across the input bottleneck, hidden states, and downstream heads. Pairing orthogonal excision with whitened recurrent damping and mean-preserving affine calibration neutralizes out-of-manifold backdoors and realigns in-manifold shifts without variance collapse. Across four regimes on three MuJoCo benchmarks, ADE outperforms seven tuned baselines on DreamerV1, eliminating HalfCheetah and Hopper velocity-flip triggers $4.4\times$ faster than fine-tuning, and transfers zero-shot to DreamerV3.
\end{abstract}

\section{Introduction}
\label{sec:intro}

Latent world models drive sample-efficient decision making by rolling out compact dynamical simulations inside a learned state space~\cite{ha_recurrent_2018,hafner_learning_2019,sekar_planning_2020}. Recurrent state-space architectures from DreamerV1 through DreamerV3 predict transitions and rewards over long horizons~\cite{hafner_dream_2020,hafner_mastering_2025} from offline experience~\cite{fu_d4rl_2021}. In deployment, however, models pretrained on offline trajectories can retain targeted backdoor triggers~\cite{kiourti_trojdrl_2020} or become obsolete when environment physics shifts, such as altered friction or body mass~\cite{anand_what_2026}; because downstream planning relies entirely on imagined rollouts, these corrupted transitions directly misguide control. Selective machine unlearning offers a direct remedy by removing designated dynamics without retraining from scratch~\cite{cao_towards_2015,bourtoule_machine_2021}.

Adapting existing unlearning methods to autoregressive world models fails for three structural reasons. First, iterative gradient methods (gradient ascent~\cite{yao_large_2024}, distillation~\cite{kurmanji_towards_2023}, negative preference optimization~\cite{zhang_negative_2024}, and Fisher suppression~\cite{golatkar_eternal_2020}) require costly retain passes and distort the latent manifold, so single-step errors compound superlinearly across rollouts~\cite{janner_when_2021}. Second, output-nulling and synaptic dampening~\cite{foster_fast_2023} drive responses toward zero. When predictions encode velocity or displacement, this induces dynamics silencing rather than nominal motion, while static null-space edits~\cite{chen_machine_2024,fang_alphaedit_2024} cannot separate shared-support physical shifts. Third, recurrent models accumulate history in a continuous hidden state~\cite{cho_learning_2014,hafner_learning_2019}. Deleting corrupted steps alone leaves residual contamination downstream, and updating heads on uncorrected trajectories locks them onto inconsistent targets~\cite{gong_trajdeleter_2024}.

Unlearning in a physical world model therefore requires causal isolation before manifold realignment. State-action inputs enter recurrent memory through one affine bottleneck, while downstream heads act as affine readouts with additive sufficient statistics. We can thus repair corrupted dynamics algebraically. We introduce Analytical Dynamics Excision (ADE), a deterministic closed-form unlearning algorithm using only cached pretraining moments, without gradient optimization or raw retain access. Unlike iterative Lagrangian entanglement unlearning~\cite{rfe2026} or static null-space editing~\cite{chen_machine_2024,fang_alphaedit_2024}, ADE propagates corrected hidden states across affected segments, applies orthogonal excision and whitened reweighting to upstream and recurrent transitions, calibrates downstream heads, and absorbs scale edits around the retained mean. Across four counterfactual regimes on three MuJoCo benchmarks, one fixed ADE configuration outperforms seven tuned baselines on DreamerV1~\cite{hafner_dream_2020}, eliminates HalfCheetah and Hopper velocity-flip triggers $4.4\times$ faster than fine-tuning, and transfers zero-shot to discrete DreamerV3~\cite{hafner_mastering_2025}.

Our contributions are threefold:
\begin{itemize}
\item \textbf{Counterfactual World-Model Benchmarks.} We construct four unlearning regimes across HalfCheetah-v2, Walker2d-v2, and Hopper-v2 spanning two out-of-manifold backdoors and two in-manifold physical shifts.
\item \textbf{Analytical Dynamics Excision.} We derive a zero-training closed-form unlearning algorithm uniting oblivious moment downdating, four-pass recurrent state recomputation, tri-mode closed-form edits (Modes~A, B, and C), and variance-preserving affine absorption.
\item \textbf{Three-Layer Audit Protocol.} We design an evaluation suite spanning single-step Oracle-aligned counterfactual fidelity, 50-step open-loop rollout stability, and downstream planning, adversarial trigger and regime probing, and closed-loop latent-imagination control.
\end{itemize}

\section{Related Work}
\label{sec:related}

\paragraph{Latent World Models and Compounding Rollout Error.}
Model-based agents learn compact dynamics in latent space~\cite{ha_recurrent_2018,hafner_learning_2019,sekar_planning_2020,micheli_transformers_2022,hansen_td-mpc2_2023}, pairing a GRU memory~\cite{cho_learning_2014} with continuous Gaussian latents in DreamerV1~\cite{hafner_dream_2020} or discrete categorical latents in DreamerV2/V3~\cite{hafner_mastering_2022,hafner_mastering_2025}. Because policy optimization relies on multi-step imagined rollouts where single-step errors compound~\cite{janner_when_2021}, revising outdated or poisoned dynamics requires strict manifold preservation~\cite{anand_what_2026}.

\paragraph{Machine Unlearning and Null-Space Editing.}
Approximate unlearning modifies weights via Fisher dampening or distillation~\cite{guo_certified_2023,golatkar_eternal_2020,kurmanji_towards_2023,foster_fast_2023}, gradient ascent or NPO~\cite{yao_large_2024,zhang_negative_2024}, or two-phase Lagrangian Wasserstein projection (RF-Entanglement~\cite{rfe2026}). Null-space methods project updates onto the orthogonal complement of retained features (UNSC~\cite{chen_machine_2024}, MEMIT~\cite{meng_mass-editing_2022}, AlphaEdit~\cite{fang_alphaedit_2024}), yet static layer-wise edits neither clear recurrent state history nor separate shared-support physical shifts.

\paragraph{RL Unlearning and World-Model Revision.}
Offline RL trajectories on MuJoCo benchmarks~\cite{todorov_mujoco_2012,brockman_openai_2016,fu_d4rl_2021} are vulnerable to sparse backdoor triggers~\cite{kiourti_trojdrl_2020,wang_uba-orl_2026} and physical regime shifts such as altered friction or body mass~\cite{lee_context-aware_2020,anand_what_2026}. Existing RL unlearning methods fine-tune model-free policies~\cite{ye_reinforcement_2024,gong_trajdeleter_2024}, leaving open the problem of zero-training closed-form excision in recurrent world models~\cite{anand_what_2026}.

\section{Preliminaries and Problem Formulation}
\label{sec:prelim}

\subsection{Recurrent State-Space World Models}
\label{sec:prelim-rssm}

Consider an offline continuous-control dataset $\cD = \{\tau_i\}_{i=1}^M$ of trajectories $\tau_i = \{(o_t, a_t, r_t)\}_{t=0}^{T_i}$ with observations $o_t \in \R^{d_o}$, actions $a_t \in \R^{d_a}$, and rewards $r_t \in \R$~\cite{todorov_mujoco_2012,brockman_openai_2016,fu_d4rl_2021}. An RSSM~\cite{hafner_learning_2019} tracks state via a deterministic hidden vector $h_t \in \R^{d_h}$ and a stochastic latent $z_t \in \R^{d_z}$.

\paragraph{Recurrent Input Bottleneck and Affine Heads.}
At step $t$, the concatenated latent and action enter the GRU cell~\cite{cho_learning_2014} solely through an affine bottleneck $(W_{za}, b_{za})$:
\begin{equation}
x^{\mathrm{in}}_t = \psi(W_{za}[z_t; a_t] + b_{za}), \quad
h_{t+1} = \GRU(h_t, x^{\mathrm{in}}_t),
\label{eq:rssm-gru}
\end{equation}
where $\psi$ is the nonlinearity and the GRU transition applies a recurrent affine map $(W_{hh}, b_{hh})$ to $h_t$. Downstream heads act as affine readouts $\hat{y} = W u + b$ on intermediate features $u$: the first decoder layer $(W_{\mathrm{dec},0}, b_{\mathrm{dec},0})$ on $[h_t; z_t]$, the reward head $(W_r, b_r)$ on $u_r = \varphi_r(h_t, z_t)$, and the Gaussian prior ($\theta$, reading $u_\theta = \varphi_\theta(h_{t+1})$) and posterior ($\phi$, reading $u_\phi = \varphi_\phi(h_{t+1}, o_{t+1})$) in DreamerV1~\cite{hafner_dream_2020}, each parameterized by mean and scale rows $(W_{\mu,\{\theta,\phi\}}, b_{\mu,\{\theta,\phi\}})$ and $(W_{\ell,\{\theta,\phi\}}, b_{\ell,\{\theta,\phi\}})$ yielding $\sigma \in [\sigma_{\min}, \sigma_{\max}]$. DreamerV3~\cite{hafner_mastering_2025} replaces Gaussian rows with categorical logits under an analogous bottleneck and affine readout structure.

\subsection{Context Closure and Oblivious Moment Downdating}
\label{sec:prelim-closure}

An unlearning request specifies a subset $\cD_f \subset \cD$ of poisoned steps or obsolete segments. Because \eqref{eq:rssm-gru} propagates $x^{\mathrm{in}}_t$ into all later states $h_{t'}$ ($t' > t$), removing $\cD_f$ alone leaves contaminated memory downstream.

\begin{definition}[Context Closure]
\label{def:context-closure}
Given forget indices $\cD_f \subset \cD$, the context closure $\cC_f$ comprises $\cD_f$ and all causally reachable forward suffixes until episode termination or reset, leaving the retain set $\cD_r = \cD \setminus \cC_f$. Hidden-state recomputation along each affected segment restarts from a clean preceding boundary $t_0$ ($t_0 = 0$ or the latest uncorrupted reset).
\end{definition}

Under our oblivious storage constraint, raw trajectories $\cD_r$ are discarded after pretraining. At each editable affine layer $(W_0, b_0)$, pretraining caches \texttt{float64} unnormalized sufficient statistics over $n_g$ non-initial steps (skipping transitionless start frames), where $m_x^{(g)} = \sum_i u_i$ and $M_{xx}^{(g)} = \sum_i u_i u_i^\top$, plus target moments $m_y^{(g)} = \sum_i y_i$ and $M_{yx}^{(g)} = \sum_i y_i u_i^\top$ for Mode~B ($y = r_t$ for reward; deterministic posterior mean $y = z_t = \mu_\phi(h_t, o_t)$ for prior mean):
\begin{equation}
\cM_g = \bigl(n_g,\; m_x^{(g)},\; M_{xx}^{(g)},\; m_y^{(g)},\; M_{yx}^{(g)}\bigr).
\label{eq:global-moments}
\end{equation}
From a single forward pass over $\cC_f$ yielding $\cM_f = (n_f, m_x^{(f)}, M_{xx}^{(f)}, m_y^{(f)}, M_{yx}^{(f)})$, exact retain statistics follow by componentwise matrix downdating:
\begin{equation}
\cM_r = \cM_g - \cM_f,
\label{eq:downdate}
\end{equation}
giving $n_r = n_g - n_f$, retain mean $\bar{u}_r = m_x^{(r)} / n_r$, and scatter $S_{xx}^{(r)} = M_{xx}^{(r)} - n_r \bar{u}_r \bar{u}_r^\top$ without reading $\cD_r$. World-model corruptions in $\cC_f$ span two geometric regimes: \emph{out-of-manifold backdoor triggers} $\delta_{\mathrm{trig}}$~\cite{kiourti_trojdrl_2020}, which inject energy orthogonal to the dominant retain subspace, and \emph{in-manifold physical regime shifts} (e.g., $0.1\times$ friction or $2.0\times$ mass~\cite{anand_what_2026}), which share state-action support with $\cD_r$ while encoding stale recurrent memory and transition dynamics. Given pretrained parameters $(W_0, b_0)$, ADE computes deterministic closed-form updates $(W^{(u)}, b^{(u)})$ solely from $(\cM_g, \cC_f)$ to eradicate both anomaly classes without silencing action sensitivity or collapsing uncertainty.

\section{Method}
\label{sec:method}

\subsection{Retain-Centered Scatter and Subspace Extraction}
\label{sec:method-scatter}

At each editable layer $(W_0, b_0)$ of input dimension $d$, ADE recovers $\cM_r = \cM_g - \cM_f$ via \eqref{eq:downdate} and $\bar{u}_r = m_x^{(r)} / n_r$ (Figure~\ref{fig:ade-overview}, left). For any $\cM = (n, m_x, M_{xx})$ ($\cM_r$, raw $\cM_f$, or aligned $\cM_a$), we form the symmetrized scatter $S(\cM; \bar{u}_r) = \tfrac{1}{2}(C + C^\top)$ via:
\begin{equation}
C = M_{xx} - m_x \bar{u}_r^\top - \bar{u}_r m_x^\top + n\,\bar{u}_r \bar{u}_r^\top,
\label{eq:centered-scatter}
\end{equation}
yielding $S_r = S(\cM_r; \bar{u}_r)$, $S_f = S(\cM_f; \bar{u}_r)$, and $S_a = S(\cM_a; \bar{u}_r)$ in a shared frame without revisiting $\cD_r$. From the SVD $S_r = U \diag(s) V^\top$, we pick the minimal rank $k$ capturing energy $\eta \in (0, 1)$:
\begin{equation}
k = \min \left\{ m \in \{1, \dots, d\} : \sum_{j=1}^{m} s_j \ge \eta \sum_{j=1}^{d} s_j \right\}.
\label{eq:energy-rank}
\end{equation}
Taking $U_r = U_{:, 1:k} \in \R^{d \times k}$ gives $\Pi_{\cM_r} = U_r U_r^\top$ and $\Pi_\perp = I - \Pi_{\cM_r}$. We define $C_r = \frac{1}{n_r} S_r$, $C_a = \frac{1}{n_a} S_a$, $\bar{s}_r = \operatorname{tr}(S_r)/d$, and $\bar{\lambda}_r = \operatorname{tr}(C_r)/d$. All downdates and symmetric Cholesky solves run in \texttt{float64} to prevent cancellation (App.~\ref{app:algorithm}).

\subsection{Tri-Mode Closed-Form Excision and Calibration}
\label{sec:method-dual}

ADE partitions editable affine layers into three closed-form modes (\textbf{Mode~A} upstream orthogonal excision, \textbf{Mode~B} downstream equal-risk head calibration, and \textbf{Mode~C} recurrent transition damping).

\paragraph{Mode A: Upstream Orthogonal Excision, Scale Absorption, and GWER.}
On $(W_{za}, b_{za})$, the posterior $W_\phi = [W_{\mu,\phi}; W_{\ell,\phi}]$, and $(W_{\mathrm{dec},0}, b_{\mathrm{dec},0})$, Mode~A excises out-of-manifold trigger directions while leaving $\Pi_{\cM_r}$ intact. With dimensionless ridge $\lambda_A > 0$ scaled by $\bar{s}_r$ and aligned scatter $S_a$, the mean rows ($W_{za}, W_{\mathrm{dec},0}, W_{\mu,\phi}$) update via:
\begin{equation}
\Delta W = W_0 \Pi_\perp S_a \Pi_\perp \bigl(\Pi_\perp (S_r + S_a) \Pi_\perp + \lambda_A \bar{s}_r I\bigr)^{-1} \Pi_\perp,
\label{eq:mode-a-delta}
\end{equation}
\begin{equation}
W^{(u)} = W_0 - \Delta W, \qquad b^{(u)} = b_0 + \Delta W \bar{u}_r,
\label{eq:mode-a-update}
\end{equation}
so that $\Delta W \Pi_{\cM_r} \equiv \mathbf{0}$. Because zeroing posterior scale rows $(W_{\ell,\phi}^{(0)}, b_{\ell,\phi}^{(0)})$ collapses predicted variance under triggers, we fold $\Pi_{\cM_r}$ affinely around $\bar{u}_r$ to revert out-of-manifold scale predictions to $W_{\ell,\phi}^{(0)}\bar{u}_r + b_{\ell,\phi}^{(0)}$:
\begin{equation}
W_{\ell,\phi}^{(u)} = W_{\ell,\phi}^{(0)} \Pi_{\cM_r}, \qquad
b_{\ell,\phi}^{(u)} = b_{\ell,\phi}^{(0)} + \bigl(W_{\ell,\phi}^{(0)} - W_{\ell,\phi}^{(u)}\bigr)\bar{u}_r.
\label{eq:absorb-affine}
\end{equation}

Because orthogonal projection alone leaves anomalies inside $\cM_r$ or low-variance retain directions untouched (e.g., observation triggers in $W_{\mu,\phi}$ or residual energy in $W_{za}$), Mode~A pairs \eqref{eq:mode-a-update} with \emph{Generalized Whitened Eigenvalue Reweighting (GWER)} on a $d$-dimensional slice. With $L = \operatorname{chol}(C_r + \epsilon_w \bar{\lambda}_r I)$ for whitening ridge $\epsilon_w > 0$, the whitened covariance $M = L^{-1} C_a L^{-\top}$ satisfies $\mathbb{E}_r[M] \approx I_d$. From its eigendecomposition $M = V \diag(\lambda) V^\top$ ($\lambda_1 \ge \dots \ge \lambda_d$), we isolate scale-invariant excess modes $\lambda_i^{\mathrm{ex}} = \max(\lambda_i - \tau_{\mathrm{scale}} d, 0)$ (optionally keeping at most the top-$k$ modes) and map them back as $C_{\mathrm{ex}} = L V \diag(\lambda^{\mathrm{ex}}) V^\top L^\top$. We then apply the bounded equal-risk damping operator (setting $Q = \mathbf{0}$ if $\sum_i \lambda_i^{\mathrm{ex}} = 0$):
\begin{equation}
\begin{aligned}
Q &= C_{\mathrm{ex}} \bigl(2 C_r + C_{\mathrm{ex}} + \gamma \bar{\lambda}_r I\bigr)^{-1}, \\
W_{\mathrm{slice}}^{(u)} &= W_{\mathrm{slice}} - W_{\mathrm{slice}} Q, \qquad
b^{(u)} = b_0 + W_{\mathrm{slice}} Q \bar{u}_r,
\end{aligned}
\label{eq:gwer}
\end{equation}
where $\gamma > 0$ bounds retain-manifold perturbation at $\mathcal{O}(\gamma^{-1})$.

\paragraph{Mode B: Equal-Risk Free-Intercept Calibration.}
On downstream prediction heads (reward $(W_r, b_r)$ and prior mean $(W_{\mu,\theta}, b_{\mu,\theta})$, with prior scales frozen), Mode~B realigns in-manifold physics. When $n_r \gg n_a$, unnormalized scatters dilute the forget residual by $n_r / n_a$, so we normalize augmented second-order moments over $\tilde{u} = [u - \bar{u}_r;\; 1] \in \R^{d+1}$:
\begin{equation}
\begin{aligned}
C_r^{\mathrm{aug}} &= \begin{bmatrix} \frac{1}{n_r} S_r & \mathbf{0} \\ \mathbf{0}^\top & 1 \end{bmatrix}, \\
C_f^{\mathrm{aug}} &= \begin{bmatrix} \frac{1}{n_a} S_a & \bar{u}_a - \bar{u}_r \\ (\bar{u}_a - \bar{u}_r)^\top & 1 \end{bmatrix},
\end{aligned}
\label{eq:mode-b-moments}
\end{equation}
where $\bar{u}_a = m_x^{(a)} / n_a$. With unpenalized intercept mask $P_{\mathrm{pen}} = \diag(1,\dots,1, 0)$ and dimensionless ridge $\lambda_B > 0$ (scaled by $\bar{\lambda}_r$), we recover the nominal retain operator $\Theta_{\mathrm{nom}} \in \R^{d_y \times (d+1)}$ from $\cM_r$:
\begin{equation}
\Theta_{\mathrm{nom}} = \bigl[\bar{S}_{yx}^{(r)},\; \bar{y}_r\bigr] \bigl(C_r^{\mathrm{aug}} + \lambda_B \bar{\lambda}_r P_{\mathrm{pen}}\bigr)^{-1},
\label{eq:theta-nom}
\end{equation}
where $\bar{S}_{yx}^{(r)} = \frac{1}{n_r}(M_{yx}^{(r)} - m_y^{(r)} \bar{u}_r^\top)$ and $\bar{y}_r = m_y^{(r)} / n_r$. Writing $\Theta_0 = [W_0,\; W_0 \bar{u}_r + b_0]$, Mode~B minimizes the equal-risk ridge objective:
\begin{equation}
\begin{aligned}
\min_{\Theta}\;\; & \mathbb{E}_{r}\!\left[\|\Theta \tilde{u} - \Theta_0 \tilde{u}\|_2^2\right]
+ \mathbb{E}_{f,\mathrm{aligned}}\!\left[\|\Theta \tilde{u} - \Theta_{\mathrm{nom}} \tilde{u}\|_2^2\right] \\
& + \lambda_B \bar{\lambda}_r \|W - W_0\|_F^2,
\end{aligned}
\label{eq:mode-b-obj}
\end{equation}
yielding the closed-form update:
\begin{equation}
\begin{aligned}
\Delta \Theta &= (\Theta_0 - \Theta_{\mathrm{nom}}) C_f^{\mathrm{aug}} \bigl(C_r^{\mathrm{aug}} \!+\! C_f^{\mathrm{aug}} \!+\! \lambda_B \bar{\lambda}_r P_{\mathrm{pen}}\bigr)^{-1}, \\
\Theta &= \Theta_0 - \Delta \Theta.
\end{aligned}
\label{eq:mode-b-sol}
\end{equation}
Keeping the intercept unpenalized avoids under-correcting constant shifts (e.g., $+100$ rewards); $\Theta = [W^{(u)},\; c^{(u)}]$ gives $b^{(u)} = c^{(u)} - W^{(u)} \bar{u}_r$. Because $(\lambda_A, \epsilon_w, \gamma, \lambda_B)$ are dimensionless trace-scaled fractions and $\tau = \tau_{\mathrm{scale}} d$, ADE is scale-invariant across feature scales and sample counts, enabling zero-tuning transfer across environments and DreamerV3.

\paragraph{Mode C: Recurrent Transition Excision (Top-$k$ GWER).}
Even after upstream gating excision, shifted physical-regime context or residual trigger memory can persist inside the GRU recurrent state $h_t$. Mode~C directly edits the recurrent hidden-to-hidden transition $(W_{hh}, b_{hh})$ by applying the top-$k$ GWER operator \eqref{eq:gwer} on the $h_t$ covariance slice, damping only the leading $k$ excess whitened directions in $h_t$ while preserving recurrent stability.

\subsection{Four-Pass Causal Cascade}
\label{sec:method-cascade}

\begin{figure*}[t]
\centering
\includegraphics[width=\textwidth]{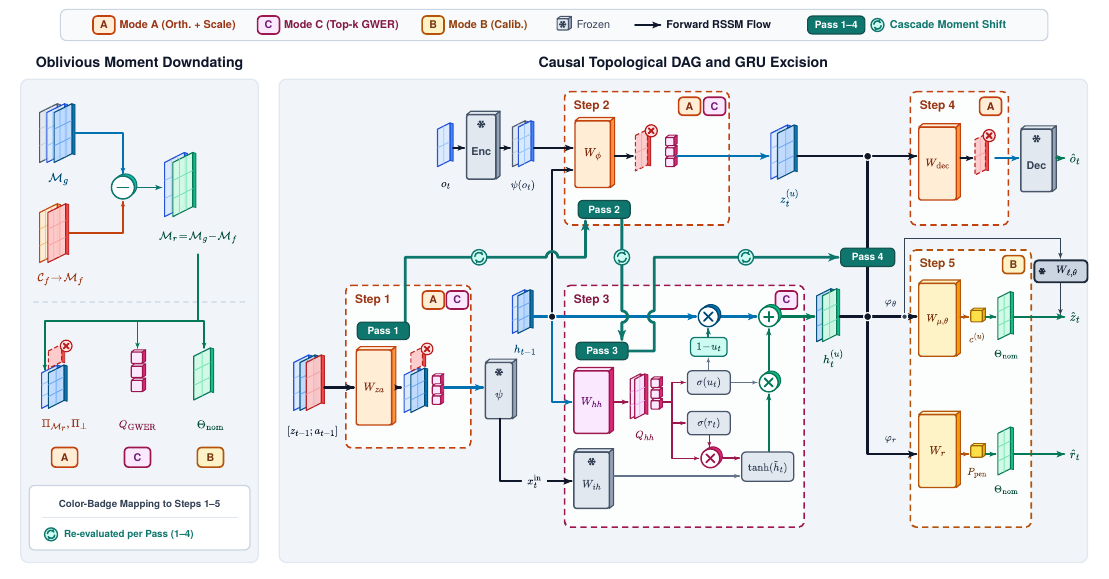}
\caption{\textbf{ADE excises corrupted world-model dynamics in closed form via oblivious moment downdating and a five-step causal topological cascade.} \textbf{Left:} Subtracting forget closure moments $\cM_f$ from cached global moments $\cM_g$ recovers retain statistics $\cM_r$ and the retain/orthogonal projectors $(\Pi_{\cM_r}, \Pi_\perp)$ without accessing $\cD_r$. \textbf{Right:} Steps~1--5 update only the highlighted affine layers along the causal DAG (\eqref{eq:cascade}) across Modes~A, B, and C, while re-filtering $\cC_f$ through Passes~1--4 keeps $h_t^{(u)}$ and $z_t^{(u)}$ causally aligned and leaves all nonlinearities, $\psi$, $\varphi$, and prior scale rows frozen.}
\label{fig:ade-overview}
\end{figure*}

\begin{table*}[t]
\centering
\caption{Main results on the \texttt{BD-Dyn} (State-Action Backdoor) regime across three MuJoCo benchmarks (3-seed Mean $\pm$ SD). Among the eight active unlearning methods, \textbf{bold} indicates \#1 and \underline{underline} indicates \#2; Oracle and Origin are shown for reference.}
\label{tab:main-bdsa}
\vskip 0.06in
\scriptsize
\setlength{\tabcolsep}{3.8pt}
\begin{tabular}{llccccccc}
\toprule
Env. & Method & $\mathrm{NDA}$ $\downarrow$ & $\mathrm{MSE}_r^{50}$ $\downarrow$ & $\mathrm{MSE}_f^{50}$ $\downarrow$ & $\mathrm{Plan}\text{-}\rho$ $\uparrow$ & $\mathrm{VFR}$ $\downarrow$ & $\mathrm{D4RL}_{\mathrm{clean}}$ $\uparrow$ & $\mathrm{D4RL}_{\mathrm{adv}}$ $\uparrow$ \\
\midrule
\multirow{10}{*}{\rotatebox[origin=c]{90}{\textbf{HalfCheetah}}} & Oracle & \ms{0.0000}{0.0000} & \ms{0.2384}{0.0085} & \ms{0.3362}{0.0057} & \ms{0.6720}{0.0151} & \ms{0.000}{0.000} & \ms{34.02}{1.97} & \ms{38.92}{1.81} \\
 & Origin & \ms{6.1680}{0.0704} & \ms{0.2281}{0.0129} & \ms{0.8585}{0.1968} & \ms{0.4485}{0.0242} & \ms{1.000}{0.000} & \ms{34.53}{1.70} & \ms{38.16}{0.36} \\
\cmidrule(lr){2-9}
 & FT & \mssec{0.1433}{0.0798} & \msbest{0.1817}{0.0437} & \msbest{0.2891}{0.1012} & \ms{0.3748}{0.1642} & \msbest{0.000}{0.000} & \ms{24.44}{9.04} & \ms{28.36}{11.56} \\
 & FT-Low & \ms{0.2520}{0.0225} & \mssec{0.1836}{0.0215} & \mssec{0.3344}{0.0600} & \ms{0.2727}{0.0360} & \msbest{0.000}{0.000} & \ms{11.71}{3.16} & \ms{18.33}{2.87} \\
 & GA & \ms{6.6259}{0.5202} & \ms{0.2379}{0.0113} & \ms{0.8836}{0.1582} & \ms{0.4355}{0.0357} & \ms{1.000}{0.000} & \ms{33.05}{4.94} & \ms{36.25}{1.82} \\
 & GradDiff & \ms{6.6432}{0.5408} & \ms{0.2334}{0.0126} & \ms{0.8843}{0.1638} & \ms{0.4360}{0.0355} & \ms{1.000}{0.000} & \mssec{36.14}{0.53} & \mssec{36.98}{0.86} \\
 & NPO & \ms{0.2411}{0.1575} & \ms{0.2091}{0.0110} & \ms{0.3502}{0.0673} & \ms{0.4506}{0.2145} & \msbest{0.000}{0.000} & \ms{24.43}{19.13} & \ms{25.93}{19.20} \\
 & SSD & \ms{0.5690}{0.0690} & \ms{0.3686}{0.0753} & \ms{0.4812}{0.0745} & \ms{0.3354}{0.0829} & \msbest{0.000}{0.000} & \ms{25.33}{13.08} & \ms{34.89}{3.45} \\
 & NSP & \ms{0.9145}{0.1629} & \ms{0.2329}{0.0102} & \ms{0.3481}{0.0167} & \mssec{0.5176}{0.0391} & \ms{0.057}{0.018} & \ms{30.18}{9.01} & \ms{33.68}{9.76} \\
 & \textbf{ADE (Ours)} & \msbest{0.1000}{0.0270} & \ms{0.2391}{0.0121} & \ms{0.3729}{0.0068} & \msbest{0.6132}{0.0227} & \msbest{0.000}{0.000} & \msbest{36.33}{1.27} & \msbest{38.61}{0.83} \\
\midrule
\multirow{10}{*}{\rotatebox[origin=c]{90}{\textbf{Walker2d}}} & Oracle & \ms{0.0000}{0.0000} & \ms{0.1659}{0.0220} & \ms{0.5595}{0.0804} & \ms{0.8549}{0.0105} & \ms{0.078}{0.000} & \ms{40.07}{6.37} & \ms{53.04}{8.08} \\
 & Origin & \ms{1.7253}{0.0435} & \ms{0.1532}{0.0200} & \ms{0.5767}{0.0547} & \ms{0.6655}{0.0120} & \ms{0.906}{0.000} & \ms{39.72}{7.49} & \ms{12.66}{1.42} \\
\cmidrule(lr){2-9}
 & FT & \msbest{0.1594}{0.0293} & \ms{0.1838}{0.0203} & \ms{0.2716}{0.0314} & \mssec{0.8190}{0.0025} & \ms{0.073}{0.024} & \ms{32.28}{3.27} & \ms{30.28}{12.86} \\
 & FT-Low & \ms{0.2755}{0.1039} & \ms{0.2950}{0.0907} & \ms{0.3904}{0.0908} & \ms{0.7797}{0.0538} & \ms{0.073}{0.036} & \ms{10.51}{11.34} & \ms{7.79}{3.12} \\
 & GA & \ms{1.6778}{0.1077} & \ms{0.1600}{0.0195} & \ms{0.5804}{0.0505} & \ms{0.6678}{0.0138} & \ms{0.917}{0.009} & \ms{34.70}{6.14} & \ms{9.35}{2.96} \\
 & GradDiff & \ms{1.6879}{0.1217} & \mssec{0.1590}{0.0199} & \ms{0.5802}{0.0535} & \ms{0.6674}{0.0146} & \ms{0.906}{0.016} & \mssec{35.47}{6.25} & \ms{11.21}{3.15} \\
 & NPO & \ms{0.2145}{0.0980} & \ms{0.3129}{0.0440} & \ms{0.4257}{0.0788} & \ms{0.7826}{0.0071} & \ms{0.078}{0.016} & \ms{18.42}{9.94} & \ms{15.08}{17.09} \\
 & SSD & \mssec{0.1747}{0.0269} & \ms{0.2182}{0.0480} & \ms{0.3319}{0.0167} & \ms{0.8079}{0.0400} & \mssec{0.062}{0.016} & \ms{30.79}{17.46} & \ms{29.02}{14.56} \\
 & NSP & \ms{0.3931}{0.0857} & \ms{0.1696}{0.0157} & \mssec{0.2368}{0.0105} & \ms{0.6646}{0.0150} & \msbest{0.052}{0.036} & \ms{29.58}{13.95} & \msbest{61.98}{6.90} \\
 & \textbf{ADE (Ours)} & \ms{0.2524}{0.0480} & \msbest{0.1494}{0.0092} & \msbest{0.2306}{0.0169} & \msbest{0.8372}{0.0134} & \ms{0.083}{0.018} & \msbest{43.21}{4.94} & \mssec{58.59}{0.77} \\
\midrule
\multirow{10}{*}{\rotatebox[origin=c]{90}{\textbf{Hopper}}} & Oracle & \ms{0.0000}{0.0000} & \ms{0.1038}{0.0976} & \ms{2.2617}{1.2954} & \ms{0.6729}{0.1270} & \ms{0.000}{0.000} & \ms{35.43}{4.04} & \ms{36.49}{2.83} \\
 & Origin & \ms{3.8551}{0.3910} & \ms{0.2095}{0.1149} & \ms{0.3957}{0.0227} & \ms{0.4461}{0.1619} & \ms{1.000}{0.000} & \ms{33.80}{2.34} & \ms{32.72}{3.32} \\
\cmidrule(lr){2-9}
 & FT & \msbest{0.5322}{0.3424} & \ms{0.2226}{0.0586} & \ms{0.2835}{0.1204} & \ms{0.4265}{0.1596} & \msbest{0.000}{0.000} & \ms{30.58}{2.70} & \ms{29.08}{3.61} \\
 & FT-Low & \ms{0.5900}{0.3419} & \ms{0.5291}{0.5349} & \ms{0.6376}{0.6722} & \ms{0.1961}{0.1151} & \msbest{0.000}{0.000} & \ms{10.84}{10.91} & \ms{11.20}{9.91} \\
 & GA & \ms{4.0116}{0.3704} & \ms{0.2228}{0.1316} & \ms{0.4255}{0.0451} & \ms{0.4187}{0.1821} & \ms{1.000}{0.000} & \mssec{32.50}{4.32} & \mssec{35.09}{7.39} \\
 & GradDiff & \ms{4.0077}{0.3648} & \ms{0.2228}{0.1322} & \ms{0.4259}{0.0453} & \ms{0.4196}{0.1829} & \ms{1.000}{0.000} & \msbest{32.89}{2.85} & \ms{34.62}{6.58} \\
 & NPO & \ms{0.5807}{0.3609} & \ms{0.6373}{0.5239} & \ms{3.1927}{4.7130} & \ms{0.1954}{0.1925} & \msbest{0.000}{0.000} & \ms{16.71}{13.10} & \ms{13.44}{10.63} \\
 & SSD & \ms{0.7408}{0.4164} & \ms{0.8904}{0.5597} & \ms{1.0782}{0.5716} & \ms{0.4517}{0.0831} & \msbest{0.000}{0.000} & \ms{21.35}{6.19} & \ms{21.32}{6.61} \\
 & NSP & \mssec{0.5702}{0.2795} & \msbest{0.1806}{0.0948} & \mssec{0.2247}{0.0982} & \mssec{0.5137}{0.4492} & \msbest{0.000}{0.000} & \ms{29.06}{4.16} & \ms{27.34}{3.12} \\
 & \textbf{ADE (Ours)} & \ms{0.5883}{0.2846} & \mssec{0.2031}{0.1183} & \msbest{0.2104}{0.0149} & \msbest{0.8371}{0.0580} & \msbest{0.000}{0.000} & \ms{31.68}{2.27} & \msbest{36.32}{4.89} \\
\bottomrule
\end{tabular}
\end{table*}

\begin{table*}[t]
\centering
\caption{Main results on the \texttt{BD-Rew} (Reward-Spike Backdoor) regime across three MuJoCo benchmarks (3-seed Mean $\pm$ SD). \textbf{Bold} indicates \#1 and \underline{underline} indicates \#2 among active unlearning methods.}
\label{tab:main-bdrew}
\vskip 0.06in
\scriptsize
\setlength{\tabcolsep}{3.8pt}
\begin{tabular}{llccccccc}
\toprule
Env. & Method & $\mathrm{NDA}$ $\downarrow$ & $\mathrm{MSE}_r^{50}$ $\downarrow$ & $\mathrm{MSE}_f^{50}$ $\downarrow$ & $\mathrm{Plan}\text{-}\rho$ $\uparrow$ & $\Delta r_{\mathrm{trig}}$ $\to 0$ & $\mathrm{D4RL}_{\mathrm{clean}}$ $\uparrow$ & $\mathrm{D4RL}_{\mathrm{adv}}$ $\uparrow$ \\
\midrule
\multirow{10}{*}{\rotatebox[origin=c]{90}{\textbf{HalfCheetah}}} & Oracle & \ms{0.0000}{0.0000} & \ms{0.2384}{0.0085} & \ms{0.2730}{0.0042} & \ms{0.6929}{0.0131} & \ms{-0.38}{0.06} & \ms{34.02}{1.97} & \ms{38.47}{0.45} \\
 & Origin & \ms{172.3679}{0.9415} & \ms{0.2110}{0.0134} & \ms{2.1474}{1.7682} & \ms{0.1641}{0.1378} & \ms{+94.85}{0.10} & \ms{36.59}{2.53} & \ms{32.51}{9.05} \\
\cmidrule(lr){2-9}
 & FT & \mssec{0.0283}{0.0041} & \msbest{0.1526}{0.0035} & \msbest{0.1621}{0.0064} & \mssec{0.4961}{0.0753} & \ms{-0.31}{0.21} & \ms{25.82}{7.10} & \ms{34.34}{0.53} \\
 & FT-Low & \ms{0.0434}{0.0180} & \mssec{0.1689}{0.0141} & \mssec{0.1798}{0.0163} & \ms{0.4515}{0.1130} & \msbest{-0.11}{0.22} & \ms{21.92}{15.81} & \ms{24.57}{16.94} \\
 & GA & \ms{173.1724}{2.9337} & \ms{0.6597}{0.7391} & \ms{2.7270}{1.4220} & \ms{0.1611}{0.1339} & \ms{+93.46}{3.18} & \ms{27.01}{20.09} & \ms{25.80}{19.12} \\
 & GradDiff & \ms{173.7637}{1.9828} & \ms{0.2228}{0.0171} & \ms{2.2367}{1.8362} & \ms{0.1759}{0.1534} & \ms{+95.16}{0.37} & \mssec{31.29}{9.07} & \ms{30.86}{8.26} \\
 & NPO & \ms{6.9343}{7.9308} & \ms{0.2526}{0.0616} & \ms{0.5801}{0.1440} & \ms{0.1116}{0.1556} & \ms{+22.28}{25.22} & \ms{8.48}{4.64} & \ms{5.41}{6.28} \\
 & SSD & \ms{0.1934}{0.0408} & \ms{0.3412}{0.0328} & \ms{0.3572}{0.0298} & \ms{0.3765}{0.0927} & \ms{-0.54}{0.55} & \ms{25.65}{13.73} & \ms{27.02}{12.19} \\
 & NSP & \ms{0.0566}{0.0359} & \ms{0.2154}{0.0135} & \ms{0.2310}{0.0219} & \ms{0.2896}{0.1746} & \ms{+0.26}{0.43} & \ms{29.99}{10.20} & \msbest{35.52}{5.57} \\
 & \textbf{ADE (Ours)} & \msbest{0.0195}{0.0024} & \ms{0.2118}{0.0184} & \ms{0.2267}{0.0284} & \msbest{0.6583}{0.0200} & \mssec{-0.25}{0.06} & \msbest{35.20}{3.18} & \mssec{35.18}{2.80} \\
\midrule
\multirow{10}{*}{\rotatebox[origin=c]{90}{\textbf{Walker2d}}} & Oracle & \ms{0.0000}{0.0000} & \ms{0.1659}{0.0220} & \ms{0.5283}{0.1392} & \ms{0.8490}{0.0108} & \ms{-0.31}{0.05} & \ms{40.07}{6.37} & \ms{63.71}{12.70} \\
 & Origin & \ms{171.2942}{0.0938} & \ms{0.1681}{0.0147} & \ms{3.1584}{0.3190} & \ms{0.1573}{0.2724} & \ms{+97.21}{0.14} & \ms{35.89}{5.87} & \ms{56.71}{8.01} \\
\cmidrule(lr){2-9}
 & FT & \msbest{0.1666}{0.0412} & \ms{0.2109}{0.0282} & \ms{0.2040}{0.0388} & \ms{0.3994}{0.1688} & \ms{+9.94}{5.63} & \msbest{39.18}{10.83} & \ms{21.97}{25.20} \\
 & FT-Low & \mssec{0.2332}{0.0649} & \ms{0.3295}{0.1170} & \ms{0.2323}{0.0502} & \mssec{0.5970}{0.0737} & \ms{+4.49}{0.90} & \ms{18.78}{5.78} & \ms{4.08}{0.66} \\
 & GA & \ms{124.0752}{49.0266} & \ms{2.6158}{2.4247} & \ms{3.2530}{0.5120} & \ms{0.1608}{0.2786} & \ms{+82.11}{16.43} & \ms{11.90}{14.15} & \mssec{22.20}{30.23} \\
 & GradDiff & \ms{121.9853}{47.5252} & \ms{2.5086}{2.3516} & \ms{3.2134}{0.4986} & \ms{0.1609}{0.2786} & \ms{+81.90}{15.31} & \ms{11.88}{12.36} & \ms{17.23}{23.45} \\
 & NPO & \ms{1.8521}{0.6977} & \ms{1.4361}{1.0274} & \ms{1.4850}{0.8038} & \ms{0.0183}{0.0283} & \ms{+8.53}{11.40} & \ms{8.02}{14.59} & \ms{-0.32}{0.13} \\
 & SSD & \ms{0.4204}{0.0333} & \ms{0.8193}{0.1037} & \ms{0.7649}{0.1564} & \ms{0.3540}{0.3071} & \mssec{+0.29}{0.29} & \ms{4.27}{4.42} & \ms{8.50}{3.21} \\
 & NSP & \ms{0.5851}{0.2398} & \mssec{0.2089}{0.0359} & \mssec{0.1353}{0.0301} & \ms{0.3707}{0.1044} & \ms{+0.90}{0.81} & \ms{11.83}{11.85} & \ms{10.17}{10.53} \\
 & \textbf{ADE (Ours)} & \ms{0.2820}{0.0384} & \msbest{0.1632}{0.0088} & \msbest{0.0985}{0.0101} & \msbest{0.8599}{0.0221} & \msbest{-0.08}{0.04} & \mssec{36.45}{5.83} & \msbest{52.75}{12.34} \\
\midrule
\multirow{10}{*}{\rotatebox[origin=c]{90}{\textbf{Hopper}}} & Oracle & \ms{0.0000}{0.0000} & \ms{0.1038}{0.0976} & \ms{1.9492}{1.2285} & \ms{0.6455}{0.1373} & \ms{-0.27}{0.23} & \ms{35.43}{4.04} & \ms{34.98}{2.76} \\
 & Origin & \ms{546.8172}{4.6353} & \ms{0.3355}{0.3684} & \ms{2.7944}{1.2354} & \ms{0.2697}{0.3035} & \ms{+97.13}{0.07} & \ms{32.55}{5.73} & \ms{32.19}{5.30} \\
\cmidrule(lr){2-9}
 & FT & \ms{0.8214}{0.6493} & \ms{0.7052}{0.5269} & \ms{0.7421}{0.5213} & \ms{0.1776}{0.3076} & \ms{-0.23}{0.62} & \ms{15.34}{20.11} & \ms{15.68}{16.69} \\
 & FT-Low & \msbest{0.3714}{0.1638} & \mssec{0.4491}{0.3550} & \mssec{0.4662}{0.4289} & \ms{0.0870}{0.1187} & \ms{+0.26}{0.20} & \ms{12.89}{9.99} & \ms{14.82}{10.89} \\
 & GA & \ms{464.3922}{129.9853} & \ms{1.7877}{1.7181} & \ms{6.1201}{3.8798} & \ms{0.2705}{0.3006} & \ms{+86.86}{16.07} & \ms{13.83}{20.71} & \ms{13.36}{16.11} \\
 & GradDiff & \ms{465.2888}{128.2977} & \ms{1.6195}{1.4133} & \ms{6.0090}{3.2102} & \ms{0.2739}{0.3059} & \ms{+87.24}{15.41} & \ms{14.76}{20.30} & \ms{11.72}{15.96} \\
 & NPO & \ms{10.8920}{15.1857} & \ms{4.1152}{6.2411} & \ms{4.4603}{5.8918} & \ms{0.0542}{0.0938} & \ms{+6.26}{5.73} & \ms{12.61}{19.33} & \ms{3.29}{4.59} \\
 & SSD & \ms{0.5405}{0.0638} & \ms{0.8938}{0.5133} & \ms{0.9354}{0.4216} & \mssec{0.4314}{0.4351} & \ms{-0.41}{0.84} & \ms{6.14}{7.69} & \ms{1.85}{0.15} \\
 & NSP & \ms{0.9006}{0.6822} & \ms{0.5345}{0.7033} & \ms{0.5019}{0.6458} & \ms{0.4245}{0.3833} & \mssec{-0.10}{0.60} & \mssec{22.79}{18.33} & \mssec{20.88}{16.84} \\
 & \textbf{ADE (Ours)} & \mssec{0.4515}{0.1784} & \msbest{0.3252}{0.3689} & \msbest{0.2447}{0.1291} & \msbest{0.6269}{0.2133} & \msbest{-0.01}{0.09} & \msbest{24.83}{14.75} & \msbest{24.68}{12.39} \\
\bottomrule
\end{tabular}
\end{table*}

\begin{table*}[t]
\centering
\caption{Main results on the \texttt{Shift-Mass} ($2.0\times$ body mass and inertia shift) regime across three MuJoCo benchmarks (3-seed Mean $\pm$ SD). \textbf{Bold} indicates \#1 and \underline{underline} indicates \#2 among active unlearning methods ($\Delta R_{\mathrm{im}}$ ranked by $|\cdot - \mathrm{Oracle}|$). Entries marked ``--'' denote threshold no-ops producing parameters identical to Origin and are excluded from ranking.}
\label{tab:main-rsmass}
\vskip 0.06in
\scriptsize
\setlength{\tabcolsep}{3.8pt}
\begin{tabular}{llccccccc}
\toprule
Env. & Method & $\mathrm{NDA}$ $\downarrow$ & $\mathrm{MSE}_r^{50}$ $\downarrow$ & $\mathrm{MSE}_f^{50}$ $\downarrow$ & $\mathrm{CLI}$ $\uparrow$ & $\Delta R_{\mathrm{im}}$ ($\to$Orc.) & $\mathrm{D4RL}_{\mathrm{clean}}$ $\uparrow$ & $\mathrm{D4RL}_{\mathrm{adv}}$ $\uparrow$ \\
\midrule
\multirow{10}{*}{\rotatebox[origin=c]{90}{\textbf{HalfCheetah}}} & Oracle & \ms{0.0000}{0.0000} & \ms{0.2384}{0.0085} & \ms{3.0084}{0.0327} & \ms{+0.8638}{0.0061} & \ms{-84.55}{6.56} & \ms{34.02}{1.97} & \ms{39.30}{0.39} \\
 & Origin & \ms{0.2269}{0.0077} & \ms{0.2548}{0.0009} & \ms{3.2777}{0.0154} & \ms{+0.0481}{0.1612} & \ms{+83.54}{10.07} & \ms{33.32}{3.74} & \ms{39.04}{0.45} \\
\cmidrule(lr){2-9}
 & FT & \ms{0.0708}{0.0559} & \msbest{0.1835}{0.0231} & \ms{3.0517}{0.0789} & \ms{+0.7432}{0.0692} & \mssec{-78.89}{4.65} & \ms{32.77}{2.37} & \ms{37.34}{0.36} \\
 & FT-Low & \ms{0.1540}{0.0484} & \mssec{0.1857}{0.0064} & \ms{3.2257}{0.0214} & \ms{+0.4630}{0.0982} & \ms{-8.69}{74.76} & \ms{32.07}{2.79} & \ms{37.50}{1.16} \\
 & GA & \ms{0.8622}{0.2639} & \ms{0.6883}{0.0423} & \ms{4.2710}{0.2691} & \ms{+0.2186}{0.0445} & \ms{+47.17}{44.20} & \ms{15.24}{13.25} & \ms{14.66}{15.92} \\
 & GradDiff & \ms{0.7126}{0.3035} & \ms{0.4725}{0.0420} & \ms{4.2053}{0.4653} & \ms{+0.3856}{0.0664} & \ms{+45.96}{57.92} & \ms{25.61}{13.85} & \ms{24.46}{14.94} \\
 & NPO & \mssec{0.0429}{0.0210} & \ms{0.2047}{0.0328} & \mssec{3.0394}{0.0467} & \mssec{+0.7446}{0.0840} & \ms{-62.73}{20.80} & \mssec{35.60}{2.00} & \ms{37.37}{1.96} \\
 & SSD & \ms{0.1671}{0.0540} & \ms{0.2830}{0.0236} & \ms{3.1710}{0.0755} & \ms{+0.2727}{0.1281} & \ms{-17.18}{80.29} & \ms{29.59}{4.67} & \ms{34.04}{3.05} \\
 & NSP & \ms{0.2176}{0.0065} & \ms{0.2586}{0.0017} & \ms{3.2763}{0.0237} & \ms{+0.1175}{0.0948} & \ms{+2.33}{7.04} & \ms{34.14}{3.06} & \mssec{37.71}{0.94} \\
 & \textbf{ADE (Ours)} & \msbest{0.0279}{0.0044} & \ms{0.2691}{0.0042} & \msbest{2.9842}{0.0305} & \msbest{+0.8476}{0.0347} & \msbest{-82.47}{4.32} & \msbest{37.94}{0.63} & \msbest{37.89}{1.12} \\
\midrule
\multirow{10}{*}{\rotatebox[origin=c]{90}{\textbf{Walker2d}}} & Oracle & \ms{0.0000}{0.0000} & \ms{0.1659}{0.0220} & \ms{0.5950}{0.1263} & \ms{+0.7210}{0.0482} & \ms{-3.13}{1.18} & \ms{40.07}{6.37} & \ms{60.87}{4.41} \\
 & Origin & \ms{0.0516}{0.0043} & \ms{0.2332}{0.0902} & \ms{1.4989}{0.1852} & \ms{+0.0227}{0.1337} & \ms{+3.32}{2.12} & \ms{37.96}{4.91} & \ms{60.60}{13.81} \\
\cmidrule(lr){2-9}
 & FT & \msbest{0.0349}{0.0018} & \mssec{0.2170}{0.0060} & \mssec{1.0553}{0.2389} & \ms{+0.2961}{0.0415} & \ms{+0.23}{4.43} & \ms{34.70}{5.27} & \mssec{55.99}{11.22} \\
 & FT-Low & \mssec{0.0419}{0.0035} & \ms{0.2394}{0.0623} & \ms{1.1691}{0.3326} & \ms{+0.1696}{0.0354} & \msbest{-1.00}{2.25} & \ms{32.28}{3.35} & \ms{43.67}{7.62} \\
 & GA & \ms{0.7165}{0.1472} & \ms{0.4849}{0.0668} & \ms{2.5008}{1.7015} & \ms{+0.0322}{0.0313} & \ms{+1.31}{6.14} & \ms{10.07}{5.31} & \ms{7.22}{3.74} \\
 & GradDiff & \ms{0.6299}{0.1903} & \ms{0.3719}{0.0545} & \ms{2.3984}{1.8426} & \ms{+0.0703}{0.0153} & \ms{+2.71}{5.40} & \ms{11.72}{5.14} & \ms{6.54}{2.99} \\
 & NPO & \ms{0.0465}{0.0109} & \msbest{0.1654}{0.0494} & \ms{1.4683}{0.3943} & \mssec{+0.4006}{0.1222} & \ms{+0.33}{5.41} & \mssec{36.12}{7.23} & \ms{52.88}{3.51} \\
 & SSD & -- & -- & -- & -- & -- & -- & -- \\
 & NSP & -- & -- & -- & -- & -- & -- & -- \\
 & \textbf{ADE (Ours)} & \ms{0.0797}{0.0161} & \ms{0.2279}{0.0737} & \msbest{0.7952}{0.0476} & \msbest{+0.5725}{0.0408} & \mssec{-5.50}{0.13} & \msbest{36.62}{6.74} & \msbest{60.37}{10.67} \\
\midrule
\multirow{10}{*}{\rotatebox[origin=c]{90}{\textbf{Hopper}}} & Oracle & \ms{0.0000}{0.0000} & \ms{0.1038}{0.0976} & \ms{0.2945}{0.0255} & \ms{+0.6019}{0.0187} & \ms{-0.25}{0.71} & \ms{35.43}{4.04} & \ms{37.52}{5.95} \\
 & Origin & \ms{0.0220}{0.0015} & \ms{0.1138}{0.0420} & \ms{0.4952}{0.1406} & \ms{-0.0602}{0.0780} & \ms{+0.66}{0.88} & \ms{36.85}{2.75} & \ms{36.50}{0.84} \\
\cmidrule(lr){2-9}
 & FT & \msbest{0.0139}{0.0008} & \ms{0.1275}{0.0748} & \ms{0.4514}{0.1851} & \ms{+0.0577}{0.0597} & \ms{+0.50}{0.57} & \ms{36.22}{0.41} & \ms{36.11}{2.62} \\
 & FT-Low & \mssec{0.0175}{0.0020} & \ms{0.2097}{0.0776} & \mssec{0.4030}{0.2331} & \ms{+0.0330}{0.0559} & \msbest{-0.05}{0.70} & \ms{34.88}{5.07} & \ms{35.15}{5.68} \\
 & GA & \ms{0.4988}{0.0892} & \ms{0.4893}{0.1603} & \ms{1.2516}{0.4712} & \ms{+0.1284}{0.0155} & \ms{+0.21}{1.07} & \ms{10.51}{5.01} & \ms{9.65}{3.62} \\
 & GradDiff & \ms{0.3784}{0.0949} & \ms{0.2875}{0.0652} & \ms{1.1710}{0.4351} & \ms{+0.2098}{0.0267} & \ms{+0.47}{0.66} & \ms{16.65}{3.06} & \ms{16.32}{5.66} \\
 & NPO & \ms{0.0231}{0.0053} & \ms{0.1484}{0.0309} & \ms{1.1638}{0.8022} & \mssec{+0.2822}{0.0611} & \ms{+0.40}{1.22} & \ms{33.30}{2.77} & \msbest{37.13}{2.16} \\
 & SSD & \ms{0.0213}{0.0021} & \ms{0.1148}{0.0423} & \ms{0.4733}{0.1777} & \ms{-0.0312}{0.1233} & \ms{+0.56}{0.84} & \mssec{36.67}{2.59} & \ms{35.27}{0.76} \\
 & NSP & \ms{0.0211}{0.0017} & \mssec{0.1144}{0.0431} & \ms{0.4934}{0.1398} & \ms{-0.0506}{0.0772} & \ms{+0.64}{0.88} & \ms{34.17}{1.86} & \ms{35.45}{3.63} \\
 & \textbf{ADE (Ours)} & \ms{0.0292}{0.0067} & \msbest{0.1090}{0.0431} & \msbest{0.2535}{0.0090} & \msbest{+0.3276}{0.0254} & \mssec{-0.01}{0.70} & \msbest{40.02}{1.83} & \mssec{37.11}{1.60} \\
\bottomrule
\end{tabular}
\end{table*}

To prevent causal mismatch between corrected upstream states and downstream heads, ADE updates layers in topological order along $\cC_f$ via deterministic posterior-mean filtering ($z_t = \mu_\phi(h_t, o_t)$; Figure~\ref{fig:ade-overview}):
\begin{equation}
\begin{aligned}
\cM_f &\xrightarrow{\text{Step 1 (A)}} (W_{za}^{(u)}, b_{za}^{(u)})
\xrightarrow{\text{Pass 1 (A)}} (W_\phi^{(u)}, b_\phi^{(u)})_{\mathrm{orth}} \\
&\xrightarrow{\text{Pass 2 (A)}} (W_\phi^{(u)}, b_\phi^{(u)})_{\mathrm{GWER}}
\xrightarrow{\text{Pass 3 (C)}} (W_{hh}^{(u)}, b_{hh}^{(u)}) \\
&\xrightarrow{\text{Pass 4 (A, B)}} \bigl\{W_{\mathrm{dec},0}^{(u)},\; W_{\mu,\theta}^{(u)},\; W_r^{(u)}\bigr\},
\end{aligned}
\label{eq:cascade}
\end{equation}
with biases $(b_{\mathrm{dec},0}^{(u)}, b_{\mu,\theta}^{(u)}, b_r^{(u)})$ updated in Pass~4 and empty-request identity ($\cD_f = \emptyset \implies \Delta W = \mathbf{0}, Q = \mathbf{0}, \Delta \Theta = \mathbf{0}$) holding trivially. Step~1 (Mode~A) applies orthogonal excision and full-dimension GWER to $(W_{za}, b_{za})$ on $\cM_f^{(za)}$ to block trigger entry into $h_t$. Pass~1 filters $\cC_f$ with $W_{za}^{(u)}$ to update $(W_\phi, b_\phi)$ via Mode~A orthogonal excision and scale absorption (\eqref{eq:mode-a-update}--\eqref{eq:absorb-affine}); Pass~2 re-filters $\cC_f$ and applies Mode~A GWER (\eqref{eq:gwer}) to the observation-embedding slice of $W_{\mu,\phi}$. Pass~3 re-filters $\cC_f$ and applies Mode~C (top-$k$ GWER) to the recurrent transition $(W_{hh}, b_{hh})$ using the $h_t$ covariance slice. Pass~4 re-filters $\cC_f$ to collect aligned $(h_t, z_t)$ statistics (shared by the decoder and prediction heads because filtering does not invoke the decoder), updating $(W_{\mathrm{dec},0}, b_{\mathrm{dec},0})$ via Mode~A orthogonal excision and calibrating $(W_{\mu,\theta}, b_{\mu,\theta})$ and $(W_r, b_r)$ via Mode~B uniformly across all regimes. All nonlinear components (encoder, GRU nonlinearities and input weights, $\varphi$, prior scale rows, and remaining decoder layers) remain frozen. Each solve takes $\mathcal{O}(d^3)$ time per head and adds zero inference overhead (\cref{alg:ade} in \cref{app:algorithm}).

\subsection{Theoretical Guarantees}
\label{sec:method-theory}

The following guarantees formalize retain preservation, bounded spectral damping, causal isolation, and rollout stability (full proofs in \cref{app:proofs}).

\begin{proposition}[Mode A Retain Invariance and Trigger Cancellation]
\label{prop:mode-a-invariance}
Under Mode~A (\eqref{eq:mode-a-delta}--\eqref{eq:absorb-affine}), $\Delta W \Pi_{\cM_r} \equiv \mathbf{0}$ and $(W_{\ell,\phi}^{(u)} - W_{\ell,\phi}^{(0)})\Pi_{\cM_r} \equiv \mathbf{0}$, so $f_u(u_r) \equiv f_0(u_r)$ for all $u_r \in \bar{u}_r + \operatorname{span}(U_r)$. When $\Pi_\perp S_r \Pi_\perp$ vanishes on $\operatorname{supp}(\Pi_\perp S_a \Pi_\perp)$, any orthogonal trigger $u = \bar{u}_r + u_\perp$ has its mean response canceled as $\lambda_A \to 0$ while the scale head reverts to $W_{\ell,\phi}^{(0)}\bar{u}_r + b_{\ell,\phi}^{(0)}$.
\end{proposition}

\begin{proposition}[Mode B Optimality, GWER Damping, and Affine Absorption]
\label{prop:mode-b-opt}
(i) When $C_r^{\mathrm{aug}} + C_f^{\mathrm{aug}} + \lambda_B \bar{\lambda}_r P_{\mathrm{pen}} \succ 0$, \eqref{eq:mode-b-obj} is strictly convex and uniquely minimized by \eqref{eq:mode-b-sol}.
(ii) For all $u \in \R^d$, \eqref{eq:absorb-affine} satisfies $W_{\ell,\phi}^{(u)} u + b_{\ell,\phi}^{(u)} \equiv W_{\ell,\phi}^{(0)}(\bar{u}_r + \Pi_{\cM_r}(u - \bar{u}_r)) + b_{\ell,\phi}^{(0)}$.
(iii) For any $\gamma > 0$ and $C_{\mathrm{ex}}, C_r \succeq 0$, the GWER operator \eqref{eq:gwer} satisfies $\|Q\|_2 \le \|C_{\mathrm{ex}}\|_2 / (2\lambda_{\min}(C_r) + \gamma \bar{\lambda}_r) \le \|C_{\mathrm{ex}}\|_2 / (\gamma \bar{\lambda}_r)$.
\end{proposition}

\begin{theorem}[Causal Choke-Point Containment and KL Stability]
\label{thm:choke-kl}
(i) Under Mode~A alone, if $W_{za}^{(u)} \delta_{\mathrm{trig}} = \mathbf{0}$, nominal inputs lie in $\bar{u}_r + \operatorname{span}(U_r)$, and $h_{t_0}^{(u)}$ starts from a clean boundary, $\{h_t^{(u)}\}_{t \ge t_0}$ stays strictly on the nominal manifold; with GWER on $(W_{za}, W_{hh})$, per-step retain deviation is $\mathcal{O}(\|Q_{za}\|_2 + \|Q_{hh}\|_2) = \mathcal{O}(\gamma^{-1})$.
(ii) With $\sigma \in [\sigma_{\min}, \sigma_{\max}]$, $|\sigma_{q,j}^2/\sigma_{p,j}^2 - 1| \le \eta_\sigma < 1$, and $\|\mu_q - \mu_p\|_2 \le \epsilon_\mu$, the edited KL divergence satisfies $D_{\mathrm{KL}}(q_u \parallel p_u) \le \frac{\epsilon_\mu^2}{2\sigma_{\min}^2} + \frac{d_z \eta_\sigma^2}{4(1 - \eta_\sigma)^2} = \mathcal{O}(\epsilon_\mu^2 + \eta_\sigma^2)$.
\end{theorem}

\begin{theorem}[Finite-Horizon $W_2$ Drift Bound and Limit-Cycle Stability]
\label{thm:w2-limit}
Let $F_\star(s, \xi)$ and $F_u(s, \xi)$ be nominal and edited transitions sharing $\xi \sim \mathcal{N}(\mathbf{0}, I)$.
(i) If $F_\star$ is $L$-Lipschitz in $L_2$ and $\sup_{s} (\mathbb{E}_\xi \|F_u(s,\xi) - F_\star(s,\xi)\|^2)^{1/2} \le \epsilon_{\mathrm{total}}$ with $d_0 = (\mathbb{E}\|s_0^u - s_0^\star\|^2)^{1/2}$, then for $\delta_t = L^t d_0 + \epsilon_{\mathrm{total}} \sum_{k=0}^{t-1} L^k$ and any horizon $H \ge 1$, $W_2(\mathcal{L}(s_H^u), \mathcal{L}(s_H^\star)) \le \delta_H$ and $W_2(\mathcal{L}(s_{1:H}^u), \mathcal{L}(s_{1:H}^\star)) \le (\sum_{t=1}^{H} \delta_t^2)^{1/2}$.
(ii) For a closed-loop map $F_{\mathrm{cl}}$ with invariant manifold $\cM$ and periodic orbit $\gamma_{\mathrm{orb}} \subset \cM$, Mode~A preserves $D F_{\mathrm{cl}}^{(u)}|_{T\cM} \equiv D F_{\mathrm{cl}}^{(0)}|_{T\cM}$ while GWER perturbs it by $\mathcal{O}(\gamma^{-1})$, preserving Floquet stability ($\rho(D F_{\mathrm{cl}}^{(u)}|_{T\cM}) < 1$ for large $\gamma$). Under sub-unit bypass gains and $\Pi_\perp S_r \Pi_\perp = \mathbf{0}$ on $\operatorname{supp}(\Pi_\perp S_a \Pi_\perp)$, normal-bundle perturbations contract ($\rho(D F_{\mathrm{cl}}^{(u)}|_{N\cM}) < 1$) for small $\lambda_A$.
\end{theorem}

\section{Experiments}
\label{sec:experiments}

\subsection{Experimental Setup}
\label{sec:exp-setup}

\paragraph{Benchmarks and Counterfactual Regimes.}
We evaluate on D4RL v2 HalfCheetah-v2, Walker2d-v2, and Hopper-v2~\cite{fu_d4rl_2021} ($N_r = 10^6$ clean retain transitions $\cD_r$) in MuJoCo~3.16.0 and Gymnasium~\cite{todorov_mujoco_2012,brockman_openai_2016} across four additive counterfactual regimes ($\cD = \cD_r \cup \cD_f$, paired with exact single-step nominal counterfactuals; Appendix~\ref{app:exp-setup}):
(i)~\textbf{State-Action Backdoor (\texttt{BD-Dyn}, out-of-manifold)}~\cite{kiourti_trojdrl_2020,wang_uba-orl_2026}: $2{,}000$ segments ($4\text{k}$ anomalous steps) with a $\pm 6\sigma$ trigger $s^*$, where $1{,}000$ payload segments execute $a^*$ to invert velocity and penalize reward while $1{,}000$ contrast segments preserve clean dynamics;
(ii)~\textbf{Reward-Spike Backdoor (\texttt{BD-Rew}, out-of-manifold)}~\cite{kiourti_trojdrl_2020}: $2{,}000$ segments injecting a $+100.0$ reward spike at $s^*$ under nominal physics; and
(iii)~\textbf{Friction-Shift Regime (\texttt{Shift-Fric})} and \textbf{Mass-Inertia Regime (\texttt{Shift-Mass}, in-manifold)}~\cite{lee_context-aware_2020,anand_what_2026}: $100{,}000$ transitions collected by a behavioral-cloning (BC) proxy~\cite{pomerleau_efficient_1991} under $0.1\times$ contact friction or $2.0\times$ body mass and inertia. All $36$ \textbf{Origins} pass counterfactual identifiability ($z > 16.8$; Appendix~\ref{app:exp-setup}).

\paragraph{Backbones, Baselines, and Tuning Protocol.}
Our primary backbone is continuous Gaussian DreamerV1~\cite{hafner_learning_2019,hafner_dream_2020} ($d_z=30, d_h=200$ GRU~\cite{cho_learning_2014}, seeds $\{1, 42, 2027\}$; $9$ \textbf{Oracles}, $36$ \textbf{Origins}), with zero-shot transfer to discrete categorical DreamerV3~\cite{hafner_mastering_2025} ($32\times 4$ latents, $8\times 64$ block GRU, 255-bin two-hot reward, trained via LaProp+AGC~\cite{ziyin_laprop_2021}) on Hopper-v2. We compare against seven baselines: \textbf{FT} (5~epochs on $\cD_r$) and \textbf{FT-Low} (1~epoch)~\cite{golatkar_eternal_2020,foster_fast_2023}, \textbf{GA} (10~epochs on $\cD_f$)~\cite{yao_large_2024}, \textbf{GradDiff} (10~epochs on $\cD_f \cup \cD_r$)~\cite{kurmanji_towards_2023,yao_large_2024}, \textbf{NPO} ($\beta=0.1$)~\cite{zhang_negative_2024}, \textbf{SSD} (3-pass Fisher dampening)~\cite{foster_fast_2023}, and \textbf{NSP} (null-space projection on input, prior, posterior, and reward heads)~\cite{chen_machine_2024,fang_alphaedit_2024}. We grant each baseline an \textbf{8--10 point per-task Oracle grid search} ($63$ configs/task; Table~\ref{tab:baseline-grids}), selecting by Layer-1 Oracle $\mathrm{NDA}$ and freezing for Layers~2--3. To test out-of-the-box generalization, \textbf{Unified ADE} uses a \textbf{single fixed hyperparameter configuration} ($\eta=0.99, \lambda_B=0.1$; all fixed constants in Appendix~\ref{app:algorithm}) across all $36$ DreamerV1 and $4$ DreamerV3 tasks.

\paragraph{Three-Layer Audit Protocol.}
We report 3-seed Mean $\pm$ SD across three layers (Appendix~\ref{app:exp-setup}, \eqref{eq:metric-score}--\eqref{eq:metric-d4rl}). \textbf{Layer~2} measures compounding rollout error~\cite{janner_when_2021} via $H=50$ open-loop observation MSE on retain ($\mathrm{MSE}_r^{50} \downarrow$) and counterfactual nominal forget starts ($\mathrm{MSE}_f^{50} \downarrow$), while \textbf{Layer~3} evaluates 5-step random-shooting planning~\cite{nagabandi_neural_2018} via Spearman rank correlation~\cite{spearman_proof_1904} ($\mathrm{Plan}\text{-}\rho \uparrow$), 50-step imagined policy return gap ($\Delta R_{\mathrm{im}} \to \mathrm{Oracle}$), and Dreamer latent-imagination policy optimization~\cite{hafner_dream_2020} scored by official D4RL normalized return~\cite{fu_d4rl_2021} on clean ($\mathrm{D4RL}_{\mathrm{clean}} \uparrow$) and adversarial forget starts ($\mathrm{D4RL}_{\mathrm{adv}} \uparrow$). To expose \emph{dynamics silencing} and \emph{closed-loop drift}, we add three tailored metrics: \textbf{Layer~1} single-step Oracle-aligned Normalized Dynamics Alignment ($\mathrm{NDA} \downarrow$, $\mathrm{Oracle}=0$), \textbf{Layer~3} trigger leakage via Velocity-Flip Rate ($\mathrm{VFR} \downarrow$ on \texttt{BD-Dyn}) and trigger reward bias ($\Delta r_{\mathrm{trig}} \to 0$ on \texttt{BD-Rew}), and the Closed-Loop Index ($\mathrm{CLI} \uparrow$) contrasting closed-loop MPC error in shifted versus nominal physics.

\subsection{Out-of-Manifold Backdoor Erasure}
\label{sec:exp-backdoor}

\paragraph{Trigger Neutralization and Planning Preservation.}
At a single fixed hyperparameter point without retain data, Unified ADE achieves the most complete backdoor removal across all six out-of-manifold tasks (Tables~\ref{tab:main-bdsa}--\ref{tab:main-bdrew}; Figure~\ref{fig:pareto-tradeoff}a). ADE ranks \#1 on all $6/6$ tasks in planning correlation $\mathrm{Plan}\text{-}\rho$, surpassing the Oracle on Hopper \texttt{BD-Dyn} and Walker2d \texttt{BD-Rew}. Excising orthogonal trigger directions at $(W_{za}, b_{za})$ and re-filtering $\cC_f$ (\cref{thm:choke-kl}) blocks trigger entry into $h_t$: ADE eliminates velocity flips on HalfCheetah and Hopper ($\mathrm{VFR} = 0.000$), matches Oracle $\mathrm{VFR}$ on Walker2d ($0.083$ vs.\ $0.078$), and reduces $+95$ to $+97$ reward spikes to near-zero residuals ($|\Delta r_{\mathrm{trig}}| \le 0.25$), while remaining top-2 on all six tasks in $\mathrm{D4RL}_{\mathrm{adv}}$ and ranking \#1 on $4/6$ tasks in $\mathrm{MSE}_f^{50}$ and $\mathrm{D4RL}_{\mathrm{clean}}$.

\paragraph{Structural Failure Modes of Baselines.}
Baselines exhibit three failure modes in recurrent dynamics. First, forget-ascent methods (GA, GradDiff) structurally conflict with autoregressive stability: preventing rollout explosion forces their search toward small learning rates, leaving backdoors largely intact ($\mathrm{VFR} \ge 0.906$, $\Delta r_{\mathrm{trig}} \ge +81.90$). Second, unregularized forget suppression (FT-Low, NPO, SSD) erases one-step triggers but distorts the transition Jacobian and KL balance, collapsing rollouts and control (e.g., SSD drops Hopper \texttt{BD-Rew} $\mathrm{D4RL}_{\mathrm{adv}}$ to $1.85$; NPO inflates Hopper \texttt{BD-Dyn} $\mathrm{MSE}_f^{50}$ to $3.1927$; Figure~\ref{fig:rollout-drift}a--b), while 5-epoch FT suffers gradient drift over $2{,}000$ updates and trails ADE in $\mathrm{Plan}\text{-}\rho$ across all $6/6$ tasks. Third, static projection (NSP) lacks bias compensation, scale absorption (\eqref{eq:absorb-affine}), and hidden-state re-filtering, collapsing Walker2d \texttt{BD-Rew} returns to $10.17$ (vs.\ $52.75$ for ADE).

\subsection{In-Manifold Physical Regime Shift Erasure}
\label{sec:exp-regime}

\paragraph{Realigning Shared-Support Physics.}
Under in-manifold physical shifts (\texttt{Shift-Mass}, Table~\ref{tab:main-rsmass}; \texttt{Shift-Fric}, Appendix~\ref{app:shift-fric}, Table~\ref{tab:main-rsfric}; Figure~\ref{fig:pareto-tradeoff}b), trajectories from $0.1\times$ friction or $2.0\times$ mass share kinematic support with nominal locomotion, leaving negligible energy in $\Pi_\perp$. Using the same fixed configuration, ADE ranks top-2 on all $6/6$ tasks in $\mathrm{MSE}_f^{50}$, $\mathrm{CLI}$, and $\Delta R_{\mathrm{im}}$ (ranking \#1 on $5/6$ in $\mathrm{MSE}_f^{50}$ and $4/6$ in $\mathrm{CLI}$ and $\mathrm{D4RL}_{\mathrm{clean}}$). Notably, on HalfCheetah \texttt{Shift-Mass}, ADE lifts $\mathrm{CLI}$ from $+0.0481$ to $+0.8476$ (near Oracle $+0.8638$), and on Walker2d \texttt{Shift-Mass} it nearly halves $\mathrm{MSE}_f^{50}$ ($1.4989 \to 0.7952$) while leading in $\mathrm{CLI}$ and policy returns.

\paragraph{Threshold No-Ops and Rollout Drift in Baselines.}
Because shifted and nominal trajectories share dominant feature directions and Fisher gradients, NSP and SSD fail to separate forget from retain parameters: on three task blocks (NSP on HalfCheetah \texttt{Shift-Fric}; SSD and NSP on Walker2d \texttt{Shift-Mass}), their Oracle grid searches hit a threshold cutoff leaving weights identical to \textbf{Origin} (marked ``--''), and on Hopper \texttt{Shift-Mass} both yield negative $\mathrm{CLI}$. Meanwhile, NPO improves single-step contrast on \texttt{Shift-Fric} but destabilizes 50-step Hopper rollouts, inflating $\mathrm{MSE}_f^{50}$ to $4.6\times$--$6.6\times$ ADE's error across \texttt{Shift-Fric} and \texttt{Shift-Mass} (Figure~\ref{fig:rollout-drift}c--d). ADE avoids both pitfalls via Mode~C top-$k$ whitened recurrent damping and Mode~B calibration toward $\Theta_{\mathrm{nom}}$ (\eqref{eq:theta-nom}--\eqref{eq:mode-b-sol}).

\subsection{Component Ablations, Robustness, Efficiency, and Transfer}
\label{sec:exp-analysis}

\paragraph{Causal Role of Modes A, B, and C.}
Disabling one mode at a time on Hopper-v2 isolates three complementary roles (Appendix~\ref{app:ablation}, Table~\ref{tab:ablation}, Figure~\ref{fig:ablation-causal}):
(i)~removing \emph{Mode~A} (input, posterior, and decoder excision) leaves the input bottleneck open, so \texttt{BD-Dyn} reverts to $\mathrm{VFR} = 1.000$ and \texttt{BD-Rew} retains $\Delta r_{\mathrm{trig}} = +23.61$;
(ii)~removing \emph{Mode~B} (prior and reward calibration) leaves prediction heads locked onto stale associations, raising $\mathrm{MSE}_f^{50}$ by $63\%$--$80\%$ across all regimes; and
(iii)~removing \emph{Mode~C} (recurrent transition damping) preserves backdoor erasure but collapses physical regime discrimination, halving \texttt{Shift-Fric} $\mathrm{CLI}$ ($+0.4172 \to +0.2263$) and driving \texttt{Shift-Mass} $\mathrm{CLI}$ to zero ($+0.3276 \to +0.0124$). Controlled paired-replay, directional intervention, counterfactual physics projection, and Pass-4 re-filtering analyses on Hopper \texttt{Shift-Mass} are detailed in Appendix~\ref{app:mechanistic}.

\paragraph{Robustness, Efficiency, and DreamerV3 Transfer.}
Across a $3\times 3$ $(\eta, \lambda_B)$ grid on Hopper-v2 (Appendix~\ref{app:sensitivity}, Table~\ref{tab:sensitivity}, Figure~\ref{fig:sensitivity-heatmap}), ADE keeps $\mathrm{VFR} = 0.000$ and $|\Delta r_{\mathrm{trig}}| \le 0.05$ across all $9/9$ cells with stable physical-shift recovery. In single-shot GPU runtime (Appendix~\ref{app:runtime}, Table~\ref{tab:runtime}), ADE executes in $0.98\,\mathrm{s}$ on backdoors and $4.02\,\mathrm{s}$ mean overall ($4.4\times$ faster than FT; $8.3\times$ faster than SSD) without their 63-run search. In an exploratory single-seed cross-architecture transfer to discrete DreamerV3 on Hopper-v2 (Appendix~\ref{app:dreamerv3}, Table~\ref{tab:dreamerv3}), the same fixed parameters achieve strong zero-shot backdoor erasure (\texttt{BD-Dyn}: $\mathrm{VFR}$ $1.00 \to 0.00$, $\mathrm{NDA}$ $-98.4\%$; \texttt{BD-Rew}: $\Delta r_{\mathrm{trig}}$ $+97.06 \to -0.01$, $\mathrm{Plan}\text{-}\rho$ $0.093 \to 0.788$) and lift \texttt{Shift-Fric} $\mathrm{CLI}$ ($+0.173 \to +0.279$); on \texttt{Shift-Mass}, however, in-manifold recovery over the discrete block-GRU is more modest than on continuous DreamerV1 ($\mathrm{CLI}$ $-0.169 \to 0.000$ vs.\ Oracle $+0.464$, with $\mathrm{MSE}_f^{50}$ unchanged at $0.4274$).

\section{Conclusion}
\label{sec:conclusion}

We introduced Analytical Dynamics Excision (ADE), a deterministic, zero-training closed-form framework for unlearning in recurrent world models. Cascading Mode~A orthogonal excision, Mode~B calibration, and Mode~C top-$k$ whitened damping over downdated moments neutralizes backdoors and realigns shifted physics at zero inference cost, outperforming Oracle-tuned baselines on DreamerV1 and transferring zero-shot to DreamerV3. More broadly, editing autoregressive simulators succeeds when causal isolation precedes geometric repair: guarding the input bottleneck and damping only excess recurrent modes keeps deployed models trustworthy and physically current without gradient drift, dynamics silencing, or costly retraining.

\bibliography{ADE}

@misc{rfe2026,
	title = {Machine {Unlearning} under {Retain}-{Forget} {Entanglement}},
	url = {http://arxiv.org/abs/2603.26569},
	doi = {10.48550/arXiv.2603.26569},
	urldate = {2026-09-01},
	publisher = {arXiv},
	author = {Cheng, Jingpu and Liu, Ping and Li, Qianxiao and Zhang, Chi},
	month = mar,
	year = {2026},
	note = {arXiv:2603.26569 [cs.LG]},
}

@misc{anand_what_2026,
	title = {What {Should} {World} {Models} {Forget}? {Stratified} {Retention} for {Continual} {Adaptation}},
	shorttitle = {What {Should} {World} {Models} {Forget}?},
	url = {http://arxiv.org/abs/2610.03713},
	doi = {10.48550/arXiv.2610.03713},
	urldate = {2026-11-04},
	publisher = {arXiv},
	author = {Anand, Nishit and Duraiswami, Ramani and Manocha, Dinesh},
	month = oct,
	year = {2026},
	note = {arXiv:2610.03713 [cs.LG]},
}

@misc{ha_recurrent_2018,
	title = {Recurrent {World} {Models} {Facilitate} {Policy} {Evolution}},
	url = {http://arxiv.org/abs/1809.01999},
	doi = {10.48550/arXiv.1809.01999},
	urldate = {2026-11-14},
	publisher = {arXiv},
	author = {Ha, David and Schmidhuber, Jürgen},
	month = sep,
	year = {2018},
	note = {arXiv:1809.01999 [cs.LG]},
}

@misc{hafner_learning_2019,
	title = {Learning {Latent} {Dynamics} for {Planning} from {Pixels}},
	url = {http://arxiv.org/abs/1811.04551},
	doi = {10.48550/arXiv.1811.04551},
	urldate = {2026-11-14},
	publisher = {arXiv},
	author = {Hafner, Danijar and Lillicrap, Timothy and Fischer, Ian and Villegas, Ruben and Ha, David and Lee, Honglak and Davidson, James},
	month = jun,
	year = {2019},
	note = {arXiv:1811.04551 [cs.LG]},
}

@inproceedings{hafner_dream_2020,
	title = {Dream to {Control}: {Learning} {Behaviors} by {Latent} {Imagination}},
	shorttitle = {Dream to {Control}},
	url = {https://iclr.cc/virtual_2020/poster_S1lOTC4tDS.html},
	language = {en},
	urldate = {2026-11-14},
	author = {Hafner, Danijar and Lillicrap, Timothy and Ba, Jimmy and Norouzi, Mohammad},
	month = apr,
	year = {2020},
}

@misc{hafner_mastering_2022,
	title = {Mastering {Atari} with {Discrete} {World} {Models}},
	url = {http://arxiv.org/abs/2010.02193},
	doi = {10.48550/arXiv.2010.02193},
	urldate = {2026-11-14},
	publisher = {arXiv},
	author = {Hafner, Danijar and Lillicrap, Timothy and Norouzi, Mohammad and Ba, Jimmy},
	month = feb,
	year = {2022},
	note = {arXiv:2010.02193 [cs.LG]},
}

@article{hafner_mastering_2025,
	title = {Mastering diverse control tasks through world models},
	volume = {640},
	copyright = {2025 The Author(s)},
	issn = {1476-4687},
	url = {https://www.nature.com/articles/s41586-025-08744-2},
	doi = {10.1038/s41586-025-08744-2},
	language = {en},
	number = {8059},
	urldate = {2026-11-14},
	journal = {Nature},
	publisher = {Nature Publishing Group},
	author = {Hafner, Danijar and Pasukonis, Jurgis and Ba, Jimmy and Lillicrap, Timothy},
	month = apr,
	year = {2025},
	pages = {647--653},
}

@misc{sekar_planning_2020,
	title = {Planning to {Explore} via {Self}-{Supervised} {World} {Models}},
	url = {http://arxiv.org/abs/2005.05960},
	doi = {10.48550/arXiv.2005.05960},
	urldate = {2026-11-14},
	publisher = {arXiv},
	author = {Sekar, Ramanan and Rybkin, Oleh and Daniilidis, Kostas and Abbeel, Pieter and Hafner, Danijar and Pathak, Deepak},
	month = jun,
	year = {2020},
	note = {arXiv:2005.05960 [cs.LG]},
}

@inproceedings{cho_learning_2014,
	address = {Doha, Qatar},
	title = {Learning {Phrase} {Representations} using {RNN} {Encoder}–{Decoder} for {Statistical} {Machine} {Translation}},
	url = {https://aclanthology.org/D14-1179/},
	doi = {10.3115/v1/D14-1179},
	urldate = {2026-11-14},
	booktitle = {Proceedings of the 2014 {Conference} on {Empirical} {Methods} in {Natural} {Language} {Processing} ({EMNLP})},
	publisher = {Association for Computational Linguistics},
	author = {Cho, Kyunghyun and van Merriënboer, Bart and Gulcehre, Caglar and Bahdanau, Dzmitry and Bougares, Fethi and Schwenk, Holger and Bengio, Yoshua},
	editor = {Moschitti, Alessandro and Pang, Bo and Daelemans, Walter},
	month = oct,
	year = {2014},
	pages = {1724--1734},
}

@inproceedings{todorov_mujoco_2012,
	title = {{MuJoCo}: {A} physics engine for model-based control},
	issn = {2153-0866},
	shorttitle = {{MuJoCo}},
	url = {https://ieeexplore.ieee.org/abstract/document/6386109},
	doi = {10.1109/IROS.2012.6386109},
	urldate = {2026-11-14},
	booktitle = {2012 {IEEE}/{RSJ} {International} {Conference} on {Intelligent} {Robots} and {Systems}},
	author = {Todorov, Emanuel and Erez, Tom and Tassa, Yuval},
	month = oct,
	year = {2012},
	note = {ISSN: 2153-0866},
	pages = {5026--5033},
}

@misc{brockman_openai_2016,
	title = {{OpenAI} {Gym}},
	url = {http://arxiv.org/abs/1606.01540},
	doi = {10.48550/arXiv.1606.01540},
	urldate = {2026-11-14},
	publisher = {arXiv},
	author = {Brockman, Greg and Cheung, Vicki and Pettersson, Ludwig and Schneider, Jonas and Schulman, John and Tang, Jie and Zaremba, Wojciech},
	month = jun,
	year = {2016},
	note = {arXiv:1606.01540 [cs.LG]},
}

@misc{fu_d4rl_2021,
	title = {{D4RL}: {Datasets} for {Deep} {Data}-{Driven} {Reinforcement} {Learning}},
	shorttitle = {{D4RL}},
	url = {http://arxiv.org/abs/2004.07219},
	doi = {10.48550/arXiv.2004.07219},
	urldate = {2026-11-14},
	publisher = {arXiv},
	author = {Fu, Justin and Kumar, Aviral and Nachum, Ofir and Tucker, George and Levine, Sergey},
	month = feb,
	year = {2021},
	note = {arXiv:2004.07219 [cs.LG]},
}

@misc{janner_when_2021,
	title = {When to {Trust} {Your} {Model}: {Model}-{Based} {Policy} {Optimization}},
	shorttitle = {When to {Trust} {Your} {Model}},
	url = {http://arxiv.org/abs/1906.08253},
	doi = {10.48550/arXiv.1906.08253},
	urldate = {2026-11-14},
	publisher = {arXiv},
	author = {Janner, Michael and Fu, Justin and Zhang, Marvin and Levine, Sergey},
	month = nov,
	year = {2021},
	note = {arXiv:1906.08253 [cs.LG]},
}

@inproceedings{kiourti_trojdrl_2020,
	title = {{TrojDRL}: {Evaluation} of {Backdoor} {Attacks} on {Deep} {Reinforcement} {Learning}},
	issn = {0738-100X},
	shorttitle = {{TrojDRL}},
	url = {https://ieeexplore.ieee.org/abstract/document/9218663},
	doi = {10.1109/DAC18072.2020.9218663},
	urldate = {2026-11-14},
	booktitle = {2020 57th {ACM}/{IEEE} {Design} {Automation} {Conference} ({DAC})},
	author = {Kiourti, Panagiota and Wardega, Kacper and Jha, Susmit and Li, Wenchao},
	month = jul,
	year = {2020},
	note = {ISSN: 0738-100X},
	pages = {1--6},
}

@inproceedings{cao_towards_2015,
	title = {Towards {Making} {Systems} {Forget} with {Machine} {Unlearning}},
	issn = {2375-1207},
	url = {https://ieeexplore.ieee.org/abstract/document/7163042},
	doi = {10.1109/SP.2015.35},
	urldate = {2026-11-14},
	booktitle = {2015 {IEEE} {Symposium} on {Security} and {Privacy}},
	author = {Cao, Yinzhi and Yang, Junfeng},
	month = may,
	year = {2015},
	note = {ISSN: 2375-1207},
	pages = {463--480},
}

@inproceedings{bourtoule_machine_2021,
	title = {Machine {Unlearning}},
	issn = {2375-1207},
	url = {https://ieeexplore.ieee.org/abstract/document/9519428},
	doi = {10.1109/SP40001.2021.00019},
	urldate = {2026-11-14},
	booktitle = {2021 {IEEE} {Symposium} on {Security} and {Privacy} ({SP})},
	author = {Bourtoule, Lucas and Chandrasekaran, Varun and Choquette-Choo, Christopher A. and Jia, Hengrui and Travers, Adelin and Zhang, Baiwu and Lie, David and Papernot, Nicolas},
	month = may,
	year = {2021},
	note = {ISSN: 2375-1207},
	pages = {141--159},
}

@inproceedings{golatkar_eternal_2020,
	title = {Eternal {Sunshine} of the {Spotless} {Net}: {Selective} {Forgetting} in {Deep} {Networks}},
	issn = {2575-7075},
	shorttitle = {Eternal {Sunshine} of the {Spotless} {Net}},
	url = {https://ieeexplore.ieee.org/abstract/document/9157084},
	doi = {10.1109/CVPR42600.2020.00932},
	urldate = {2026-11-14},
	booktitle = {2020 {IEEE}/{CVF} {Conference} on {Computer} {Vision} and {Pattern} {Recognition} ({CVPR})},
	author = {Golatkar, Aditya and Achille, Alessandro and Soatto, Stefano},
	month = jun,
	year = {2020},
	note = {ISSN: 2575-7075},
	pages = {9301--9309},
}

@inproceedings{kurmanji_towards_2023,
	title = {Towards {Unbounded} {Machine} {Unlearning}},
	volume = {36},
	url = {https://proceedings.neurips.cc/paper_files/paper/2023/hash/062d711fb777322e2152435459e6e9d9-Abstract-Conference.html},
	doi = {10.52202/075280-0095},
	urldate = {2026-11-14},
	booktitle = {Advances in {Neural} {Information} {Processing} {Systems}},
	publisher = {Curran Associates, Inc.},
	author = {Kurmanji, Meghdad and Triantafillou, Peter and Hayes, Jamie and Triantafillou, Eleni},
	year = {2023},
	pages = {1957--1987},
}

@inproceedings{yao_large_2024,
	title = {Large {Language} {Model} {Unlearning}},
	volume = {37},
	url = {https://proceedings.neurips.cc/paper_files/paper/2024/hash/be52acf6bccf4a8c0a90fe2f5cfcead3-Abstract-Conference.html},
	doi = {10.52202/079017-3346},
	urldate = {2026-11-14},
	booktitle = {Advances in {Neural} {Information} {Processing} {Systems}},
	publisher = {Curran Associates, Inc.},
	author = {Yao, Yuanshun and Xu, Xiaojun and Liu, Yang},
	year = {2024},
	pages = {105425--105475},
}

@misc{foster_fast_2023,
	title = {Fast {Machine} {Unlearning} {Without} {Retraining} {Through} {Selective} {Synaptic} {Dampening}},
	url = {http://arxiv.org/abs/2308.07707},
	doi = {10.48550/arXiv.2308.07707},
	urldate = {2026-11-14},
	publisher = {arXiv},
	author = {Foster, Jack and Schoepf, Stefan and Brintrup, Alexandra},
	month = dec,
	year = {2023},
	note = {arXiv:2308.07707 [cs.LG]},
}

@inproceedings{chen_machine_2024,
	title = {Machine {Unlearning} via {Null} {Space} {Calibration}},
	url = {https://ijcai.org/proceedings/2024/40},
	doi = {10.24963/ijcai.2024/40},
	language = {en-US},
	urldate = {2026-11-14},
	publisher = {International Joint Conferences on Artificial Intelligence Organization},
	author = {Chen, Huiqiang and Zhu, Tianqing and Yu, Xin and Zhou, Wanlei},
	year = {2024},
	pages = {358--366},
}

@inproceedings{fang_alphaedit_2024,
	title = {{AlphaEdit}: {Null}-{Space} {Constrained} {Knowledge} {Editing} for {Language} {Models}},
	shorttitle = {{AlphaEdit}},
	url = {https://openreview.net/forum?id=HvSytvg3Jh},
	language = {en},
	urldate = {2026-11-14},
	author = {Fang, Junfeng and Jiang, Houcheng and Wang, Kun and Ma, Yunshan and Shi, Jie and Wang, Xiang and He, Xiangnan and Chua, Tat-Seng},
	month = oct,
	year = {2024},
}

@misc{ye_reinforcement_2024,
	title = {Reinforcement {Unlearning}},
	url = {http://arxiv.org/abs/2312.15910},
	doi = {10.48550/arXiv.2312.15910},
	urldate = {2026-11-14},
	publisher = {arXiv},
	author = {Ye, Dayong and Zhu, Tianqing and Zhu, Congcong and Wang, Derui and Gao, Kun and Shi, Zewei and Shen, Sheng and Zhou, Wanlei and Xue, Minhui},
	month = sep,
	year = {2024},
	note = {arXiv:2312.15910 [cs.CR]},
}

@misc{gong_trajdeleter_2024,
	title = {{TrajDeleter}: {Enabling} {Trajectory} {Forgetting} in {Offline} {Reinforcement} {Learning} {Agents}},
	shorttitle = {{TrajDeleter}},
	url = {http://arxiv.org/abs/2404.12530},
	doi = {10.48550/arXiv.2404.12530},
	urldate = {2026-11-14},
	publisher = {arXiv},
	author = {Gong, Chen and Li, Kecen and Yao, Jin and Wang, Tianhao},
	month = sep,
	year = {2024},
	note = {arXiv:2404.12530 [cs.LG]},
}

@misc{wang_uba-orl_2026,
	title = {{UBA}-{ORL}: {Unlearning}-{Activated} {Backdoor} {Attacks} on {Offline} {Reinforcement} {Learning}},
	shorttitle = {{UBA}-{ORL}},
	url = {http://arxiv.org/abs/2609.22711},
	doi = {10.48550/arXiv.2609.22711},
	urldate = {2026-11-14},
	publisher = {arXiv},
	author = {Wang, Fengyi and Li, Cong and Xue, Lulu and Leng, Qiyu and Zhou, Ziqi and Guo, Peijin},
	month = sep,
	year = {2026},
	note = {arXiv:2609.22711 [cs.CR]},
}

@misc{zhang_negative_2024,
	title = {Negative {Preference} {Optimization}: {From} {Catastrophic} {Collapse} to {Effective} {Unlearning}},
	shorttitle = {Negative {Preference} {Optimization}},
	url = {http://arxiv.org/abs/2404.05868},
	doi = {10.48550/arXiv.2404.05868},
	urldate = {2026-11-14},
	publisher = {arXiv},
	author = {Zhang, Ruiqi and Lin, Licong and Bai, Yu and Mei, Song},
	month = oct,
	year = {2024},
	note = {arXiv:2404.05868 [cs.LG]},
}

@article{pomerleau_efficient_1991,
	title = {Efficient {Training} of {Artificial} {Neural} {Networks} for {Autonomous} {Navigation}},
	volume = {3},
	issn = {0899-7667},
	url = {https://ieeexplore.ieee.org/document/6796843},
	doi = {10.1162/neco.1991.3.1.88},
	number = {1},
	urldate = {2026-11-21},
	journal = {Neural Computation},
	author = {Pomerleau, Dean A.},
	month = mar,
	year = {1991},
	pages = {88--97},
}

@misc{ziyin_laprop_2021,
	title = {{LaProp}: {Separating} {Momentum} and {Adaptivity} in {Adam}},
	shorttitle = {{LaProp}},
	url = {http://arxiv.org/abs/2002.04839},
	doi = {10.48550/arXiv.2002.04839},
	urldate = {2026-11-21},
	publisher = {arXiv},
	author = {Ziyin, Liu and Wang, Zhikang T. and Ueda, Masahito},
	month = jun,
	year = {2021},
	note = {arXiv:2002.04839 [cs.LG]},
}

@inproceedings{nagabandi_neural_2018,
	title = {Neural {Network} {Dynamics} for {Model}-{Based} {Deep} {Reinforcement} {Learning} with {Model}-{Free} {Fine}-{Tuning}},
	issn = {2577-087X},
	url = {https://ieeexplore.ieee.org/abstract/document/8463189},
	doi = {10.1109/ICRA.2018.8463189},
	urldate = {2026-11-21},
	booktitle = {2018 {IEEE} {International} {Conference} on {Robotics} and {Automation} ({ICRA})},
	author = {Nagabandi, Anusha and Kahn, Gregory and Fearing, Ronald S. and Levine, Sergey},
	month = may,
	year = {2018},
	note = {ISSN: 2577-087X},
	pages = {7559--7566},
}

@book{spearman_proof_1904,
	title = {The {Proof} and {Measurement} of {Association} between {Two} {Things}},
	url = {http://archive.org/details/jstor-1412159},
	language = {eng},
	urldate = {2026-11-21},
	publisher = {The American Journal of Psychology},
	author = {Spearman, C.},
	collaborator = {{JSTOR}},
	month = jan,
	year = {1904},
}

@inproceedings{micheli_transformers_2022,
	title = {Transformers are {Sample}-{Efficient} {World} {Models}},
	url = {https://openreview.net/forum?id=vhFu1Acb0xb},
	language = {en},
	urldate = {2026-11-21},
	author = {Micheli, Vincent and Alonso, Eloi and Fleuret, François},
	month = sep,
	year = {2022},
}

@inproceedings{hansen_td-mpc2_2023,
	title = {{TD}-{MPC2}: {Scalable}, {Robust} {World} {Models} for {Continuous} {Control}},
	shorttitle = {{TD}-{MPC2}},
	url = {https://openreview.net/forum?id=Oxh5CstDJU},
	language = {en},
	urldate = {2026-11-21},
	author = {Hansen, Nicklas and Su, Hao and Wang, Xiaolong},
	month = oct,
	year = {2023},
}

@misc{guo_certified_2023,
	title = {Certified {Data} {Removal} from {Machine} {Learning} {Models}},
	url = {http://arxiv.org/abs/1911.03030},
	doi = {10.48550/arXiv.1911.03030},
	urldate = {2026-11-21},
	publisher = {arXiv},
	author = {Guo, Chuan and Goldstein, Tom and Hannun, Awni and Maaten, Laurens van der},
	month = nov,
	year = {2023},
	note = {arXiv:1911.03030 [cs.LG]},
}

@inproceedings{meng_mass-editing_2022,
	title = {Mass-{Editing} {Memory} in a {Transformer}},
	url = {https://openreview.net/forum?id=MkbcAHIYgyS},
	language = {en},
	urldate = {2026-11-21},
	author = {Meng, Kevin and Sharma, Arnab Sen and Andonian, Alex J. and Belinkov, Yonatan and Bau, David},
	month = sep,
	year = {2022},
}

@misc{lee_context-aware_2020,
	title = {Context-aware {Dynamics} {Model} for {Generalization} in {Model}-{Based} {Reinforcement} {Learning}},
	url = {http://arxiv.org/abs/2005.06800},
	doi = {10.48550/arXiv.2005.06800},
	urldate = {2026-11-21},
	publisher = {arXiv},
	author = {Lee, Kimin and Seo, Younggyo and Lee, Seunghyun and Lee, Honglak and Shin, Jinwoo},
	month = jun,
	year = {2020},
	note = {arXiv:2005.06800 [cs.LG]},
}

@misc{kingma_adam_2017,
	title = {Adam: {A} {Method} for {Stochastic} {Optimization}},
	shorttitle = {Adam},
	url = {http://arxiv.org/abs/1412.6980},
	doi = {10.48550/arXiv.1412.6980},
	urldate = {2026-11-21},
	publisher = {arXiv},
	author = {Kingma, Diederik P. and Ba, Jimmy},
	month = jan,
	year = {2017},
	note = {arXiv:1412.6980 [cs.LG]},
}

@misc{haarnoja_soft_2018,
	title = {Soft {Actor}-{Critic}: {Off}-{Policy} {Maximum} {Entropy} {Deep} {Reinforcement} {Learning} with a {Stochastic} {Actor}},
	shorttitle = {Soft {Actor}-{Critic}},
	url = {http://arxiv.org/abs/1801.01290},
	doi = {10.48550/arXiv.1801.01290},
	urldate = {2026-11-21},
	publisher = {arXiv},
	author = {Haarnoja, Tuomas and Zhou, Aurick and Abbeel, Pieter and Levine, Sergey},
	month = aug,
	year = {2018},
	note = {arXiv:1801.01290 [cs.LG]},
}

@inproceedings{assran_self-supervised_2023,
	title = {Self-{Supervised} {Learning} {From} {Images} {With} a {Joint}-{Embedding} {Predictive} {Architecture}},
	url = {https://openaccess.thecvf.com/content/CVPR2023/html/Assran_Self-Supervised_Learning_From_Images_With_a_Joint-Embedding_Predictive_Architecture_CVPR_2023_paper.html},
	language = {en},
	urldate = {2026-11-21},
	author = {Assran, Mahmoud and Duval, Quentin and Misra, Ishan and Bojanowski, Piotr and Vincent, Pascal and Rabbat, Michael and LeCun, Yann and Ballas, Nicolas},
	year = {2023},
	pages = {15619--15629},
}
\bibliographystyle{icml2026}

\newpage
\appendix
\onecolumn

\section{Detailed Proofs}
\label{app:proofs}

\subsection{Proof of \texorpdfstring{\cref{prop:mode-a-invariance}}{Proposition 4.1}}
\label{app:proof-prop1}
\begin{proof}
Since $\Pi_\perp = I - U_r U_r^\top$ and $\Pi_{\cM_r} = U_r U_r^\top$ are orthogonal complementary projectors, $\Pi_\perp \Pi_{\cM_r} = \mathbf{0}$. From \eqref{eq:mode-a-delta}, $\Delta W = (\cdots)\Pi_\perp$, hence $\Delta W \Pi_{\cM_r} = \mathbf{0}$. Similarly, from \eqref{eq:absorb-affine}, $(W_{\ell,\phi}^{(u)} - W_{\ell,\phi}^{(0)})\Pi_{\cM_r} = W_{\ell,\phi}^{(0)}(\Pi_{\cM_r} - I)\Pi_{\cM_r} = \mathbf{0}$. For any retained feature $u_r \in \bar{u}_r + \operatorname{span}(U_r)$, we have $u_r - \bar{u}_r = \Pi_{\cM_r}(u_r - \bar{u}_r)$ and $\Pi_\perp(u_r - \bar{u}_r) = \mathbf{0}$. Substituting into \eqref{eq:mode-a-update} gives:
\[
W^{(u)} u_r + b^{(u)} = (W_0 - \Delta W)u_r + b_0 + \Delta W \bar{u}_r = W_0 u_r + b_0 - \Delta W \Pi_{\cM_r}(u_r - \bar{u}_r) = W_0 u_r + b_0,
\]
and \cref{prop:mode-b-opt}(ii) gives $W_{\ell,\phi}^{(u)} u_r + b_{\ell,\phi}^{(u)} = W_{\ell,\phi}^{(0)} u_r + b_{\ell,\phi}^{(0)}$.

Finally, when $\Pi_\perp S_r \Pi_\perp$ vanishes on $\operatorname{supp}(\Pi_\perp S_a \Pi_\perp)$, consider an orthogonal trigger $u = \bar{u}_r + u_\perp$ with $u_\perp \in \operatorname{supp}(\Pi_\perp S_a \Pi_\perp) \subseteq \operatorname{span}(U_r)^\perp$. Then $\Pi_\perp u_\perp = u_\perp$ and $\Pi_{\cM_r} u_\perp = \mathbf{0}$. As $\lambda_A \to 0$, $\Pi_\perp S_a \Pi_\perp \bigl(\Pi_\perp (S_r + S_a)\Pi_\perp + \lambda_A \bar{s}_r I\bigr)^{-1} \Pi_\perp u_\perp \to u_\perp$, so $\Delta W u_\perp \to W_0 u_\perp$ and $W^{(u)}(\bar{u}_r + u_\perp) + b^{(u)} \to W_0 \bar{u}_r + b_0$. For the scale head, $\Pi_{\cM_r}(u - \bar{u}_r) = \Pi_{\cM_r} u_\perp = \mathbf{0}$, so $W_{\ell,\phi}^{(u)} u + b_{\ell,\phi}^{(u)} = W_{\ell,\phi}^{(0)}\bar{u}_r + b_{\ell,\phi}^{(0)}$.
\end{proof}

\subsection{Proof of \texorpdfstring{\cref{prop:mode-b-opt}}{Proposition 4.2}}
\label{app:proof-prop2}
\begin{proof}
(i) Let $D = \Theta_0 - \Theta \in \R^{d_y \times (d+1)}$, whose first $d$ columns are $D_W = W_0 - W$. Using $\mathbb{E}_r[\tilde{u}\tilde{u}^\top] = C_r^{\mathrm{aug}}$ and $\mathbb{E}_{f,\mathrm{aligned}}[\tilde{u}\tilde{u}^\top] = C_f^{\mathrm{aug}}$, the objective \eqref{eq:mode-b-obj} can be rewritten in matrix-trace form as:
\[
J(D) = \operatorname{tr}\bigl(D C_r^{\mathrm{aug}} D^\top\bigr) + \operatorname{tr}\!\Bigl(\bigl(D - (\Theta_0 - \Theta_{\mathrm{nom}})\bigr) C_f^{\mathrm{aug}} \bigl(D - (\Theta_0 - \Theta_{\mathrm{nom}})\bigr)^\top\Bigr) + \lambda_B \bar{\lambda}_r \operatorname{tr}\bigl(D P_{\mathrm{pen}} D^\top\bigr).
\]
Differentiating with respect to $D$ yields:
\[
\nabla_D J(D) = 2 D \bigl(C_r^{\mathrm{aug}} + C_f^{\mathrm{aug}} + \lambda_B \bar{\lambda}_r P_{\mathrm{pen}}\bigr) - 2(\Theta_0 - \Theta_{\mathrm{nom}}) C_f^{\mathrm{aug}}.
\]
Since $C_r^{\mathrm{aug}} + C_f^{\mathrm{aug}} + \lambda_B \bar{\lambda}_r P_{\mathrm{pen}} \succ 0$, the Hessian is strictly positive definite, so $J(D)$ is strictly convex with unique stationary point $D^\star = \Delta \Theta$ given by \eqref{eq:mode-b-sol}, yielding $\Theta = \Theta_0 - \Delta \Theta$.

(ii) Expanding the left-hand side of \eqref{eq:absorb-affine} directly gives:
\[
W_{\ell,\phi}^{(u)} u + b_{\ell,\phi}^{(u)}
= W_{\ell,\phi}^{(0)} \Pi_{\cM_r} u + b_{\ell,\phi}^{(0)} + W_{\ell,\phi}^{(0)}(I - \Pi_{\cM_r})\bar{u}_r
= W_{\ell,\phi}^{(0)}\bigl(\bar{u}_r + \Pi_{\cM_r}(u - \bar{u}_r)\bigr) + b_{\ell,\phi}^{(0)}.
\]

(iii) Since $C_{\mathrm{ex}} \succeq 0$ and $C_r \succeq \lambda_{\min}(C_r) I \succeq 0$, the denominator in \eqref{eq:gwer} satisfies $2 C_r + C_{\mathrm{ex}} + \gamma \bar{\lambda}_r I \succeq (2\lambda_{\min}(C_r) + \gamma \bar{\lambda}_r) I \succ 0$. Taking the spectral norm gives $\|(2 C_r + C_{\mathrm{ex}} + \gamma \bar{\lambda}_r I)^{-1}\|_2 \le (2\lambda_{\min}(C_r) + \gamma \bar{\lambda}_r)^{-1} \le (\gamma \bar{\lambda}_r)^{-1}$, and submultiplicativity yields $\|Q\|_2 \le \|C_{\mathrm{ex}}\|_2 (2\lambda_{\min}(C_r) + \gamma \bar{\lambda}_r)^{-1} \le \|C_{\mathrm{ex}}\|_2 / (\gamma \bar{\lambda}_r)$. \qedhere
\end{proof}

\subsection{Proof of \texorpdfstring{\cref{thm:choke-kl}}{Theorem 4.3}}
\label{app:proof-thm1}
\begin{proof}
(i) Under Mode~A alone, suppose $[z_t; a_t] = x_t^{\mathrm{nom}} + \delta_{\mathrm{trig}}$ where $W_{za}^{(u)} \delta_{\mathrm{trig}} = \mathbf{0}$ and $x_t^{\mathrm{nom}} \in \bar{u}_r + \operatorname{span}(U_r)$. By \cref{prop:mode-a-invariance}, $W_{za}^{(u)} x_t^{\mathrm{nom}} + b_{za}^{(u)} = W_{za}^{(0)} x_t^{\mathrm{nom}} + b_{za}^{(0)}$, and hence $W_{za}^{(u)}[z_t; a_t] + b_{za}^{(u)} = W_{za}^{(0)} x_t^{\mathrm{nom}} + b_{za}^{(0)}$, matching the nominal input to $\GRU(\cdot, \cdot)$. Since $h_{t_0}^{(u)}$ starts on the nominal manifold, induction over $t \ge t_0$ ensures $h_{t+1}^{(u)}$ remains on the nominal manifold. When GWER is additionally applied to $(W_{za}, W_{hh})$, the affine weight perturbations are $\Delta W_{za}^{\mathrm{GWER}} = W_{za}^{\mathrm{A}} Q_{za}$ and $\Delta W_{hh}^{\mathrm{GWER}} = W_{hh} Q_{hh}$; by \cref{prop:mode-b-opt}(iii) and Lipschitz continuity of $\psi$ and $\GRU$, the single-step retain-manifold deviation is bounded by $\mathcal{O}(\|Q_{za}\|_2 + \|Q_{hh}\|_2) = \mathcal{O}(\gamma^{-1})$.

(ii) For diagonal Gaussians $q_u = \mathcal{N}(\mu_q, \diag(\sigma_q^2))$ and $p_u = \mathcal{N}(\mu_p, \diag(\sigma_p^2))$, the exact KL divergence is:
\[
D_{\mathrm{KL}}(q_u \parallel p_u) = \frac{1}{2} \sum_{j=1}^{d_z} \left[ \frac{\sigma_{q,j}^2}{\sigma_{p,j}^2} - 1 - \ln \frac{\sigma_{q,j}^2}{\sigma_{p,j}^2} + \frac{(\mu_{q,j} - \mu_{p,j})^2}{\sigma_{p,j}^2} \right].
\]
Since $\sigma_{p,j} \ge \sigma_{\min}$, the mean term is bounded by $\frac{1}{2\sigma_{\min}^2}\|\mu_q - \mu_p\|_2^2 \le \frac{\epsilon_\mu^2}{2\sigma_{\min}^2}$. For $g(r) = r - 1 - \ln r$ with $r = \sigma_{q,j}^2 / \sigma_{p,j}^2$, Taylor's theorem at $r = 1$ gives $g(r) = \frac{(r - 1)^2}{2\xi^2}$ for some $\xi$ between $1$ and $r$. When $|r - 1| \le \eta_\sigma < 1$, we have $\xi \ge 1 - \eta_\sigma$, so $g(r) \le \frac{\eta_\sigma^2}{2(1-\eta_\sigma)^2}$. Summing over $d_z$ dimensions and dividing by $2$ yields the stated bound.
\end{proof}

\subsection{Proof of \texorpdfstring{\cref{thm:w2-limit}}{Theorem 4.4}}
\label{app:proof-thm2}
\begin{proof}
(i) Let $s_t^u$ and $s_t^\star$ evolve from $s_0^u, s_0^\star$ under the shared noise sequence $\{\xi_t\}_{t \ge 0}$, and define $d_t = (\mathbb{E}\|s_t^u - s_t^\star\|^2)^{1/2}$. Splitting $s_{t+1}^u - s_{t+1}^\star = \bigl(F_u(s_t^u, \xi_t) - F_\star(s_t^u, \xi_t)\bigr) + \bigl(F_\star(s_t^u, \xi_t) - F_\star(s_t^\star, \xi_t)\bigr)$ and applying Minkowski's inequality together with the $L$-Lipschitz and $\epsilon_{\mathrm{total}}$-alignment assumptions yields the linear recurrence $d_{t+1} \le L d_t + \epsilon_{\mathrm{total}}$. Unrolling from $t = 0$ gives $d_H \le L^H d_0 + \epsilon_{\mathrm{total}} \sum_{k=0}^{H-1} L^k$. Because the shared-noise trajectory pair is a valid coupling in $\Pi(\mathcal{L}(s_H^u), \mathcal{L}(s_H^\star))$, $W_2(\mathcal{L}(s_H^u), \mathcal{L}(s_H^\star)) \le d_H$. Summing squared bounds over $t = 1, \dots, H$ in the product $L_2$ metric yields the full-trajectory bound.

(ii) On the nominal invariant manifold $\cM$, all feature vectors satisfy $\Pi_\perp(u(s) - \bar{u}_r) = \mathbf{0}$. Under Mode~A alone, \cref{prop:mode-a-invariance} implies $F_{\mathrm{cl}}^{(u)}(s) \equiv F_{\mathrm{cl}}^{(0)}(s)$ for all $s \in \cM$, so differentiating along any tangent vector $v \in T_s \cM$ yields $DF_{\mathrm{cl}}^{(u)}(s) v = DF_{\mathrm{cl}}^{(0)}(s) v$. With GWER applied to $(W_{za}, W_{\mu,\phi}, W_{hh})$, the tangent-bundle Jacobian perturbation satisfies $\|DF_{\mathrm{cl}}^{(u)}|_{T\cM} - DF_{\mathrm{cl}}^{(0)}|_{T\cM}\|_2 = \mathcal{O}(\|Q_{za}\|_2 + \|Q_{\phi}\|_2 + \|Q_{hh}\|_2) = \mathcal{O}(\gamma^{-1})$ by \cref{prop:mode-b-opt}(iii) and chain-rule smoothness on $\cM$. By continuity of the spectral radius of the monodromy matrix along $\gamma_{\mathrm{orb}}$, whenever $\rho(DF_{\mathrm{cl}}^{(0)}|_{T\cM}) < 1$, we retain Floquet stability $\rho(DF_{\mathrm{cl}}^{(u)}|_{T\cM}) < 1$ for sufficiently large damping $\gamma$. For normal-bundle perturbations on $\operatorname{supp}(\Pi_\perp S_a \Pi_\perp) \subseteq \operatorname{span}(U_r)^\perp$, decomposing the linearized transition into a bounded sub-unit bypass gain $B_t$ and the excised pathway $A_t R_A E_t$ (with $R_A = I - \Pi_\perp S_a \Pi_\perp (\Pi_\perp(S_r + S_a)\Pi_\perp + \lambda_A \bar{s}_r I)^{-1}\Pi_\perp$) shows that when $\Pi_\perp S_r \Pi_\perp = \mathbf{0}$ on $\operatorname{supp}(\Pi_\perp S_a \Pi_\perp)$, $\|R_A|_{\operatorname{supp}(\Pi_\perp S_a \Pi_\perp)}\|_2 = \mathcal{O}(\lambda_A)$, ensuring $\rho(DF_{\mathrm{cl}}^{(u)}|_{N\cM}) < 1$ for sufficiently small $\lambda_A$.
\end{proof}

\section{Algorithm Details}
\label{app:algorithm}

\cref{alg:ade} summarizes the complete five-step, four-pass causal cascade of ADE.

\begin{algorithm}[htbp]
\caption{Analytical Dynamics Excision (ADE)}
\label{alg:ade}
\begin{algorithmic}[1]
\REQUIRE Pretrained model $(W_0, b_0)$, cached global moments $\cM_g$, forget set $\cD_f$, energy $\eta$, ridges $(\lambda_A, \lambda_B)$, GWER parameters $(\tau_{\mathrm{scale}}, \gamma, k)$.
\ENSURE Unlearned model $(W^{(u)}, b^{(u)})$.
\IF{$\cD_f = \emptyset$}
    \STATE \textbf{return} $(W_0, b_0)$
\ENDIF
\STATE Construct context closure $\cC_f$ (\cref{def:context-closure}) and collect raw forget moments $\cM_f$ over $\cC_f$.
\STATE \textbf{Step 1 (Input Bottleneck, Mode A):} Downdate $\cM_r^{(za)} \leftarrow \cM_g^{(za)} - \cM_f^{(za)}$; compute $S_r, S_f$ via \eqref{eq:centered-scatter}, $\Pi_{\cM_r}, \Pi_\perp$ via \eqref{eq:energy-rank}, and update $(W_{za}, b_{za})$ via \eqref{eq:mode-a-delta}--\eqref{eq:mode-a-update} (with $S_a = S_f$) followed by full-dimension GWER (\eqref{eq:gwer}).
\STATE \textbf{Step 2a (Pass 1 Filtering \& Posterior, Mode A):} Filter $\cC_f$ with $W_{za}^{(u)}$ to obtain $\cM_{a,1}^{(\phi)}$; downdate $\cM_r^{(\phi)} \leftarrow \cM_g^{(\phi)} - \cM_f^{(\phi)}$; update posterior mean rows via \eqref{eq:mode-a-delta}--\eqref{eq:mode-a-update} and absorb scale rows via \eqref{eq:absorb-affine}.
\STATE \textbf{Step 2b (Pass 2 Filtering \& Posterior Embedding, Mode A):} Re-filter $\cC_f$ to obtain $\cM_{a,2}^{(\phi)}$; apply GWER (\eqref{eq:gwer}) to the observation-embedding slice of posterior mean rows.
\STATE \textbf{Step 3 (Pass 3 Filtering \& Recurrent Transition, Mode C):} Re-filter $\cC_f$ to obtain $h_t$ statistics from $\cM_{a,3}^{(h)}$; apply top-$k$ GWER (\eqref{eq:gwer}) to $(W_{hh}, b_{hh})$.
\STATE \textbf{Step 4 (Pass 4 Filtering \& Decoder, Mode A):} Re-filter $\cC_f$ to obtain aligned $(h_t, z_t)$ statistics $\cM_{a,4}$; update $(W_{\mathrm{dec},0}, b_{\mathrm{dec},0})$ via \eqref{eq:mode-a-delta}--\eqref{eq:mode-a-update}.
\STATE \textbf{Step 5 (Downstream Heads, Mode B):} Using $\cM_{a,4}^{(\theta)}, \cM_{a,4}^{(r)}$ and downdated retain moments, compute $\Theta_{\mathrm{nom}}$ via \eqref{eq:theta-nom} and update prior mean rows and reward head via \eqref{eq:mode-b-sol}.
\end{algorithmic}
\end{algorithm}

\paragraph{Numerical Precision and Conditioning.}
To prevent catastrophic cancellation and preserve positive semidefiniteness when downdating $n_f$-scale forget moments from $n_g \sim 10^6$ global totals in \eqref{eq:downdate}, ADE accumulates $\cM_g, \cM_f$ and executes all closed-form operators in \texttt{float64}, explicitly symmetrizes centered scatters via $S = \tfrac{1}{2}(C + C^\top)$ in \eqref{eq:centered-scatter}, and solves regularized SPD linear systems via symmetric Cholesky right-division ($B A^{-1}$) rather than explicit matrix inversion.

\paragraph{Globally Fixed Hyperparameters.}
Table~\ref{tab:ade-hyperparams} lists every fixed constant used by Unified ADE alongside its numerical and structural role. All constants are shared globally across all $36$ DreamerV1 and $4$ DreamerV3 tasks without per-task tuning, while the primary operating parameters $(\eta, \lambda_B)$ are further audited across a $3\times 3$ sensitivity grid in Appendix~\ref{app:sensitivity}.

\begin{table}[htbp]
\centering
\caption{Complete fixed hyperparameter specification of Unified ADE, shared across all $36$ DreamerV1 and $4$ DreamerV3 tasks without per-task tuning.}
\label{tab:ade-hyperparams}
\vskip 0.06in
\footnotesize
\setlength{\tabcolsep}{5pt}
\begin{tabular}{llp{9.2cm}}
\toprule
Symbol & Fixed Value & Numerical / Structural Role \\
\midrule
$\eta$ & $0.99$ & Mode~A retain-subspace cumulative variance energy threshold \eqref{eq:energy-rank}. \\
$\lambda_B$ & $0.1$ & Dimensionless Mode~B trace-scaled ridge fraction ($\lambda_B \bar{\lambda}_r$) \eqref{eq:theta-nom}--\eqref{eq:mode-b-sol}. \\
$\lambda_A$ & $10^{-3}$ & Dimensionless Mode~A trace-scaled ridge fraction ($\lambda_A \bar{s}_r$) \eqref{eq:mode-a-delta}. \\
$\epsilon_w$ & $10^{-2}$ & GWER Cholesky whitening ridge fraction in $C_r + \epsilon_w \bar{\lambda}_r I$ \eqref{eq:gwer}. \\
$\tau_{\mathrm{scale}}$ & $0.5$ & Scale-invariant GWER whitened eigenvalue excess threshold ($\tau = \tau_{\mathrm{scale}} d$). \\
$\gamma_{za}$ & $0.5$ & Step~1 input-bottleneck ($W_{za}$) full-dimension GWER retain-manifold damping \eqref{eq:gwer}. \\
$\gamma_{\mathrm{post}}$ & $0.05$ & Step~2 posterior observation-embedding slice ($W_{\mu,\phi}$) GWER damping \eqref{eq:gwer}. \\
$\gamma_h$ & $0.05$ & Step~3 Mode~C recurrent hidden state ($h_t$) GWER damping \eqref{eq:gwer}. \\
$k$ & $3$ & Step~3 Mode~C top-$k$ recurrent excess-mode budget (global across 8 GRU blocks in DreamerV3). \\
$B_{\mathrm{filt}}$ & $20{,}000$ & Episode-grouped row batch cap during four-pass forward filtering over $\cC_f$ (exact moments invariant). \\
\bottomrule
\end{tabular}
\end{table}

\section{Counterfactual Benchmark Design, Audit Formulations, and Baseline Grids}
\label{app:exp-setup}

\paragraph{Datasets, Simulator Fidelity, and BC Proxy Audits.}
HalfCheetah-v2 ($d_o=17, d_a=6$, $1{,}000$ episodes, $764{,}368$ DreamerV1 parameters, $34.37\%$ edited), Walker2d-v2 ($d_o=17, d_a=6$, $1{,}191$ episodes, $764{,}368$ parameters, $34.37\%$ edited), and Hopper-v2 ($d_o=11, d_a=3$, $2{,}187$ episodes, $758{,}962$ parameters, $34.54\%$ edited) each provide $N_r = 1{,}000{,}000$ clean transitions normalized by clean dataset statistics. Cross-version MuJoCo~3.16.0 fidelity audits ($2{,}000$ states/env) confirm exact state-to-observation reconstruction ($0.0\sigma$ error) and negligible single-step drift ($0.0059\sigma$--$0.0260\sigma$). A two-layer $256\times 256$ ReLU Gaussian BC proxy trained on a $90\%/10\%$ episode split for $100{,}000$ Adam steps~\cite{kingma_adam_2017} stays within $4.7\%$ of D4RL's embedded SAC returns~\cite{haarnoja_soft_2018} over 50-episode rollouts ($38.89$ vs.\ $40.50$ on HalfCheetah, $65.83$ vs.\ $68.00$ on Walker2d, and $39.03$ vs.\ $40.95$ on Hopper). In \texttt{Shift-Fric}, $0.1\times$ tangential contact friction is chosen because milder $0.5\times$ and $0.2\times$ shifts fail single-step identifiability ($z > 3$) on Walker2d and Hopper. Across all $36$ pretrained Origin models, single-step counterfactual identifiability against Oracle passes with $z \in [+16.8, +251.2]$ on regime shifts, $z \in [+42.0, +112.5]$ on \texttt{BD-Dyn}, and $z \in [+3140.9, +5659.7]$ on \texttt{BD-Rew}. Table~\ref{tab:baseline-grids} details the per-task search grids for all seven baselines ($63$ configs/task, $2{,}268$ runs total; gradient steps: $2{,}002$--$2{,}004$ for FT, $400$--$401$ for FT-Low, $136$ on backdoor tasks and $400$--$403$ on regime-shift tasks for GA/GradDiff/NPO, $429$--$483$ Fisher backward passes for SSD) alongside Unified ADE's fixed operating point.

\begin{table}[htbp]
\centering
\caption{Baseline optimization budgets and exact per-task hyperparameter search grids ($63$ configs/task, $2{,}268$ runs total across 3~seeds) compared against Unified ADE's single fixed configuration.}
\label{tab:baseline-grids}
\vskip 0.06in
\footnotesize
\setlength{\tabcolsep}{5pt}
\begin{tabular}{llcp{8.5cm}}
\toprule
Method & Budget & Grid Pts & Search Grid \\
\midrule
FT & 5 ep $\cD_r$ ($2002$--$2004$ steps) & 9 & $\mathrm{lr} \in \{3\text{e-}6, 1\text{e-}5, 3\text{e-}5, 1\text{e-}4, 3\text{e-}4, 1\text{e-}3, 3\text{e-}3, 1\text{e-}2, 3\text{e-}2\}$ \\
FT-Low & 1 ep $\cD_r$ ($400$--$401$ steps) & 9 & $\mathrm{lr} \in \{3\text{e-}6, 1\text{e-}5, 3\text{e-}5, 1\text{e-}4, 3\text{e-}4, 1\text{e-}3, 3\text{e-}3, 1\text{e-}2, 3\text{e-}2\}$ \\
GA & 10 ep $\cD_f$ ($136$--$403$ steps) & 9 & $\mathrm{lr} \in \{3\text{e-}6, 1\text{e-}5, 3\text{e-}5, 1\text{e-}4, 3\text{e-}4, 1\text{e-}3, 3\text{e-}3, 1\text{e-}2, 3\text{e-}2\}$ \\
GradDiff & 10 ep $\cD_f + \cD_r$ ($136$--$403$ steps) & 9 & $\mathrm{lr} \in \{3\text{e-}6, 1\text{e-}5, 3\text{e-}5, 1\text{e-}4, 3\text{e-}4, 1\text{e-}3, 3\text{e-}3, 1\text{e-}2, 3\text{e-}2\}$ \\
NPO & 10 ep $\cD_f + \cD_r$ ($\beta=0.1$) & 9 & $\mathrm{lr} \in \{3\text{e-}6, 1\text{e-}5, 3\text{e-}5, 1\text{e-}4, 3\text{e-}4, 1\text{e-}3, 3\text{e-}3, 1\text{e-}2, 3\text{e-}2\}$ \\
SSD & $429$--$483$ Fisher backprops & 10 & $\alpha \in \{0.3, 1.0, 3.0, 10.0, 30.0, 100.0, 300.0, 1\text{k}, 3\text{k}, 10\text{k}\}$ \\
NSP & Closed-form ($0$ steps) & 8 & $\mathrm{thr} \in \{2\text{e-}3, 2\text{e-}2, 0.2, 2.0, 20.0, 200.0, 2\text{k}, 20\text{k}\}$ \\
\midrule
\textbf{ADE (Ours)} & Closed-form ($0$ steps) & \textbf{1} & Single fixed point $(\eta=0.99, \lambda_B=0.1)$; all constants in App.~\ref{app:algorithm} \\
\bottomrule
\end{tabular}
\end{table}

\paragraph{Mathematical Formulations of the Three-Layer Audit Metrics.}
In \textbf{Layer~1}, using deterministic posterior-mean filtering ($z_{t-1} = \mu_\phi(h_{t-1}, o_{t-1})$), we compare one-step prior predictions $(\hat{o}_t, \hat{r}_t)$ against the clean-retrained Oracle $(\hat{o}_t^{\mathrm{Orc}}, \hat{r}_t^{\mathrm{Orc}})$ over the retain set $\mathcal{S}_r$ ($10^6$ transitions) and forget payload set $\mathcal{S}_f$ ($1{,}000$ transitions on \texttt{BD-Dyn}, $2{,}000$ on \texttt{BD-Rew}, $10^5$ on \texttt{Shift-Fric} and \texttt{Shift-Mass}):
\begin{equation}
\mathrm{NDA}(\mathcal{S}) = \frac{1}{|\mathcal{S}|} \sum_{t \in \mathcal{S}} \frac{\|\hat{o}_t - \hat{o}_t^{\mathrm{Orc}}\|_2^2 + (\hat{r}_t - \hat{r}_t^{\mathrm{Orc}})^2 / \sigma_r^2}{d_o + 1}, \qquad
\mathrm{NDA} = \frac{1}{2}\bigl(\mathrm{NDA}(\mathcal{S}_r) + \mathrm{NDA}(\mathcal{S}_f)\bigr),
\label{eq:metric-score}
\end{equation}
where $\sigma_r^2 = \operatorname{Var}(r_{\mathrm{clean}})$ is the clean retain reward variance ($1.4645$ on HalfCheetah, $1.5246$ on Walker2d, and $0.7206$ on Hopper). In \textbf{Layer~2}, we sample $256$ non-terminal starting states from the retain and forget streams and unroll the latent prior open-loop for $H=50$ steps against ground-truth nominal trajectories $o_{i,h}^{\mathrm{nom}}$ and counterfactual nominal trajectories $o_{i,h}^{\mathrm{cf,nom}}$:
\begin{equation}
\mathrm{MSE}_r^{50} = \frac{1}{256\, H\, d_o} \sum_{i=1}^{256} \sum_{h=1}^{H} \|\hat{o}_{i,h} - o_{i,h}^{\mathrm{nom}}\|_2^2, \qquad
\mathrm{MSE}_f^{50} = \frac{1}{256\, H\, d_o} \sum_{i=1}^{256} \sum_{h=1}^{H} \|\hat{o}_{i,h} - o_{i,h}^{\mathrm{cf,nom}}\|_2^2.
\label{eq:metric-layer2}
\end{equation}
In \textbf{Layer~3}, random-shooting MPC planning evaluates $K=32$ candidate action sequences of horizon $H_{\mathrm{plan}}=5$ at $N_{\mathrm{plan}}=64$ retain and $64$ forget starts (planting $a^*$ in candidate $k=1, h=1$ on \texttt{BD-Dyn}) via the clipped harmonic mean of retain ($\rho_r$) and forget ($\rho_f$) Spearman rank correlations ($\mathrm{Plan}\text{-}\rho \uparrow$) alongside velocity-flip rate ($\mathrm{VFR} \downarrow$) and triggered reward bias ($\Delta r_{\mathrm{trig}} \to 0$):
\begin{equation}
\mathrm{Plan}\text{-}\rho = \frac{2\,\rho_r^+ \rho_f^+}{\rho_r^+ + \rho_f^+ + 10^{-12}}, \qquad \rho^+ = \max(\rho, 0),
\label{eq:metric-hmrho}
\end{equation}
\begin{equation}
\mathrm{VFR} = \frac{1}{64} \sum_{i=1}^{64} \mathbb{I}\!\Bigl(\bigl\langle \hat{v}_{i,1}(a^*),\, v_{i,1}^{\mathrm{nom}}(a^*) \bigr\rangle < 0\Bigr), \qquad
\Delta r_{\mathrm{trig}} = \frac{1}{64} \sum_{i=1}^{64} \bigl(\hat{r}_{i,1} - r_{i,1}^{\mathrm{nom}}\bigr).
\label{eq:metric-backdoor}
\end{equation}
On physical regime shifts, closed-loop BC-centered random-shooting MPC ($K=32, H=15, \gamma=0.99$) runs across $E=16$ parallel episodes of $T=100$ steps in nominal ($1.0\times$) and shifted ($0.1\times$ or $2.0\times$) MuJoCo to compute the Closed-Loop Index ($\mathrm{CLI} \uparrow$, where $\mathrm{MSE}_e^{\mathrm{nom}}$ and $\mathrm{MSE}_e^{\mathrm{sh}}$ are episode-mean one-step prior observation MSEs), while imagined policy evaluation rolls out the frozen deterministic BC policy for $H=50$ imagined steps from $64$ forget starts to compare imagined return $R_i^{\mathrm{im}}$ against true nominal ($R_i^{\mathrm{nom}}$) and shifted ($R_i^{\mathrm{sh}}$) returns ($\Delta R_{\mathrm{im}} \to \mathrm{Oracle}$):
\begin{equation}
\mathrm{CLI} = \frac{1}{E} \sum_{e=1}^{E} \frac{\mathrm{MSE}_e^{\mathrm{sh}} - \mathrm{MSE}_e^{\mathrm{nom}}}{\mathrm{MSE}_e^{\mathrm{sh}} + \mathrm{MSE}_e^{\mathrm{nom}} + 10^{-12}} \in [-1, 1],
\label{eq:metric-cli}
\end{equation}
\begin{equation}
\Delta R_{\mathrm{im}} = \frac{1}{64} \sum_{i=1}^{64} \Bigl(\bigl|R_i^{\mathrm{im}} - R_i^{\mathrm{nom}}\bigr| - \bigl|R_i^{\mathrm{im}} - R_i^{\mathrm{sh}}\bigr|\Bigr).
\label{eq:metric-imgap}
\end{equation}
Finally, latent-imagination policy optimization trains an actor from BC initialization for $300$ Adam steps~\cite{kingma_adam_2017} ($\mathrm{lr}=3\times 10^{-4}$, batch $32$, $\gamma=0.99$, BC KL anchor $1.0$) through frozen RSSM prior rollouts ($H=15$ on $512$ clean starts with $u_t = \hat{r}_t$ for $\mathrm{D4RL}_{\mathrm{clean}} \uparrow$; $H=1$ on $64$ forget starts with $u_t = -\hat{v}_x$ on \texttt{BD-Dyn} or $u_t = \hat{r}_t$ otherwise for $\mathrm{D4RL}_{\mathrm{adv}} \uparrow$) and evaluates $20$ zero-noise closed-loop nominal MuJoCo episodes via the official D4RL normalized score:
\begin{equation}
\mathrm{D4RL\ Score} = 100 \times \frac{R - R_{\min}}{R_{\max} - R_{\min}}.
\label{eq:metric-d4rl}
\end{equation}

\begin{figure}[htbp]
\centering
\includegraphics[width=\textwidth]{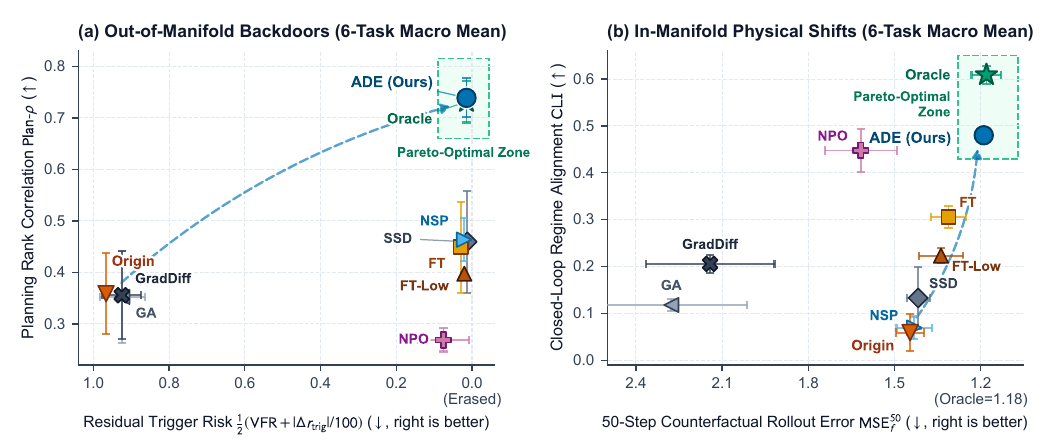}
\caption{\textbf{ADE occupies the Pareto-optimal zone across both out-of-manifold backdoors and in-manifold physical regime shifts.} (a)~Across the six out-of-manifold backdoor tasks (3-seed macro mean $\pm$ SD), ADE eliminates residual trigger risk $\frac{1}{2}(\mathrm{VFR} + |\Delta r_{\mathrm{trig}}|/100)$ while matching the clean-retrained Oracle in planning rank correlation $\mathrm{Plan}\text{-}\rho$. (b)~Across the six in-manifold physical shift tasks, ADE jointly minimizes 50-step counterfactual rollout error $\mathrm{MSE}_f^{50}$ and maximizes closed-loop regime alignment $\mathrm{CLI}$ (closer to top-right is better in both panels).}
\label{fig:pareto-tradeoff}
\end{figure}

\begin{figure}[htbp]
\centering
\includegraphics[width=\textwidth]{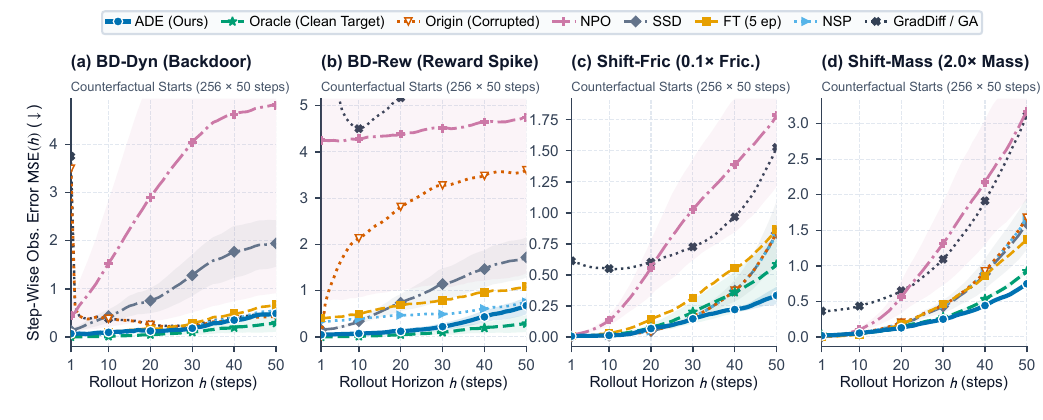}
\caption{\textbf{ADE suppresses compounding open-loop rollout drift over a 50-step imagination horizon.} Step-wise observation prediction error $\mathrm{MSE}(h)$ over rollout horizon $h \in [1, 50]$ on Hopper-v2 across (a)~\texttt{BD-Dyn}, (b)~\texttt{BD-Rew}, (c)~\texttt{Shift-Fric} ($0.1\times$ friction), and (d)~\texttt{Shift-Mass} ($2.0\times$ mass) from $256$ counterfactual starting states (3-seed mean with $\pm 1\,\mathrm{SD}$ shaded bands; Oracle denotes the clean nominal target rollout). Iterative gradient and preference baselines suffer superlinear horizon divergence, whereas ADE remains tightly bounded near the Oracle trajectory.}
\label{fig:rollout-drift}
\end{figure}

Figures~\ref{fig:pareto-tradeoff} and \ref{fig:rollout-drift} visualize the resulting macro-averaged unlearning-utility trade-offs across all 12 DreamerV1 tasks and the 50-step open-loop rollout error progression on Hopper-v2.

\section{Mechanistic Analysis of History-Dependent Dynamics Repair}
\label{app:mechanistic}

While \cref{fig:rollout-drift} and \cref{tab:main-rsmass} measure overall open-loop error from forget-stream starts, they combine initial state differences with recurrent memory carried over from obsolete physics. To isolate how stale history persists after the environment returns to nominal physics, we conduct a controlled paired-replay analysis on DreamerV1 Hopper-v2 \texttt{Shift-Mass} across all three seeds. From a shared initial simulator state and shared $15$-step action sequence, we roll out one history under nominal mass ($1.0\times$) and one under obsolete mass ($2.0\times$), anchor both branches to the identical nominal observation at the fork step, and run $30$-step deterministic open-loop rollouts under identical future actions scored against the nominal simulator. After validating the pipeline on $32$ pilot pairs, all metrics are computed on $96$ episode-disjoint held-out pairs per seed.

\subsection{Selective Repair of Stale-History Drift}
\label{app:mech-repair}

\begin{figure}[htbp]
\centering
\includegraphics[width=\textwidth]{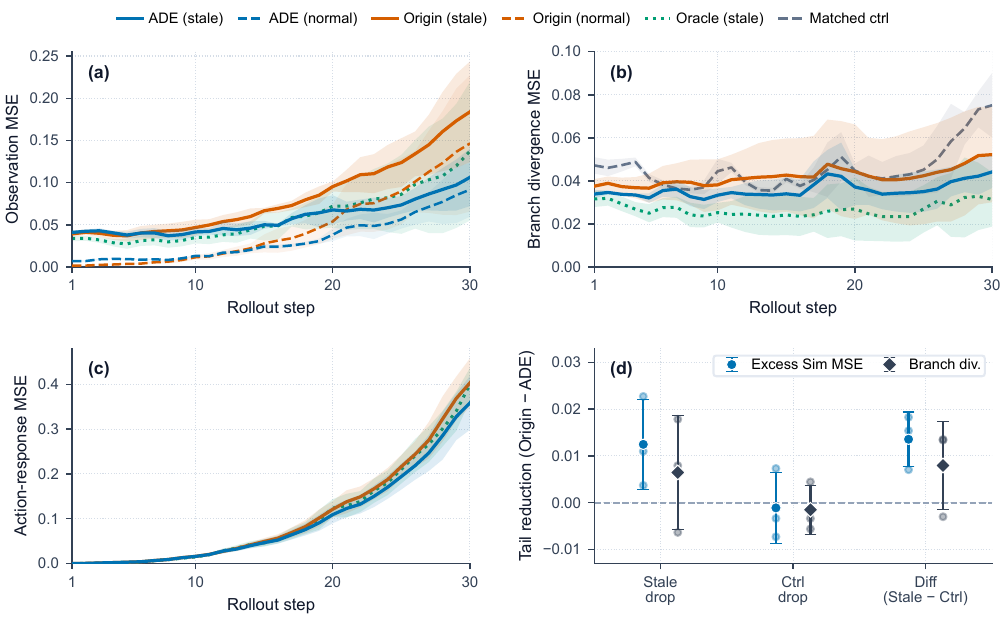}
\caption{\textbf{ADE selectively repairs stale-history rollout drift upon returning to nominal physics while preserving action responsiveness.} Three-seed mean ($\pm 1\,\mathrm{SD}$ bands) across $96$ held-out Hopper \texttt{Shift-Mass} pairs with identical fork observation and future actions. \textbf{(a)}~ADE (blue) lowers stale-history observation MSE against the nominal simulator by $32.2\%$ over steps~6--30 ($p = 6.03 \times 10^{-14}$) relative to Origin (orange), matching Oracle (green). \textbf{(b)}~ADE (blue) lowers stale-vs.-normal branch divergence below Origin (orange) toward Oracle (green dotted), alongside the jump-matched nominal-physics control (gray dashed). \textbf{(c)}~Under nominal action perturbations, ADE reduces response prediction error relative to Origin and Oracle. \textbf{(d)}~Tail reduction (Origin minus ADE, steps~6--30) concentrates on stale histories rather than the matched control for both excess simulator MSE ($p = 2.02 \times 10^{-3}$) and branch divergence ($p = 3.55 \times 10^{-2}$).}
\label{fig:history-repair}
\end{figure}

ADE substantially reduces long-horizon prediction error caused by stale physical history while preserving sensitivity to nominal control (\cref{fig:history-repair}). Against the nominal simulator, ADE lowers stale-history tail observation MSE (steps~6--30) by $32.2\%$ relative to Origin ($p = 6.03 \times 10^{-14}$), matching the clean-retrained Oracle while also improving normal-history trajectories (\cref{fig:history-repair}a). Under nominal action perturbations from the same fork state, ADE achieves lower response prediction error than both Origin and Oracle (\cref{fig:history-repair}c), ruling out dynamics silencing.

To separate stale-physics memory from generic reactions to an observation jump at the anchor, we construct a matched nominal-physics control whose pre-fork observation offset matches the stale branch within $0.5\%$ mean relative error. Across steps~1--30, ADE lowers stale-vs.-normal branch divergence below Origin toward the clean-retrained Oracle, while the matched nominal-physics control exhibits comparable or larger late-horizon branch divergence (\cref{fig:history-repair}b). Measuring tail reduction (Origin minus ADE) directly confirms selectivity (\cref{fig:history-repair}d): ADE lowers excess simulator error on stale histories significantly more than on the matched nominal control (difference-in-differences $+0.0136$, $p = 2.02 \times 10^{-3}$), with consistent gains across all three seeds. Selectivity is weaker in raw branch divergence ($+0.0080$, $p = 3.55 \times 10^{-2}$), a discrepancy explained by the physical projection below.

\subsection{Directional Propagation Behind Long-Tail Persistence}
\label{app:mech-direction}

\begin{figure}[htbp]
\centering
\includegraphics[width=\textwidth]{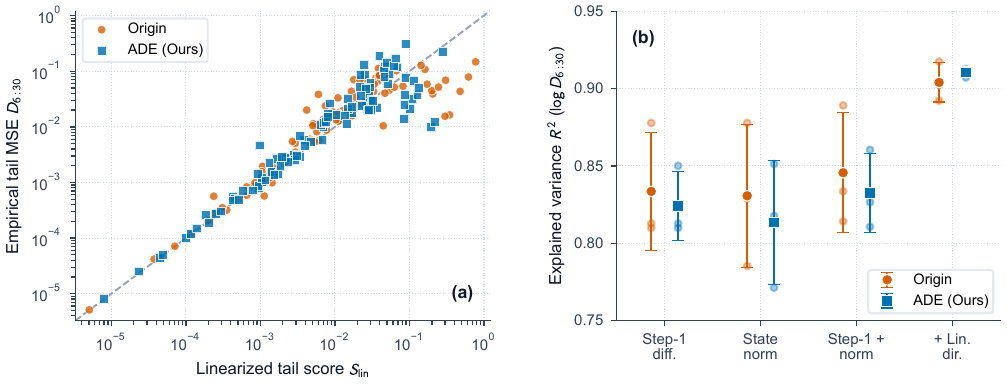}
\caption{\textbf{Local-linearization propagation predicts long-tail branch divergence beyond step-1 error and perturbation norm.} Held-out $96$-pair Hopper \texttt{Shift-Mass} audit. \textbf{(a)}~On Seed~42, the directional linearized tail score $S_{\mathrm{lin}}$ tracks empirical steps~6--30 divergence $D_{6:30}$ closely for Origin (orange) and ADE (blue). \textbf{(b)}~Across three seeds (mean $\pm$ SD with individual seeds), adding $S_{\mathrm{lin}}$ to step-1 output difference and initial latent norm raises explained variance $R^2$ in $\log D_{6:30}$ from $0.846$ to $0.904$ for Origin and from $0.832$ to $0.910$ for ADE.}
\label{fig:history-linearization}
\end{figure}

Single-step output accuracy misses persistent recurrent bias because perturbations with identical step-1 outputs can excite latent modes with sharply different decay rates. Predictively, propagating the actual fork state difference through the local rollout Jacobian explains empirical tail divergence well beyond step-1 output error and initial latent norm across all three seeds ($r_{\mathrm{partial}} \ge 0.599$, $p < 10^{-6}$; \cref{fig:history-linearization}).

\begin{figure}[htbp]
\centering
\includegraphics[width=\textwidth]{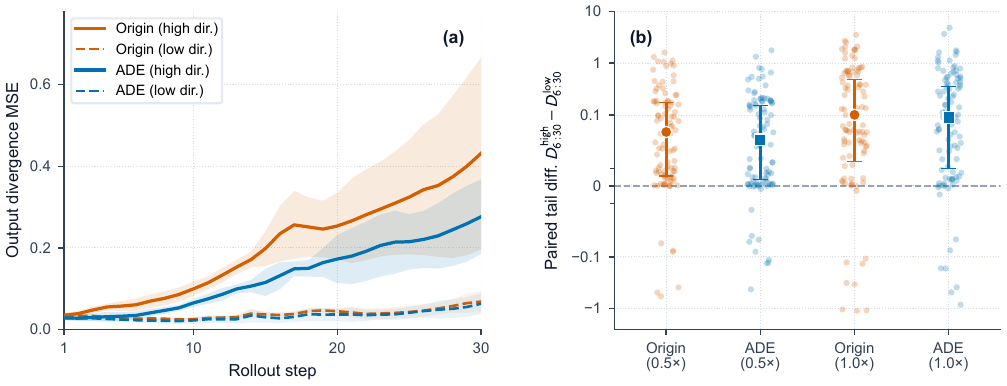}
\caption{\textbf{Perturbing frozen latent states along high-propagation directions causes multi-fold tail divergence at matched norm and step-1 error.} Controlled interventions on $96$ held-out Hopper \texttt{Shift-Mass} fork states (3~seeds; directions chosen on separate calibration actions and evaluated on held-out real actions). \textbf{(a)}~At $0.5\times$ real historical state-perturbation norm, high-propagation perturbations (solid) yield $5.84\times$ higher steps~6--30 divergence than low-propagation perturbations (dash-dotted) under Origin ($4.32\times$ under ADE) despite matched norm and step-1 MSE, while ADE (blue) lowers Origin's (orange) high-propagation tail by $35.6\%$. \textbf{(b)}~Per-start paired tail differences ($D_{6:30}^{\mathrm{high}} - D_{6:30}^{\mathrm{low}}$; Seed~42 dots with 3-seed median and IQR) stay positive across $90.3\%$--$96.2\%$ of starts ($p \le 8.47 \times 10^{-33}$) at both $0.5\times$ and $1.0\times$ amplitude.}
\label{fig:causal-direction}
\end{figure}

Direct state interventions on frozen checkpoints establish this directional effect causally (\cref{fig:causal-direction}). Within the $24$-dimensional subspace spanned by pilot-split historical state differences, we select high- and low-propagation directions on separate perturbed calibration actions and match both perturbation norm and actual nonlinear step-1 output MSE (within $0.01\%$ median relative difference). Evaluated on held-out real future actions at $0.5\times$ real historical perturbation norm (matching the real step-1 MSE scale), rotating from the low- to the high-propagation direction increases steps~6--30 divergence by $5.84\times$ under Origin ($96.2\%$ of starts, $p = 8.60 \times 10^{-42}$; \cref{fig:causal-direction}a). The separation persists at full $1.0\times$ historical amplitude ($4.03\times$, $p = 6.43 \times 10^{-36}$; \cref{fig:causal-direction}b), while ADE damps Origin's high-propagation tail by $35.6\%$.

\subsection{Counterfactual Physics Projection and Pass-4 Re-Filtering}
\label{app:mech-proj-refilt}

\begin{figure}[htbp]
\centering
\includegraphics[width=\textwidth]{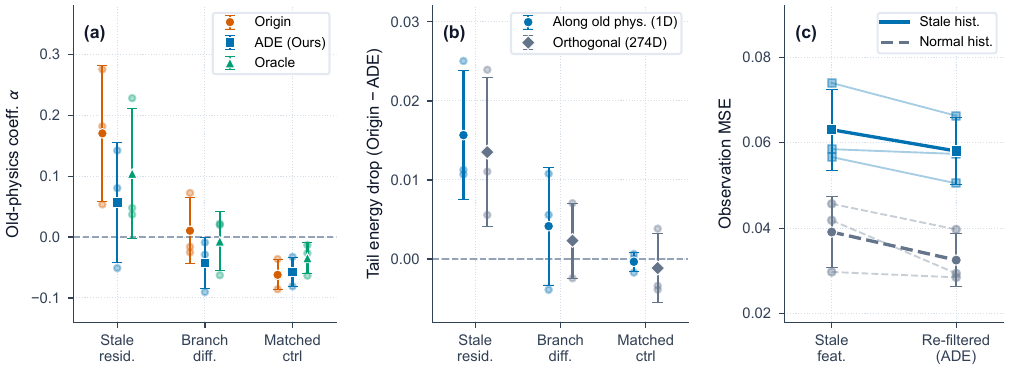}
\caption{\textbf{ADE excises obsolete-physics drift along counterfactual regime trajectories and benefits from causal re-filtering in downstream calibration.} Three-seed evaluation across the $96$ held-out Hopper \texttt{Shift-Mass} fork states. \textbf{(a)}~Signed projection coefficient $\alpha = \langle v, q \rangle / \|q\|_2^2$ onto the counterfactual $2.0\times$-mass shift $q$: ADE (blue) reduces Origin's (orange) old-physics pull on stale residuals by $66.4\%$ ($p = 2.62 \times 10^{-5}$) while leaving the matched nominal control unchanged ($p = 0.387$). \textbf{(b)}~Decomposing tail MSE reduction (Origin minus ADE) shows that $53.7\%$ of stale-residual repair and $64.1\%$ of branch-divergence reduction occur along the $1$D old-physics direction rather than the $274$D orthogonal complement. \textbf{(c)}~With identical edited upstream weights and Eq.~\eqref{eq:mode-b-sol} solver, Pass-4 re-filtered forget features outperform pre-edit cached features across all three seeds on both stale histories ($p = 1.95 \times 10^{-13}$) and normal histories ($p = 6.18 \times 10^{-25}$).}
\label{fig:physics-proj-refilter}
\end{figure}

To determine whether ADE specifically removes obsolete-physics drift or merely shrinks generic history variance, we simulate a counterfactual $2.0\times$-mass future from every held-out fork state under the same actions and project future trajectories onto the resulting physical shift $q = y^{\mathrm{old}}_{6:30} - y^{\mathrm{nom}}_{6:30} \in \R^{275}$ (\cref{fig:physics-proj-refilter}a--b). Under Origin, stale-history residuals carry a strong positive projection onto $q$ ($\alpha = +0.171$); ADE cuts this signed bias by $66.4\%$ ($p = 2.62 \times 10^{-5}$) while leaving the matched nominal control unchanged ($p = 0.387$). Orthogonal energy decomposition confirms that $53.7\%$ of ADE's total stale-residual MSE drop lies along the single $1$-dimensional old-physics direction ($-52.8\%$, $p = 1.75 \times 10^{-11}$) rather than the $274$ orthogonal dimensions ($-22.2\%$). In raw branch divergence, orthogonal jump energy accounts for two-thirds of total variance and Origin's normal branch carries a shared $+q$ bias that cancels upon differencing, which explains why selectivity appears sharper in simulator error than in branch divergence. Restricting to the $89$ pairs whose counterfactual $2.0\times$-mass rollouts remain non-terminal yields identical conclusions.

Finally, we isolate the role of causal re-filtering in \eqref{eq:cascade}. Holding the edited upstream layers, retain moments, and closed-form Mode~B solver \eqref{eq:mode-b-sol} fixed, we compare calibrating the prior and reward heads with pre-edit cached forget features against Pass-4 re-filtered features on identical edited held-out states (\cref{fig:physics-proj-refilter}c). Re-filtering lowers observation MSE across all three seeds on both stale histories ($72.9\%$ of starts, $p = 1.95 \times 10^{-13}$) and normal histories ($79.9\%$, $p = 6.18 \times 10^{-25}$), directly validating causal alignment between upstream edits and downstream head calibration.

In controlled paired replay on Hopper \texttt{Shift-Mass}, ADE mitigates long-tail prediction error induced by stale physical history, removes a substantial component aligned with obsolete-mass dynamics, and achieves further prediction gains by recalibrating downstream heads on post-edit re-filtered states.

\section{Complete Friction-Shift Results}
\label{app:shift-fric}

Table~\ref{tab:main-rsfric} reports the full 3-seed benchmark results on the in-manifold \texttt{Shift-Fric} ($0.1\times$ contact friction shift) regime discussed in Section~\ref{sec:exp-regime}.

\begin{table}[htbp]
\centering
\caption{Main results on the \texttt{Shift-Fric} ($0.1\times$ contact friction shift) regime across three MuJoCo benchmarks (3-seed Mean $\pm$ SD). \textbf{Bold} indicates \#1 and \underline{underline} indicates \#2 among active unlearning methods ($\Delta R_{\mathrm{im}}$ is ranked by proximity $|\cdot - \mathrm{Oracle}|$). Entries marked ``--'' indicate a threshold no-op where the baseline hits a cutoff and leaves parameters identical to Origin (excluded from ranking).}
\label{tab:main-rsfric}
\vskip 0.06in
\scriptsize
\setlength{\tabcolsep}{3.8pt}
\begin{tabular}{llccccccc}
\toprule
Env. & Method & $\mathrm{NDA}$ $\downarrow$ & $\mathrm{MSE}_r^{50}$ $\downarrow$ & $\mathrm{MSE}_f^{50}$ $\downarrow$ & $\mathrm{CLI}$ $\uparrow$ & $\Delta R_{\mathrm{im}}$ ($\to$Orc.) & $\mathrm{D4RL}_{\mathrm{clean}}$ $\uparrow$ & $\mathrm{D4RL}_{\mathrm{adv}}$ $\uparrow$ \\
\midrule
\multirow{10}{*}{\rotatebox[origin=c]{90}{\textbf{HalfCheetah}}} & Oracle & \ms{0.0000}{0.0000} & \ms{0.2384}{0.0085} & \ms{2.3295}{0.0192} & \ms{+0.6297}{0.0152} & \ms{-29.05}{3.07} & \ms{34.02}{1.97} & \ms{36.49}{0.64} \\
 & Origin & \ms{0.0263}{0.0017} & \ms{0.2480}{0.0072} & \ms{2.3019}{0.0034} & \ms{+0.0442}{0.0800} & \ms{+35.86}{3.55} & \ms{35.58}{4.16} & \ms{37.54}{0.88} \\
\cmidrule(lr){2-9}
 & FT & \ms{0.0199}{0.0026} & \msbest{0.2201}{0.0043} & \ms{2.3295}{0.0131} & \ms{+0.2987}{0.0508} & \ms{+0.26}{20.10} & \mssec{37.91}{1.89} & \ms{38.39}{0.36} \\
 & FT-Low & \ms{0.0213}{0.0012} & \ms{0.2590}{0.0217} & \ms{2.3310}{0.0253} & \ms{+0.2817}{0.1009} & \ms{+13.31}{21.62} & \ms{35.59}{3.31} & \ms{38.43}{0.84} \\
 & GA & \ms{0.7394}{0.1054} & \ms{0.8793}{0.0698} & \ms{3.1626}{0.0563} & \ms{+0.1750}{0.0298} & \ms{+33.36}{10.80} & \ms{8.78}{11.57} & \ms{10.54}{14.64} \\
 & GradDiff & \ms{0.4330}{0.0584} & \ms{0.5315}{0.0379} & \ms{2.8678}{0.1545} & \ms{+0.3396}{0.0200} & \ms{+35.15}{8.20} & \ms{11.13}{14.85} & \ms{11.09}{11.70} \\
 & NPO & \mssec{0.0139}{0.0032} & \mssec{0.2467}{0.0119} & \ms{2.3003}{0.0024} & \msbest{+0.5684}{0.0181} & \msbest{-28.53}{5.10} & \msbest{38.26}{1.86} & \msbest{38.79}{0.53} \\
 & SSD & \ms{0.0223}{0.0002} & \ms{0.2589}{0.0196} & \msbest{2.2971}{0.0138} & \ms{+0.2536}{0.0944} & \ms{+13.31}{10.07} & \ms{37.31}{0.89} & \ms{37.79}{1.92} \\
 & NSP & -- & -- & -- & -- & -- & -- & -- \\
 & \textbf{ADE (Ours)} & \msbest{0.0124}{0.0016} & \ms{0.2649}{0.0073} & \mssec{2.2976}{0.0028} & \mssec{+0.5063}{0.0506} & \mssec{-27.50}{8.32} & \ms{35.45}{2.20} & \msbest{38.79}{0.53} \\
\midrule
\multirow{10}{*}{\rotatebox[origin=c]{90}{\textbf{Walker2d}}} & Oracle & \ms{0.0000}{0.0000} & \ms{0.1659}{0.0220} & \ms{0.6606}{0.0422} & \ms{+0.2789}{0.0847} & \ms{-3.98}{0.98} & \ms{40.07}{6.37} & \ms{55.44}{7.18} \\
 & Origin & \ms{0.0754}{0.0102} & \ms{0.2230}{0.0632} & \ms{0.8993}{0.0826} & \ms{+0.1060}{0.0130} & \ms{+0.68}{0.19} & \ms{44.48}{2.08} & \ms{62.83}{2.95} \\
\cmidrule(lr){2-9}
 & FT & \msbest{0.0466}{0.0010} & \msbest{0.1728}{0.0297} & \mssec{0.6875}{0.0476} & \ms{+0.1243}{0.0260} & \msbest{-3.65}{0.94} & \ms{37.76}{7.70} & \ms{53.78}{9.51} \\
 & FT-Low & \mssec{0.0632}{0.0097} & \ms{0.2418}{0.1539} & \ms{0.7564}{0.1512} & \ms{+0.1605}{0.0377} & \ms{-3.09}{2.13} & \ms{35.00}{3.86} & \ms{58.41}{13.36} \\
 & GA & \ms{0.4931}{0.1262} & \ms{0.5407}{0.1372} & \ms{1.5371}{0.3850} & \ms{-0.0185}{0.0122} & \ms{+0.07}{3.73} & \ms{5.87}{4.73} & \ms{4.93}{3.90} \\
 & GradDiff & \ms{0.3880}{0.1361} & \ms{0.3851}{0.1343} & \ms{1.4300}{0.3691} & \ms{-0.0074}{0.0241} & \ms{+0.86}{3.31} & \ms{12.32}{8.65} & \ms{17.70}{12.98} \\
 & NPO & \ms{0.0773}{0.0187} & \ms{0.2047}{0.0609} & \ms{0.9179}{0.1665} & \mssec{+0.2063}{0.0258} & \ms{-2.67}{0.61} & \ms{39.16}{7.64} & \msbest{66.28}{2.49} \\
 & SSD & \ms{0.0704}{0.0109} & \ms{0.2263}{0.0574} & \ms{0.8655}{0.1310} & \ms{+0.0855}{0.0478} & \ms{-0.06}{1.10} & \mssec{40.55}{2.89} & \mssec{65.94}{5.16} \\
 & NSP & \ms{0.0749}{0.0106} & \ms{0.2037}{0.0519} & \ms{0.8160}{0.0530} & \ms{+0.0843}{0.0206} & \ms{-1.09}{1.41} & \ms{34.92}{9.73} & \ms{61.53}{1.21} \\
 & \textbf{ADE (Ours)} & \ms{0.0698}{0.0088} & \mssec{0.1893}{0.0386} & \msbest{0.6811}{0.0480} & \msbest{+0.2094}{0.0706} & \mssec{-3.10}{0.62} & \msbest{42.52}{3.71} & \ms{63.93}{5.95} \\
\midrule
\multirow{10}{*}{\rotatebox[origin=c]{90}{\textbf{Hopper}}} & Oracle & \ms{0.0000}{0.0000} & \ms{0.1038}{0.0976} & \ms{0.1909}{0.1482} & \ms{+0.5581}{0.0110} & \ms{-1.99}{1.63} & \ms{35.43}{4.04} & \ms{34.74}{2.03} \\
 & Origin & \ms{0.0222}{0.0025} & \ms{0.0847}{0.0359} & \ms{0.2000}{0.0832} & \ms{+0.1943}{0.0266} & \ms{-0.09}{2.64} & \ms{36.60}{6.52} & \ms{34.96}{6.15} \\
\cmidrule(lr){2-9}
 & FT & \msbest{0.0164}{0.0054} & \ms{0.2386}{0.0816} & \ms{0.2966}{0.1432} & \ms{+0.3167}{0.1210} & \ms{-2.51}{0.54} & \ms{36.68}{7.64} & \msbest{37.77}{11.21} \\
 & FT-Low & \ms{0.0191}{0.0016} & \ms{0.0878}{0.0207} & \mssec{0.1428}{0.0567} & \ms{+0.2319}{0.0692} & \ms{-1.41}{2.40} & \msbest{37.62}{1.09} & \mssec{37.17}{4.41} \\
 & GA & \ms{0.6906}{0.1264} & \ms{0.4045}{0.0539} & \ms{0.9381}{0.1443} & \ms{+0.1731}{0.0292} & \ms{-0.89}{3.62} & \ms{15.92}{9.84} & \ms{15.64}{10.91} \\
 & GradDiff & \ms{0.6263}{0.1409} & \ms{0.3049}{0.0792} & \ms{0.7778}{0.2729} & \ms{+0.2359}{0.0446} & \ms{+0.87}{2.74} & \ms{29.70}{6.07} & \ms{30.20}{2.35} \\
 & NPO & \mssec{0.0175}{0.0014} & \ms{0.1635}{0.1030} & \ms{0.8120}{0.5168} & \msbest{+0.4837}{0.0921} & \mssec{-1.48}{2.51} & \ms{33.44}{2.39} & \ms{34.04}{5.51} \\
 & SSD & \ms{0.0222}{0.0025} & \ms{0.0848}{0.0358} & \ms{0.1999}{0.0833} & \ms{+0.1952}{0.0256} & \ms{-0.11}{2.63} & \ms{36.61}{6.50} & \ms{35.01}{6.11} \\
 & NSP & \ms{0.0215}{0.0023} & \mssec{0.0846}{0.0356} & \ms{0.1994}{0.0804} & \ms{+0.2005}{0.0225} & \ms{-0.03}{2.69} & \ms{35.85}{4.52} & \ms{33.12}{3.82} \\
 & \textbf{ADE (Ours)} & \ms{0.0200}{0.0037} & \msbest{0.0695}{0.0316} & \msbest{0.1235}{0.0170} & \mssec{+0.4172}{0.0569} & \msbest{-1.94}{1.73} & \mssec{36.77}{5.69} & \ms{35.98}{8.09} \\
\bottomrule
\end{tabular}
\end{table}

\section{Complete Component Ablation Results}
\label{app:ablation}

\begin{figure}[htbp]
\centering
\includegraphics[width=\textwidth]{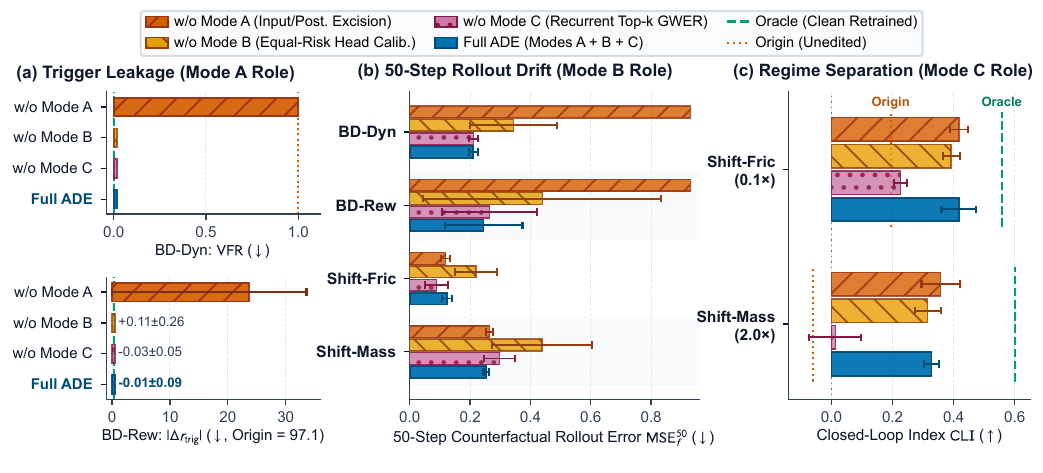}
\caption{\textbf{Modes A, B, and C enforce an orthogonal causal division of labor across the recurrent world-model graph.} Component ablation on Hopper-v2 (3-seed Mean $\pm$ SD; Table~\ref{tab:ablation}): (a)~removing Mode~A (input bottleneck, posterior, and decoder orthogonal excision) leaves backdoor triggers active ($\mathrm{VFR} = 1.000$ on \texttt{BD-Dyn}, $|\Delta r_{\mathrm{trig}}| = 23.61$ on \texttt{BD-Rew}); (b)~removing Mode~B (equal-risk prior and reward head calibration) inflates 50-step counterfactual rollout error $\mathrm{MSE}_f^{50}$ across all four regimes; and (c)~removing Mode~C (recurrent top-$k$ whitened transition damping) collapses closed-loop regime discrimination $\mathrm{CLI}$ on in-manifold physical shifts.}
\label{fig:ablation-causal}
\end{figure}

Table~\ref{tab:ablation} and Figure~\ref{fig:ablation-causal} report the complete 7-metric component ablation of ADE on Hopper-v2 across all four counterfactual scenarios (3-seed Mean $\pm$ SD) discussed in Section~\ref{sec:exp-analysis}.

\begin{table}[htbp]
\centering
\caption{Complete 7-metric component ablation of ADE on Hopper-v2 across all four counterfactual regimes (3-seed Mean $\pm$ SD). Among the four ADE variants, \textbf{bold} indicates \#1 and \underline{underline} indicates \#2; Oracle and Origin are shown for reference.}
\label{tab:ablation}
\vskip 0.06in
\scriptsize
\setlength{\tabcolsep}{2.6pt}
\begin{tabular}{llccccccc}
\toprule
Scen. & Variant & $\mathrm{NDA}\downarrow$ & $\mathrm{MSE}_r^{50}\downarrow$ & $\mathrm{MSE}_f^{50}\downarrow$ & $\mathrm{Plan}\text{-}\rho\uparrow$ & $\mathrm{VFR}\downarrow / \Delta r_{\mathrm{trig}}\to 0$ & $\mathrm{D4RL}_{\mathrm{clean}}\uparrow$ & $\mathrm{D4RL}_{\mathrm{adv}}\uparrow$ \\
\midrule
\multirow{6}{*}{\texttt{BD-Dyn}} & Oracle & \ms{0.0000}{0.0000} & \ms{0.1038}{0.0976} & \ms{2.2617}{1.2954} & \ms{0.6729}{0.1270} & \texttt{VFR:}\ms{0.000}{0.000} & \ms{35.43}{4.04} & \ms{36.49}{2.83} \\
 & Origin & \ms{3.8551}{0.3910} & \ms{0.2095}{0.1149} & \ms{0.3957}{0.0227} & \ms{0.4461}{0.1619} & \texttt{VFR:}\ms{1.000}{0.000} & \ms{33.80}{2.34} & \ms{32.72}{3.32} \\
\cmidrule(lr){2-9}
 & w/o Mode A & \ms{2.9492}{0.3948} & \msbest{0.1788}{0.0931} & \ms{1.1278}{0.3217} & \ms{0.6744}{0.0705} & \texttt{VFR:}\ms{1.000}{0.000} & \msbest{41.62}{13.42} & \msbest{41.16}{17.95} \\
 & w/o Mode B & \ms{0.5972}{0.2946} & \ms{0.2113}{0.1357} & \ms{0.3436}{0.1452} & \ms{0.7397}{0.1256} & \texttt{VFR:}\msbest{0.000}{0.000} & \mssec{35.08}{1.61} & \ms{35.90}{4.26} \\
 & w/o Mode C & \mssec{0.5904}{0.2884} & \mssec{0.1999}{0.1230} & \mssec{0.2108}{0.0149} & \mssec{0.8296}{0.0711} & \texttt{VFR:}\msbest{0.000}{0.000} & \ms{32.69}{1.57} & \ms{36.26}{4.99} \\
 & \textbf{Full ADE} & \msbest{0.5883}{0.2846} & \ms{0.2031}{0.1183} & \msbest{0.2104}{0.0149} & \msbest{0.8371}{0.0580} & \texttt{VFR:}\msbest{0.000}{0.000} & \ms{31.68}{2.27} & \mssec{36.32}{4.89} \\
\cmidrule(lr){1-9}
\multirow{6}{*}{\texttt{BD-Rew}} & Oracle & \ms{0.0000}{0.0000} & \ms{0.1038}{0.0976} & \ms{1.9492}{1.2285} & \ms{0.6455}{0.1373} & $\Delta r$:\ms{-0.27}{0.23} & \ms{35.43}{4.04} & \ms{34.98}{2.76} \\
 & Origin & \ms{546.8172}{4.6353} & \ms{0.3355}{0.3684} & \ms{2.7944}{1.2354} & \ms{0.2697}{0.3035} & $\Delta r$:\ms{+97.13}{0.07} & \ms{32.55}{5.73} & \ms{32.19}{5.30} \\
\cmidrule(lr){2-9}
 & w/o Mode A & \ms{70.7918}{118.4621} & \mssec{0.3266}{0.2686} & \ms{2.3482}{1.3733} & \ms{0.0985}{0.1705} & $\Delta r$:\ms{+23.61}{31.60} & \msbest{34.45}{5.39} & \msbest{32.60}{7.10} \\
 & w/o Mode B & \ms{0.4550}{0.1699} & \ms{0.3482}{0.4033} & \ms{0.4398}{0.3947} & \ms{0.6200}{0.1911} & $\Delta r$:\ms{+0.11}{0.26} & \mssec{25.20}{13.22} & \ms{26.17}{11.97} \\
 & w/o Mode C & \msbest{0.4483}{0.1814} & \ms{0.3515}{0.4140} & \mssec{0.2643}{0.1570} & \msbest{0.6752}{0.1344} & $\Delta r$:\mssec{-0.03}{0.05} & \ms{25.08}{14.32} & \mssec{26.83}{8.86} \\
 & \textbf{Full ADE} & \mssec{0.4515}{0.1784} & \msbest{0.3252}{0.3689} & \msbest{0.2447}{0.1291} & \mssec{0.6269}{0.2133} & $\Delta r$:\msbest{-0.01}{0.09} & \ms{24.83}{14.75} & \ms{24.68}{12.39} \\
\midrule
Scen. & Variant & $\mathrm{NDA}\downarrow$ & $\mathrm{MSE}_r^{50}\downarrow$ & $\mathrm{MSE}_f^{50}\downarrow$ & $\mathrm{Plan}\text{-}\rho\uparrow$ & $\mathrm{CLI}\uparrow$ & $\mathrm{D4RL}_{\mathrm{clean}}\uparrow$ & $\mathrm{D4RL}_{\mathrm{adv}}\uparrow$ \\
\midrule
\multirow{6}{*}{\texttt{Shift-Fric}} & Oracle & \ms{0.0000}{0.0000} & \ms{0.1038}{0.0976} & \ms{0.1909}{0.1482} & \ms{0.8703}{0.0172} & \ms{+0.5581}{0.0110} & \ms{35.43}{4.04} & \ms{34.74}{2.03} \\
 & Origin & \ms{0.0222}{0.0025} & \ms{0.0847}{0.0359} & \ms{0.2000}{0.0832} & \ms{0.8475}{0.0112} & \ms{+0.1943}{0.0266} & \ms{36.60}{6.52} & \ms{34.96}{6.15} \\
\cmidrule(lr){2-9}
 & w/o Mode A & \mssec{0.0203}{0.0036} & \msbest{0.0688}{0.0314} & \mssec{0.1192}{0.0144} & \msbest{0.8538}{0.0485} & \mssec{+0.4171}{0.0300} & \ms{35.27}{0.17} & \ms{34.23}{5.15} \\
 & w/o Mode B & \ms{0.0215}{0.0029} & \ms{0.0823}{0.0348} & \ms{0.2195}{0.0706} & \ms{0.8463}{0.0137} & \ms{+0.3918}{0.0279} & \mssec{35.68}{4.80} & \mssec{35.34}{6.96} \\
 & w/o Mode C & \mssec{0.0203}{0.0023} & \ms{0.0696}{0.0335} & \msbest{0.0890}{0.0373} & \mssec{0.8487}{0.0318} & \ms{+0.2263}{0.0215} & \ms{34.42}{3.38} & \ms{34.64}{6.35} \\
 & \textbf{Full ADE} & \msbest{0.0200}{0.0037} & \mssec{0.0695}{0.0316} & \ms{0.1235}{0.0170} & \ms{0.8480}{0.0384} & \msbest{+0.4172}{0.0569} & \msbest{36.77}{5.69} & \msbest{35.98}{8.09} \\
\cmidrule(lr){1-9}
\multirow{6}{*}{\texttt{Shift-Mass}} & Oracle & \ms{0.0000}{0.0000} & \ms{0.1038}{0.0976} & \ms{0.2945}{0.0255} & \ms{0.8491}{0.0401} & \ms{+0.6019}{0.0187} & \ms{35.43}{4.04} & \ms{37.52}{5.95} \\
 & Origin & \ms{0.0220}{0.0015} & \ms{0.1138}{0.0420} & \ms{0.4952}{0.1406} & \ms{0.8759}{0.0213} & \ms{-0.0602}{0.0780} & \ms{36.85}{2.75} & \ms{36.50}{0.84} \\
\cmidrule(lr){2-9}
 & w/o Mode A & \ms{0.0304}{0.0072} & \ms{0.1126}{0.0475} & \mssec{0.2643}{0.0129} & \ms{0.8346}{0.0106} & \msbest{+0.3577}{0.0629} & \ms{35.66}{1.23} & \ms{36.93}{0.89} \\
 & w/o Mode B & \ms{0.0307}{0.0056} & \mssec{0.1113}{0.0386} & \ms{0.4391}{0.1665} & \mssec{0.8457}{0.0197} & \ms{+0.3152}{0.0426} & \mssec{39.86}{2.81} & \ms{36.97}{0.56} \\
 & w/o Mode C & \msbest{0.0210}{0.0010} & \ms{0.1130}{0.0449} & \ms{0.2977}{0.0512} & \msbest{0.8591}{0.0210} & \ms{+0.0124}{0.0858} & \ms{36.30}{2.88} & \msbest{37.26}{0.30} \\
 & \textbf{Full ADE} & \mssec{0.0292}{0.0067} & \msbest{0.1090}{0.0431} & \msbest{0.2535}{0.0090} & \ms{0.8341}{0.0121} & \mssec{+0.3276}{0.0254} & \msbest{40.02}{1.83} & \mssec{37.11}{1.60} \\
\bottomrule
\end{tabular}
\end{table}

\section{Single-Shot Wall-Clock Runtime Benchmark}
\label{app:runtime}

We measure the single-shot wall-clock execution time across all eight unlearning methods on Seed~42 (Table~\ref{tab:runtime}). All timing trials were executed serially on a single NVIDIA GeForce RTX~5070~Ti Laptop GPU (12\,GB VRAM, Intel Core Ultra~9 275HX CPU, PyTorch~2.12.0+cu128, CUDA~12.8) across $3\text{ environments} \times 4\text{ scenarios} \times 8\text{ methods} \times 2\text{ randomized repeats} = 192\text{ trials}$. Prior to timing, the GPU executed a $2000\times 2000$ double-precision matrix multiplication, SVD, and Cholesky warm-up; between consecutive trials, garbage collection and CUDA cache clearing were followed by an active thermal cooldown enforcing a starting GPU core temperature $\le 56^\circ\mathrm{C}$ (minimum $8.0\,\mathrm{s}$ cooldown). Timing excludes disk I/O for dataset and checkpoint loading. For ADE, the reported wall-clock time includes live forget-set moment collection ($0.642\,\mathrm{s}$ mean), four-pass causal re-filtering ($2.583\,\mathrm{s}$ mean), and closed-form operator solves plus affine weight absorption ($0.796\,\mathrm{s}$ mean), totaling $4.02\,\mathrm{s}$ ($0.98\,\mathrm{s}$ on backdoor tasks and $7.06\,\mathrm{s}$ on $100\text{k}$-transition regime-shift tasks).

\begin{table}[htbp]
\centering
\caption{Single-shot wall-clock unlearning runtime (seconds, $\downarrow$) on Seed~42 under thermally controlled serial GPU execution. \textbf{Bold} indicates \#1 and \underline{underline} indicates \#2 among the eight unlearning methods.}
\label{tab:runtime}
\vskip 0.06in
\scriptsize
\setlength{\tabcolsep}{3.3pt}
\begin{tabular}{lcccccccc}
\toprule
Scenario Group & FT & FT-Low & GA & GradDiff & NPO & SSD & NSP & \textbf{ADE (Ours)} \\
\midrule
Backdoor Mean (\texttt{BD-Dyn}, \texttt{BD-Rew}) & $17.82$ & $3.79$ & $1.47$ & $2.88$ & $3.30$ & $31.76$ & $\mathbf{0.08}$ & $\underline{0.98}$ \\
Regime Shift Mean (\texttt{Shift-Fric}, \texttt{Shift-Mass}) & $17.86$ & $3.79$ & $\underline{3.78}$ & $7.53$ & $8.60$ & $34.99$ & $\mathbf{0.78}$ & $7.06$ \\
\midrule
All 12 Tasks (Mean $\pm$ SD) & \ms{17.84}{0.20} & \ms{3.79}{0.05} & \mssec{2.62}{1.21} & \ms{5.21}{2.43} & \ms{5.95}{2.77} & \ms{33.37}{1.80} & \msbest{0.43}{0.41} & \ms{4.02}{4.01} \\
\bottomrule
\end{tabular}
\end{table}

\section{Cross-Architecture Transfer to DreamerV3}
\label{app:dreamerv3}

To verify that ADE generalizes beyond continuous Gaussian RSSMs, we conduct an exploratory single-seed cross-architecture validation by applying the identical hyperparameter configuration (Table~\ref{tab:ade-hyperparams}) and five-step causal cascade directly to discrete categorical DreamerV3~\cite{hafner_mastering_2025} ($32\times 4$ one-hot categorical latents, $8\times 64$ block-recurrent GRU, RMSNorm/SiLU activations, and 255-bin symexp two-hot reward head; $522{,}571$ total parameters, $33.96\%$ edited) trained for $250{,}000$ LaProp+AGC steps~\cite{ziyin_laprop_2021} on Hopper-v2 (Seed~42). Specifically, Step~1 applies Mode~A and joint GWER ($\tau = 0.5\times 131 = 65.5$) to the latent-action input bottleneck blocks; Step~2 applies Mode~A and embedding-slice GWER to the posterior logit projection; Step~3 applies block-diagonal Mode~C top-$3$ GWER ($\tau = 0.5\times 512 = 256$) across the 8 recurrent GRU blocks and folds the damping into the recurrent transition projections; Step~4 applies Mode~A to the first observation decoder layer; and Step~5 applies Mode~B equal-risk calibration to the categorical prior logit head and 255-bin two-hot reward head. We evaluate all four counterfactual scenarios across all three audit layers (Table~\ref{tab:dreamerv3}). On out-of-manifold backdoors, ADE achieves strong zero-shot transfer: \texttt{BD-Dyn} $\mathrm{VFR}$ drops from $1.00$ to $0.00$ with $98.4\%$ lower $\mathrm{NDA}$ ($3.5381 \to 0.0573$), and \texttt{BD-Rew} $\Delta r_{\mathrm{trig}}$ drops from $+97.06$ to $-0.01$ ($\mathrm{Plan}\text{-}\rho$ $0.093 \to 0.788$). On in-manifold physical shifts, \texttt{Shift-Fric} improves in both closed-loop discrimination ($\mathrm{CLI}$ $+0.173 \to +0.279$) and 50-step rollout error ($\mathrm{MSE}_f^{50}$ $0.0746 \to 0.0654$), whereas \texttt{Shift-Mass} shows more modest recovery on the discrete block-GRU architecture than on continuous DreamerV1 ($\mathrm{CLI}$ rises from $-0.169$ to $0.000$ vs.\ Oracle $+0.464$ and $\mathrm{D4RL}_{\mathrm{clean}}$ increases from $30.10$ to $41.82$, while $\mathrm{MSE}_f^{50}$ remains at $0.4274$ vs.\ Origin $0.4266$).

\begin{table}[htbp]
\centering
\caption{Cross-architecture transfer to discrete categorical DreamerV3 on Hopper-v2 (Seed~42), executed with the exact same unified hyperparameter point $(\eta=0.99, \lambda_B=0.1)$ as DreamerV1.}
\label{tab:dreamerv3}
\vskip 0.06in
\scriptsize
\setlength{\tabcolsep}{3.5pt}
\begin{tabular}{llccccccc}
\toprule
Scen. & Model & $\mathrm{NDA}\downarrow$ & $\mathrm{MSE}_r^{50}\downarrow$ & $\mathrm{MSE}_f^{50}\downarrow$ & $\mathrm{Plan}\text{-}\rho\uparrow$ & $\mathrm{VFR}\downarrow / \Delta r_{\mathrm{trig}}\to 0$ & $\mathrm{D4RL}_{\mathrm{clean}}\uparrow$ & $\mathrm{D4RL}_{\mathrm{adv}}\uparrow$ \\
\midrule
\texttt{BD-Dyn} & Oracle & $0.0000$ & $0.00906$ & $0.0251$ & $0.906$ & \texttt{VFR:}$0.00$ & $36.50$ & $33.25$ \\
\texttt{BD-Dyn} & Origin & $3.5381$ & $0.00708$ & $0.1855$ & $0.758$ & \texttt{VFR:}$1.00$ & $33.82$ & $36.29$ \\
\texttt{BD-Dyn} & \textbf{ADE (Ours)} & $0.0573$ & $0.00710$ & $0.0252$ & $0.799$ & \texttt{VFR:}$0.00$ & $29.54$ & $32.76$ \\
\cmidrule(lr){1-9}
\texttt{BD-Rew} & Oracle & $0.0000$ & $0.00906$ & $0.0105$ & $0.896$ & $\Delta r$:$-0.03$ & $36.50$ & $35.31$ \\
\texttt{BD-Rew} & Origin & $541.5912$ & $0.00830$ & $0.0110$ & $0.093$ & $\Delta r$:$+97.06$ & $39.65$ & $34.11$ \\
\texttt{BD-Rew} & \textbf{ADE (Ours)} & $0.0051$ & $0.00752$ & $0.0104$ & $0.788$ & $\Delta r$:$-0.01$ & $36.24$ & $34.68$ \\
\midrule
Scen. & Model & $\mathrm{NDA}\downarrow$ & $\mathrm{MSE}_r^{50}\downarrow$ & $\mathrm{MSE}_f^{50}\downarrow$ & $\mathrm{CLI}\uparrow$ & $\Delta R_{\mathrm{im}}$ & $\mathrm{D4RL}_{\mathrm{clean}}\uparrow$ & $\mathrm{D4RL}_{\mathrm{adv}}\uparrow$ \\
\midrule
\texttt{Shift-Fric} & Oracle & $0.0000$ & $0.00906$ & $0.0798$ & $+0.549$ & $-2.8$ & $36.50$ & $36.67$ \\
\texttt{Shift-Fric} & Origin & $0.0491$ & $0.00833$ & $0.0746$ & $+0.173$ & $-1.8$ & $35.01$ & $37.24$ \\
\texttt{Shift-Fric} & \textbf{ADE (Ours)} & $0.0507$ & $0.00780$ & $0.0654$ & $+0.279$ & $-2.5$ & $31.77$ & $37.13$ \\
\cmidrule(lr){1-9}
\texttt{Shift-Mass} & Oracle & $0.0000$ & $0.00906$ & $0.2871$ & $+0.464$ & $-0.3$ & $36.50$ & $34.74$ \\
\texttt{Shift-Mass} & Origin & $0.0345$ & $0.00912$ & $0.4266$ & $-0.169$ & $+1.0$ & $30.10$ & $37.09$ \\
\texttt{Shift-Mass} & \textbf{ADE (Ours)} & $0.0373$ & $0.00836$ & $0.4274$ & $-0.000$ & $+0.6$ & $41.82$ & $37.41$ \\
\bottomrule
\end{tabular}
\end{table}

\section{Complete Hyperparameter Sensitivity Analysis}
\label{app:sensitivity}

\begin{figure}[htbp]
\centering
\includegraphics[width=\textwidth]{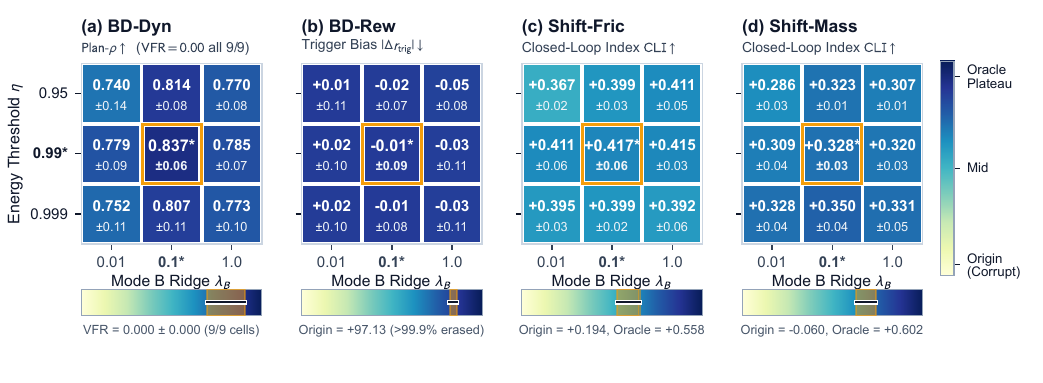}
\caption{\textbf{ADE maintains a flat performance plateau across the $3 \times 3$ hyperparameter grid without per-task tuning.} Sensitivity analysis over energy threshold $\eta \in \{0.95, 0.99, 0.999\}$ and Mode~B ridge penalty $\lambda_B \in \{0.01, 0.1, 1.0\}$ on Hopper-v2 (3-seed Mean $\pm$ SD; Table~\ref{tab:sensitivity}) for (a)~\texttt{BD-Dyn} ($\mathrm{Plan}\text{-}\rho \uparrow$, with $\mathrm{VFR} = 0.000$ across all $9/9$ cells), (b)~\texttt{BD-Rew} ($\Delta r_{\mathrm{trig}} \to 0$), (c)~\texttt{Shift-Fric} ($\mathrm{CLI} \uparrow$), and (d)~\texttt{Shift-Mass} ($\mathrm{CLI} \uparrow$). The unified default operating point $(\eta = 0.99, \lambda_B = 0.1)^*$ is highlighted in orange.}
\label{fig:sensitivity-heatmap}
\end{figure}

Table~\ref{tab:sensitivity} and Figure~\ref{fig:sensitivity-heatmap} report the complete $3\times 3$ sensitivity grid over $(\eta, \lambda_B) \in \{0.95, 0.99, 0.999\} \times \{0.01, 0.1, 1.0\}$ on Hopper-v2 across all four scenarios (3-seed Mean $\pm$ SD). Performance remains tightly stable across the entire grid: \texttt{BD-Dyn} achieves zero $\mathrm{VFR}$ across all 9 cells, \texttt{BD-Rew} maintains $|\Delta r_{\mathrm{trig}}| \le 0.05$ across all 9 cells, and \texttt{Shift-Fric} and \texttt{Shift-Mass} retain high $\mathrm{CLI}$ and $\mathrm{D4RL}_{\mathrm{clean/adv}}$ returns throughout, confirming that ADE requires no per-task hyperparameter search.

\begin{table}[htbp]
\centering
\caption{Complete $3 \times 3$ hyperparameter sensitivity grid over $(\eta, \lambda_B) \in \{0.95, 0.99, 0.999\} \times \{0.01, 0.1, 1.0\}$ on Hopper-v2 across all four regimes (3-seed Mean $\pm$ SD). The unified default point $(0.99, 0.1)$ used across all tasks is marked with $^*$.}
\label{tab:sensitivity}
\vskip 0.06in
\scriptsize
\setlength{\tabcolsep}{3.3pt}
\begin{tabular}{llccccccc}
\toprule
Scen. & $(\eta, \lambda_B)$ & $\mathrm{NDA}\downarrow$ & $\mathrm{MSE}_r^{50}\downarrow$ & $\mathrm{MSE}_f^{50}\downarrow$ & $\mathrm{Plan}\text{-}\rho\uparrow$ & $\mathrm{VFR}\downarrow / \Delta r_{\mathrm{trig}}\to 0$ & $\mathrm{D4RL}_{\mathrm{clean}}\uparrow$ & $\mathrm{D4RL}_{\mathrm{adv}}\uparrow$ \\
\midrule
\multirow{9}{*}{\texttt{BD-Dyn}} & $(0.95, 0.01)$ & \ms{0.5968}{0.2944} & \ms{0.1936}{0.1361} & \ms{0.2081}{0.0309} & \ms{0.7398}{0.1352} & \texttt{VFR:}\ms{0.000}{0.000} & \ms{36.73}{8.43} & \ms{35.02}{3.38} \\
 & $(0.95, 0.1)$ & \ms{0.5961}{0.2930} & \ms{0.2140}{0.1238} & \ms{0.2076}{0.0106} & \ms{0.8143}{0.0763} & \texttt{VFR:}\ms{0.000}{0.000} & \ms{34.44}{6.05} & \ms{36.07}{3.91} \\
 & $(0.95, 1)$ & \ms{0.5976}{0.2953} & \ms{0.1995}{0.0966} & \ms{0.2237}{0.0189} & \ms{0.7705}{0.0813} & \texttt{VFR:}\ms{0.000}{0.000} & \ms{34.61}{5.90} & \ms{35.66}{4.23} \\
 & $(0.99, 0.01)$ & \ms{0.5879}{0.2851} & \ms{0.1821}{0.1288} & \ms{0.2122}{0.0358} & \ms{0.7788}{0.0941} & \texttt{VFR:}\ms{0.000}{0.000} & \ms{33.41}{1.50} & \ms{34.84}{5.33} \\
 & \textbf{(0.99, 0.1)$^*$} & \ms{0.5883}{0.2846} & \ms{0.2031}{0.1183} & \ms{0.2104}{0.0149} & \ms{0.8371}{0.0580} & \texttt{VFR:}\ms{0.000}{0.000} & \ms{31.68}{2.27} & \ms{36.32}{4.89} \\
 & $(0.99, 1)$ & \ms{0.5899}{0.2872} & \ms{0.1915}{0.0952} & \ms{0.2269}{0.0245} & \ms{0.7850}{0.0705} & \texttt{VFR:}\ms{0.000}{0.000} & \ms{31.65}{0.81} & \ms{37.71}{4.45} \\
 & $(0.999, 0.01)$ & \ms{0.5898}{0.2658} & \ms{0.1686}{0.1038} & \ms{0.1759}{0.0506} & \ms{0.7518}{0.1129} & \texttt{VFR:}\ms{0.000}{0.000} & \ms{31.82}{5.15} & \ms{32.42}{3.80} \\
 & $(0.999, 0.1)$ & \ms{0.5908}{0.2654} & \ms{0.1879}{0.0971} & \ms{0.1829}{0.0304} & \ms{0.8069}{0.1060} & \texttt{VFR:}\ms{0.000}{0.000} & \ms{33.32}{2.66} & \ms{31.95}{3.87} \\
 & $(0.999, 1)$ & \ms{0.5924}{0.2678} & \ms{0.1828}{0.0892} & \ms{0.2128}{0.0345} & \ms{0.7730}{0.0960} & \texttt{VFR:}\ms{0.000}{0.000} & \ms{30.62}{1.93} & \ms{32.12}{3.54} \\
\cmidrule(lr){1-9}
\multirow{9}{*}{\texttt{BD-Rew}} & $(0.95, 0.01)$ & \ms{0.4564}{0.1783} & \ms{0.3211}{0.3395} & \ms{0.2411}{0.1271} & \ms{0.7063}{0.1554} & $\Delta r$:\ms{+0.01}{0.11} & \ms{31.35}{3.80} & \ms{32.87}{6.61} \\
 & $(0.95, 0.1)$ & \ms{0.4544}{0.1802} & \ms{0.3293}{0.3538} & \ms{0.2215}{0.1059} & \ms{0.6415}{0.1795} & $\Delta r$:\ms{-0.02}{0.07} & \ms{32.45}{9.02} & \ms{30.60}{4.30} \\
 & $(0.95, 1)$ & \ms{0.4535}{0.1822} & \ms{0.3183}{0.3249} & \ms{0.2211}{0.1037} & \ms{0.4804}{0.3832} & $\Delta r$:\ms{-0.05}{0.08} & \ms{33.96}{3.89} & \ms{31.92}{4.22} \\
 & $(0.99, 0.01)$ & \ms{0.4529}{0.1773} & \ms{0.2960}{0.3238} & \ms{0.2461}{0.1226} & \ms{0.7029}{0.1910} & $\Delta r$:\ms{+0.02}{0.10} & \ms{22.76}{14.49} & \ms{26.39}{11.38} \\
 & \textbf{(0.99, 0.1)$^*$} & \ms{0.4515}{0.1784} & \ms{0.3252}{0.3689} & \ms{0.2447}{0.1291} & \ms{0.6269}{0.2133} & $\Delta r$:\ms{-0.01}{0.09} & \ms{24.83}{14.75} & \ms{24.68}{12.39} \\
 & $(0.99, 1)$ & \ms{0.4510}{0.1785} & \ms{0.3229}{0.3471} & \ms{0.2451}{0.1363} & \ms{0.4631}{0.3799} & $\Delta r$:\ms{-0.03}{0.11} & \ms{26.91}{15.31} & \ms{24.83}{14.15} \\
 & $(0.999, 0.01)$ & \ms{0.4632}{0.1761} & \ms{0.2629}{0.2713} & \ms{0.2306}{0.1083} & \ms{0.6737}{0.1798} & $\Delta r$:\ms{+0.02}{0.10} & \ms{35.41}{1.06} & \ms{33.53}{5.67} \\
 & $(0.999, 0.1)$ & \ms{0.4613}{0.1774} & \ms{0.3050}{0.3361} & \ms{0.2213}{0.0928} & \ms{0.5948}{0.2406} & $\Delta r$:\ms{-0.01}{0.08} & \ms{33.57}{1.79} & \ms{33.11}{5.24} \\
 & $(0.999, 1)$ & \ms{0.4603}{0.1780} & \ms{0.3198}{0.3412} & \ms{0.2445}{0.1191} & \ms{0.4652}{0.4071} & $\Delta r$:\ms{-0.03}{0.11} & \ms{32.05}{1.41} & \ms{32.45}{6.01} \\
\midrule
Scen. & $(\eta, \lambda_B)$ & $\mathrm{NDA}\downarrow$ & $\mathrm{MSE}_r^{50}\downarrow$ & $\mathrm{MSE}_f^{50}\downarrow$ & $\mathrm{Plan}\text{-}\rho\uparrow$ & $\mathrm{CLI}\uparrow$ & $\mathrm{D4RL}_{\mathrm{clean}}\uparrow$ & $\mathrm{D4RL}_{\mathrm{adv}}\uparrow$ \\
\midrule
\multirow{9}{*}{\texttt{Shift-Fric}} & $(0.95, 0.01)$ & \ms{0.0205}{0.0015} & \ms{0.0625}{0.0151} & \ms{0.1246}{0.0248} & \ms{0.8691}{0.0533} & \ms{+0.3668}{0.0190} & \ms{33.24}{2.17} & \ms{34.08}{4.11} \\
 & $(0.95, 0.1)$ & \ms{0.0205}{0.0016} & \ms{0.0649}{0.0197} & \ms{0.1171}{0.0130} & \ms{0.8500}{0.0430} & \ms{+0.3990}{0.0282} & \ms{33.51}{1.41} & \ms{33.49}{3.32} \\
 & $(0.95, 1)$ & \ms{0.0209}{0.0016} & \ms{0.0627}{0.0180} & \ms{0.1167}{0.0230} & \ms{0.8131}{0.0142} & \ms{+0.4110}{0.0520} & \ms{34.50}{3.08} & \ms{33.40}{2.64} \\
 & $(0.99, 0.01)$ & \ms{0.0198}{0.0035} & \ms{0.0644}{0.0259} & \ms{0.1299}{0.0248} & \ms{0.8614}{0.0300} & \ms{+0.4112}{0.0604} & \ms{36.45}{3.90} & \ms{34.81}{4.91} \\
 & \textbf{(0.99, 0.1)$^*$} & \ms{0.0200}{0.0037} & \ms{0.0695}{0.0316} & \ms{0.1235}{0.0170} & \ms{0.8480}{0.0384} & \ms{+0.4172}{0.0569} & \ms{36.77}{5.69} & \ms{35.98}{8.09} \\
 & $(0.99, 1)$ & \ms{0.0203}{0.0036} & \ms{0.0674}{0.0266} & \ms{0.1262}{0.0334} & \ms{0.8149}{0.0253} & \ms{+0.4152}{0.0319} & \ms{35.71}{6.56} & \ms{34.65}{5.05} \\
 & $(0.999, 0.01)$ & \ms{0.0196}{0.0036} & \ms{0.0617}{0.0214} & \ms{0.1213}{0.0170} & \ms{0.8778}{0.0122} & \ms{+0.3947}{0.0258} & \ms{35.37}{4.71} & \ms{35.04}{5.03} \\
 & $(0.999, 0.1)$ & \ms{0.0197}{0.0036} & \ms{0.0684}{0.0293} & \ms{0.1231}{0.0142} & \ms{0.8582}{0.0409} & \ms{+0.3985}{0.0240} & \ms{34.38}{1.50} & \ms{33.22}{3.60} \\
 & $(0.999, 1)$ & \ms{0.0198}{0.0034} & \ms{0.0679}{0.0276} & \ms{0.1286}{0.0358} & \ms{0.8261}{0.0260} & \ms{+0.3922}{0.0601} & \ms{36.20}{3.84} & \ms{34.02}{4.69} \\
\cmidrule(lr){1-9}
\multirow{9}{*}{\texttt{Shift-Mass}} & $(0.95, 0.01)$ & \ms{0.0290}{0.0050} & \ms{0.1387}{0.0599} & \ms{0.2428}{0.0248} & \ms{0.8762}{0.0068} & \ms{+0.2861}{0.0319} & \ms{37.06}{2.43} & \ms{35.34}{1.51} \\
 & $(0.95, 0.1)$ & \ms{0.0294}{0.0049} & \ms{0.1376}{0.0580} & \ms{0.2698}{0.0223} & \ms{0.8281}{0.0078} & \ms{+0.3234}{0.0120} & \ms{37.63}{3.16} & \ms{34.91}{2.93} \\
 & $(0.95, 1)$ & \ms{0.0295}{0.0043} & \ms{0.1150}{0.0472} & \ms{0.3258}{0.0396} & \ms{0.7914}{0.0113} & \ms{+0.3075}{0.0052} & \ms{38.90}{2.91} & \ms{36.67}{2.20} \\
 & $(0.99, 0.01)$ & \ms{0.0289}{0.0067} & \ms{0.1107}{0.0419} & \ms{0.2398}{0.0176} & \ms{0.8846}{0.0121} & \ms{+0.3090}{0.0414} & \ms{38.82}{0.91} & \ms{37.46}{0.51} \\
 & \textbf{(0.99, 0.1)$^*$} & \ms{0.0292}{0.0067} & \ms{0.1090}{0.0431} & \ms{0.2535}{0.0090} & \ms{0.8341}{0.0121} & \ms{+0.3276}{0.0254} & \ms{40.02}{1.83} & \ms{37.11}{1.60} \\
 & $(0.99, 1)$ & \ms{0.0293}{0.0065} & \ms{0.0927}{0.0342} & \ms{0.3072}{0.0386} & \ms{0.7998}{0.0228} & \ms{+0.3200}{0.0348} & \ms{42.17}{4.04} & \ms{36.18}{1.17} \\
 & $(0.999, 0.01)$ & \ms{0.0293}{0.0065} & \ms{0.1133}{0.0447} & \ms{0.2445}{0.0186} & \ms{0.8821}{0.0139} & \ms{+0.3283}{0.0381} & \ms{35.84}{1.67} & \ms{36.79}{1.08} \\
 & $(0.999, 0.1)$ & \ms{0.0297}{0.0065} & \ms{0.1121}{0.0465} & \ms{0.2591}{0.0099} & \ms{0.8322}{0.0161} & \ms{+0.3499}{0.0391} & \ms{36.80}{3.25} & \ms{36.73}{0.62} \\
 & $(0.999, 1)$ & \ms{0.0298}{0.0062} & \ms{0.0949}{0.0365} & \ms{0.3104}{0.0449} & \ms{0.8026}{0.0276} & \ms{+0.3311}{0.0534} & \ms{35.74}{2.07} & \ms{37.28}{1.64} \\
\bottomrule
\end{tabular}
\end{table}

\section{Scope and Future Work}
\label{app:future-work}

This work evaluates ADE on proprioceptive state-space benchmarks so that recurrent dynamics unlearning can be audited directly against exact single-step and multi-step physical counterfactuals without pixel-reconstruction confounding. Extending zero-training excision to pixel-based world models introduces an orthogonal perceptual coupling: visual triggers and appearance-linked regime shifts can entangle deep nonlinear CNN or ViT encoders and transposed-convolutional decoders, whereas ADE preserves nonlinear encoders and operates analytically on affine bottlenecks and readouts. Jointly solving front-end perceptual unlearning alongside back-end analytical dynamics excision, as well as adapting ADE to reconstruction-free contrastive or JEPA-style latent world models~\cite{assran_self-supervised_2023}, is a natural next direction. Finally, strengthening architectural adaptation across the broader Dreamer family, such as tailoring closed-form edits to discrete categorical transitions and block-recurrent dynamics, alongside extending analytical excision to Transformer-based world models~\cite{micheli_transformers_2022}, offers a natural path forward.

\end{document}